\pdfoutput=1
\documentclass{article}
\usepackage{ijcai22}
\usepackage{times}
\usepackage{soul}
\usepackage{url}
\usepackage[hidelinks]{hyperref}
\usepackage[utf8]{inputenc}
\usepackage[small]{caption}
\usepackage{graphicx}
\usepackage{amsmath}
\usepackage{amsthm}
\usepackage{booktabs}
\usepackage{algorithm}
\usepackage{algorithmic}
\usepackage{color}
\usepackage[dvipsnames]{xcolor}
\title{Zero-Reference Image Restoration for Under-Display Camera of UAV}

\author{Zhuoran Zheng$^{1}$, Xiuyi Jia$^{1}$\thanks{Corresponding authors.} and Yunliang Zhuang$^{2}$\\
	$^{1}$CSE, Nanjing University of Science and Technology~~~
	$^{2}$SISE, Shandong Normal University ~~~ \\
}

\begin{document}

\maketitle

\begin{abstract}
  The exposed cameras of UAV can shake, shift, or even malfunction under the influence of harsh weather, while the add-on devices (Dupont lines) are very vulnerable to damage. 
  We can place a low-cost T-OLED overlay around the  camera to protect it, but this would also introduce image degradation issues. 
  In particular, the temperature variations in the atmosphere can create mist that adsorbs to the T-OLED, which can cause secondary disasters (i.e., more severe image degradation) during the UAV's filming process. 
  To solve the image degradation problem caused by overlaying T-OLEDs, in this paper we propose a new method to enhance the visual experience by enhancing the texture and color of images. 
  Specifically, our method trains a lightweight network to estimate a low-rank affine grid on the input image, and then utilizes the grid to enhance the input image at block granularity. 
  The advantages of our method are that no reference image is required and the loss function is developed from visual experience. 
  In addition, our model can perform high-quality recovery of images of arbitrary resolution in real time.
  In the end, the limitations of our model and the collected datasets (including the daytime and nighttime scenes) are discussed.
  %The harsh challenges of full-screen devices are position a camera behind a screen, especially for UAVs.
   
\end{abstract}

\begin{figure}[t] 
	\begin{center}
		\begin{tabular}{@{}c@{}}
			\includegraphics[width = 0.44\textwidth]{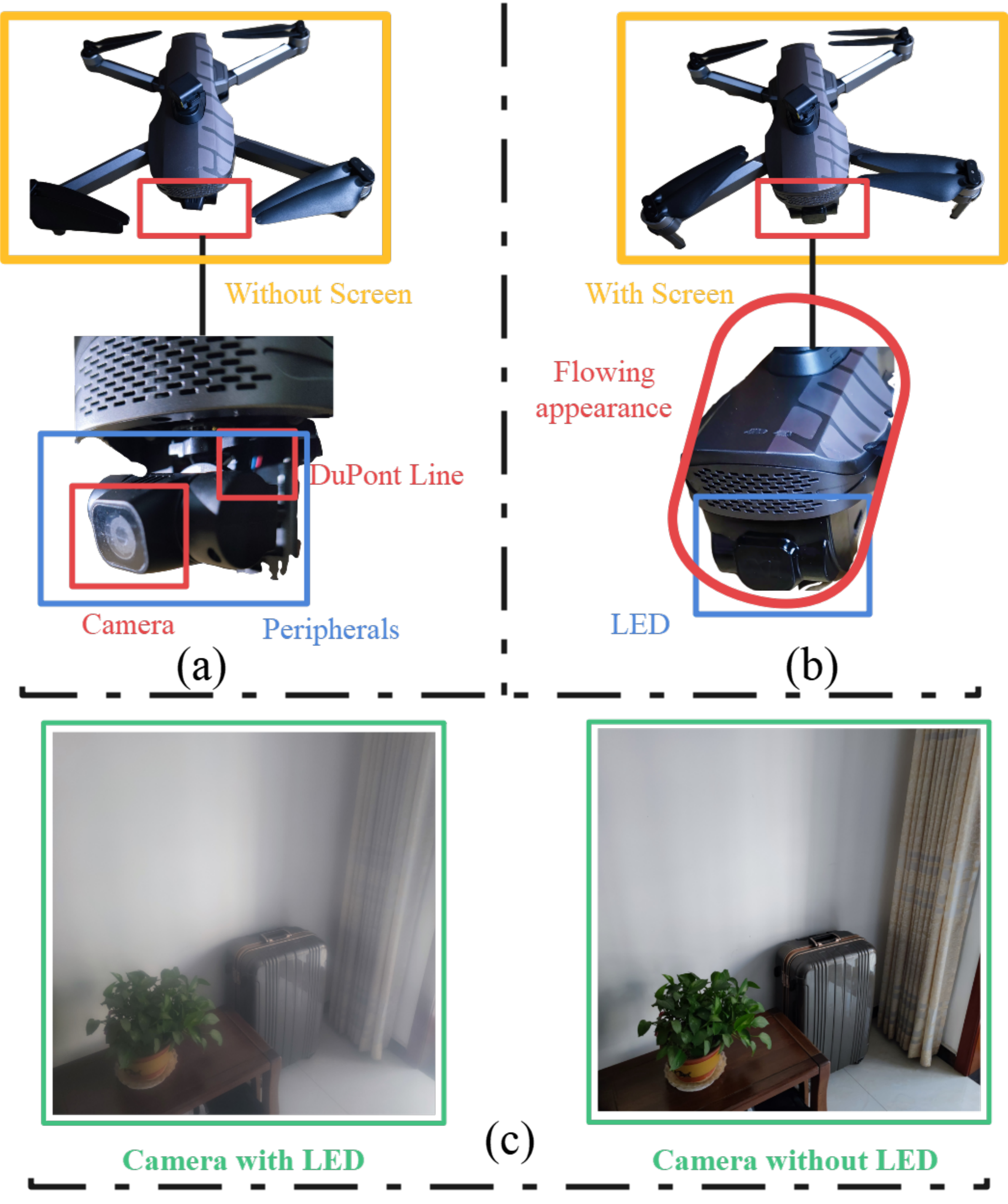}                   
		\end{tabular}
	\end{center}
	\vspace{-4mm}
	\caption{The newly proposed camera system on the UAV. We mount display screen on top of a traditional digital camera lens. Note the newly proposed method causes degradation of the image.} 
	
	%Our approach needs to consider images of arbitrary resolution since the resolution of images are different from photo (the resolution is $5000 \times 4000$) and video (the resolution is $1280 \times 720$) operations in camera systems.}
	\vspace{-5mm}
	\label{f1}
\end{figure}
\vspace{-4mm}
\section{Introduction}
\begin{figure*}[t] 
	\begin{center}
		\begin{tabular}{@{}c@{}}
			\includegraphics[width = 1\textwidth]{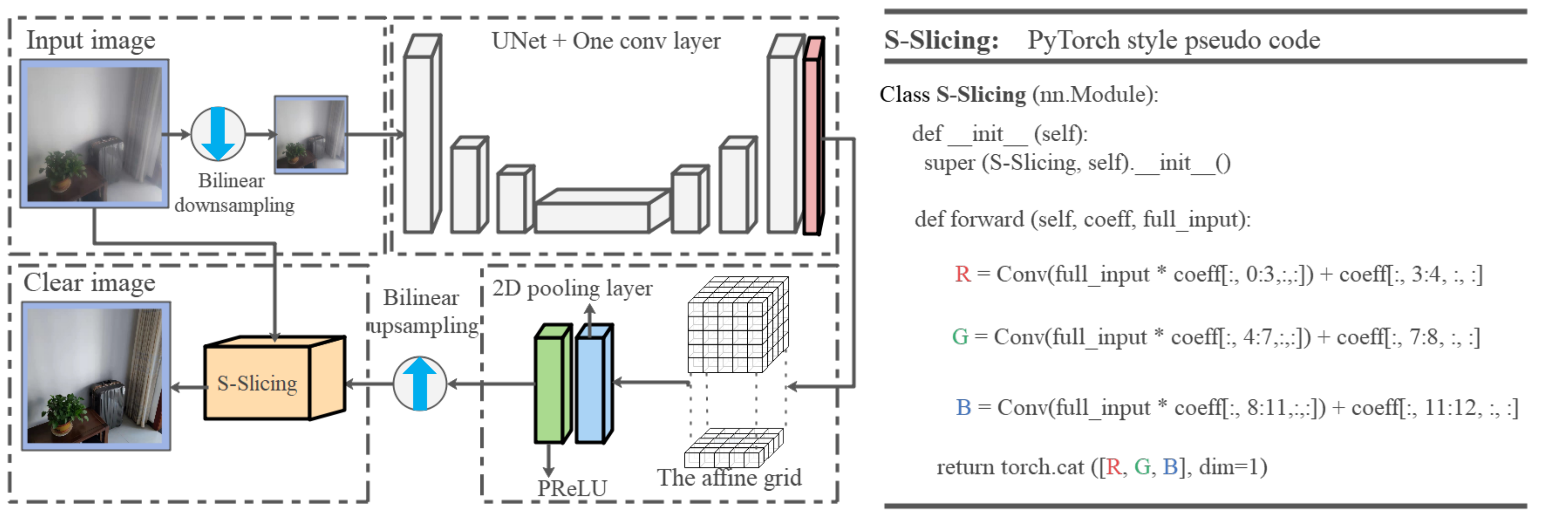}                   
		\end{tabular}
	\end{center}
	\vspace{-4mm}
	\caption{The architecture and configuration of the proposed single UDC restoration approach, which consists of four components. First, we downsample the input image (the resolution is $3 \times W \times H$) to obtain a fixed resolution image ($3 \times 256 \times 256$).
		Then we develop the UNet and extra convolution layer to generate an affine grid (the resolution is $12 \times 256 \times 256$). Next, we enforce a regular term to clean up the noise from the affine grid by using a 2D pooling layer (MaxPool). Final, we upsample the affine grid to yield a full resolution clear image (the resolution is $12 \times W \times H$) and apply it to the raw input image through the Smoothing Slicing operator (S-Slicing).
		We use filters (Conv (convolutional layer with kernel is 3)) to decompress the channels to get a smoother image. 
		}
	\vspace{-5mm}
	\label{framework}
\end{figure*}
%
%Under-display Camera (UDC) as a new image system, it reduces the burden on the device while also guarding the stability of the camera system.
%
%In extreme environments, the exposed camera of the UAV is vulnerable, due to wind and meteorological artifacts that can hit the exposed camera devices(see Figure~\ref{f1} (a)).
%
In extreme environments, the exposed cameras of UAVs are susceptible to wind and meteorological factors that can lead to camera failure.
To solve this issue, we design a low-cost scheme to keep the stability of the camera system of the exposed camera on the UAV, as shown in Figure~\ref{f1} (b).
Specifically, we use a low-cost Transparent OLED (T-OLED) to wrap around the camera device. 
%Inspired by the Under-display Camera (UDC), we design a simple scheme to guard the stability of the camera system of the exposed camera of the UAV, as shown in Figure~\ref{f1} (b).
%
%Such a design addresses two main problems.
%
%First, UAVs generally provide specialized anti-shake measures for the camera system, which obviously requires the affordance of more stand off devices and a large number of redundant models to be installed in the storage unit.
%
All-in-one design can provide better user perceptive (see Figure~\ref{f1} (b), for instance the flowing appearance) and other intelligent experience~\cite{irudc}. 
More importantly, this design offers a reference for saving the complicated peripherals around the camera, thereby reducing the weight of the UAV.
%
%Second, exposed devices suffer from haze, dust storms and other weather hinder the normal workings of the camera system, especially the exposed DuPont Line may be broken (see Figure~\ref{f1} (a)).
%
%For this, a protective cover can defend these sensitive components.
%
%To achieve convenience, we mount the T-OLED scree [?] on top of camera lens on the UAV.
%
However, this new system is mired in the same morass as the Under-Display Camera (UDC)~\cite{irudc,dagf,cirudc,removeudc,eccvworks,ddpludc}, because it is relatively difficult to retain full functionality of an imaging sensor after mounting it behind a display.
Specifically, the imaging quality of a camera will be degraded due to lower light transmission rate and diffraction effects~\cite{irudc}.
Moreover, the image quality degradation is further exacerbated because the temperature difference between the land and the sky creates a haze that adheres to the protector. 
As a result, the captured image clings to the surface of the image as if there is a white veil (the camera with T-OLED in Figure~\ref{f1} (c)).

Currently, image restoration methods focused on UDC are trained on a pair of the blurred and clear images~\cite{irudc,dagf}. 
Unfortunately, it is difficult for us to obtain a pair of the blurred and the clear images due to the instability of the UAV at high altitude.
In this study, we endeavor to implement the image restoration task without reference images by leveraging the UDC prior rather than considering the high-cost GAN techniques~\cite{englightengan,HistoGAN}.
In simple terms, our approach does not require any paired information in the training process.

\vspace{-1mm}
To achieve our goal, we propose a novel lightweight deep learning-based method, Zero-Reference Under-Display Camera Network (ZRUDC-Net), for UDC image enhancement on the UAV.
It can cope with a variety of luminance conditions, including uniform chromatic aberration and water haze disturbance.
Instead of performing image-to-image mapping, we reformulate the task as an image-specific affine estimation problem.
In detail, we estimate an affine transformation tensor (the channel is $12$) acting on the raw input image by using a smoothing slicing operator.
For the affine transformation tensor (affine grid), we can learn it end-to-end through the U-Net~\cite{UNET}.
%
%The proposed approach is lightweight and the unique advantage is \textbf{zero-reference}.
%\textbf{Second}, we have designed a \textbf{zero-reference} learning pattern.
%
Second, we develop a zero-reference learning pattern thanks to sophisticated technologies~\cite{dsd,dcploss,zerolli}.
This is achieved through a set of specially designed non-reference loss functions, including \textit{spatial consistency loss, exposure control loss, color constancy loss, \textit{dark and bright channel priors loss}, and light smoothing loss}.
%
%For example, the UAV will generate water mist adhering to the lens at high altitude, yielding an image similar to the haze uniformly distributed on the image, and the Dark Channel Prior (DCP) prior can tackle this problem efficiently.
%
At last, we discuss in the work the limitations and struggles imposed by collecting data sets, especially for nighttime scenes.

\vspace{-1mm}
In summary, the main contributions of our work are: (1) We empirically develop a pipeline based on the visual experience and use this pipeline to guide us in formulating the final solution, including the design of the network and the configuration of the loss functions.
(2) We propose the first zero-reference UDC enhancement network that is independent of paired and unpaired training information. Numerous experiments have demonstrated the effectiveness of our method.
(3) We collect the first UDC dataset on the UAV (diverse visual scenes are included) which will be released and evaluated by the public.

\vspace{-3mm}
\section{Related Work}

{\flushleft \textbf{UDC imaging and restoration.}} 
Several previous work proposed positioning the camera under the display panel to obtain high-resolution photos~\cite{dtp,irudc,icta}.
% 期刊
%In particular, there has been a focus on using Transparent-OLED (T-OLED)~\cite{dtp}, the display that are usually used in commercial televisions, cellphones and mobile devices.
%
T-OLED is a better display panel~\cite{dtp} and is commonly used in commercial televisions, cellphones and mobile devices.
%
%In UAVs, we try to use TOLEDs to defend the camera device while keeping the high quality image capturing capability.
%
% CVPR
Recently, several works~\cite{mlt,sti,dsti} described and analyzed the diffraction effects of UDC systems.
However, these works cannot tackle the image restoration problem.
Although several works~\cite{bidi,add,suh2012p} proposed camera-behind-display design, these are low-resolution images with very low-quality content and are not suitable for everyday photography.
%
%To address these challenges, ~\cite{irudc}and the ECCV challenge~\cite{eccvworks} try to address the problem of UDC image restoration through publicly released datasets.
%
Zhou et al.~\shortcite{irudc} and the ECCV challenge~\cite{eccvworks} attempted to address the problem of UDC image restoration with publicly released datasets.
However, the above methods have an obvious drawback that the non-clear images carry only one form of image degradation.
%
% 期刊
Some approaches~\cite{id,tdp} attempted to develop deep CNNs to handle the low SNR and large blur in under-panel imagery.
%
%Zhou et al. use the CNNs to deblur and denoise to generate high quality images under T-OLED.
%
%Sundar et al. [13] deblur on low-resolution images and conducts a guided filer network to restore a high resolution images.
%
%Puthussery et al. [14] develop a densely connected convolutional layers with different dilations acting on UNet to improve the performance of image restoration.
%
%Further, [15, 16] redesigned the UNet to restore the degraded images.
%
We also refer to the architecture of these methods to generate the required affine coefficients for the raw images.

{\flushleft \textbf{Unsupervised single image restoration.}} 
Currently numerous unsupervised techniques are employed for image enhancement tasks such as GAN-based approaches~\cite{HistoGAN,englightengan}.
%
%For conventional methods, there is a strong reliance on a priori knowledge of the image, which is not friendly for outdoor scenes. 
%
Although GAN-based methods can yield outstanding results in image restoration, such methods require paired or unpaired images as an auxiliary, which is costly.
Fortunately, Zero-Net~\cite{zerolli} provides an efficient and sophisticated idea to tackle the problem of image degradation. Therefore, we can refer this approach to solve the image degradation problem by combining dark and bright channel priors~\cite{dcploss,dbloss}.

\vspace{-2mm}
\section{Motivation and Methodology}
Our task is a blind image degradation problem, and the point spread function (PSF) is hard to estimate from the degraded image.
Because the photography conditions of the UAV in the air are enormously complex, with not only blur but also saturation overflow and other problems, we can only solve these ill-posed problems from visual experience.
Based on the intuitive visual experience and analysis, we design a new pipeline (see Figure~\ref{f3} (a) $\rightarrow$ (c)) that attempts to address the problem of image degradation on the UAV.
Finally, on the basis of this pipeline, we design a zero-reference image enhancement strategy.
\vspace{-2mm}
\subsection{Motivation}

\begin{figure}[t] 
	\scriptsize
	\begin{center}
		\begin{tabular}{@{}ccc@{}}
			\includegraphics[width = 0.14\textwidth]{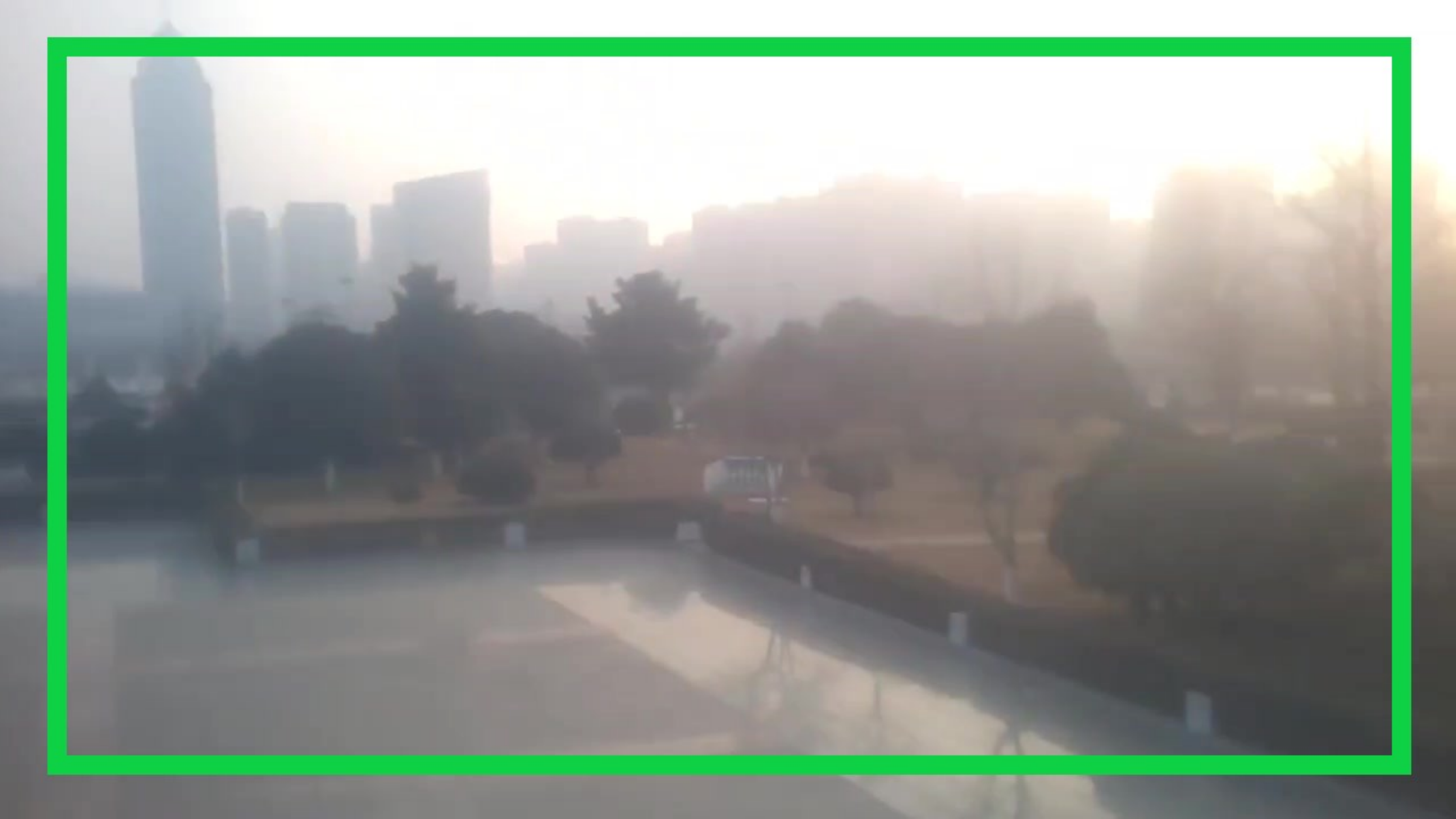} &
			\includegraphics[width = 0.14\textwidth]{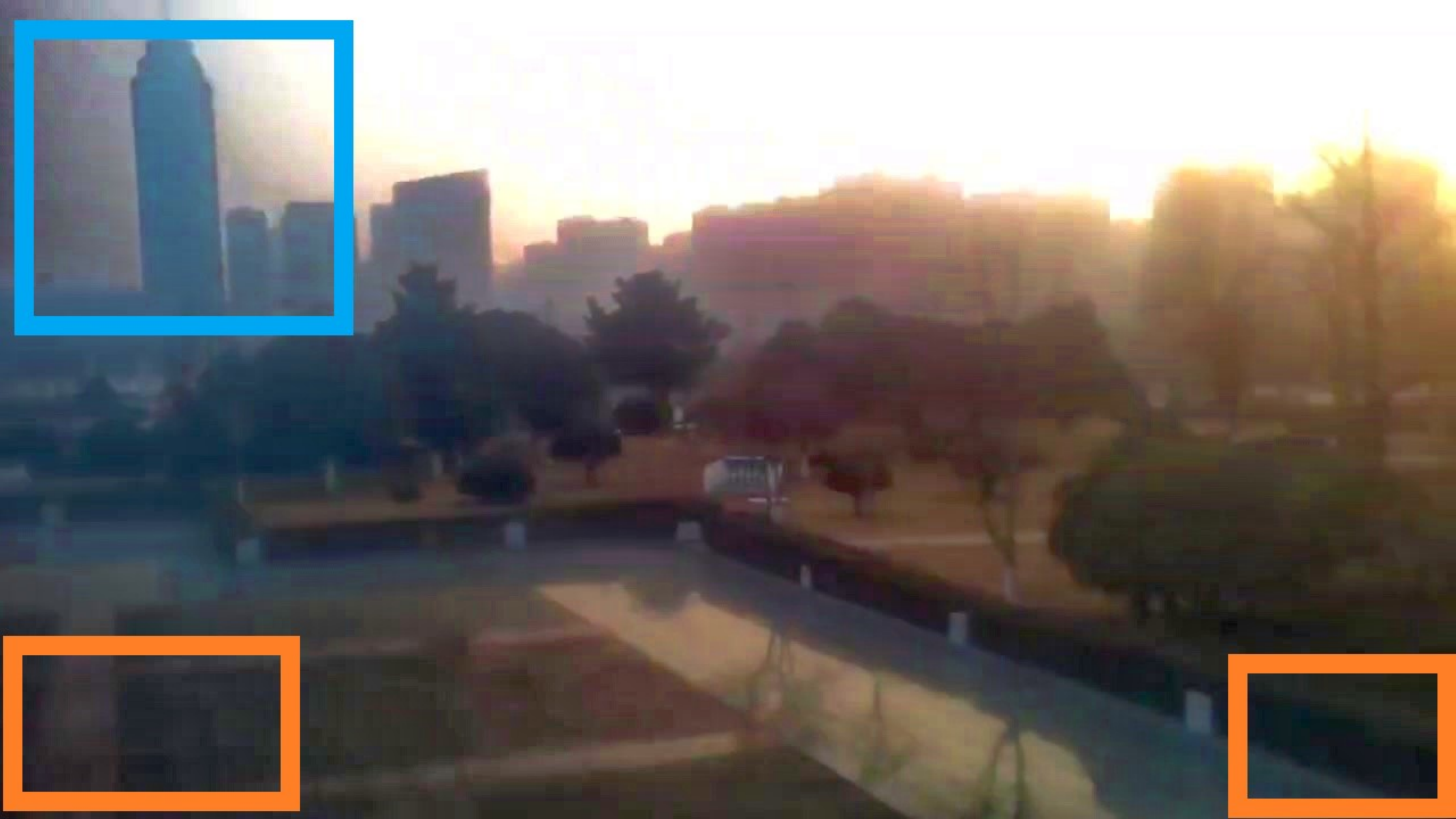}  &
			\includegraphics[width = 0.14\textwidth]{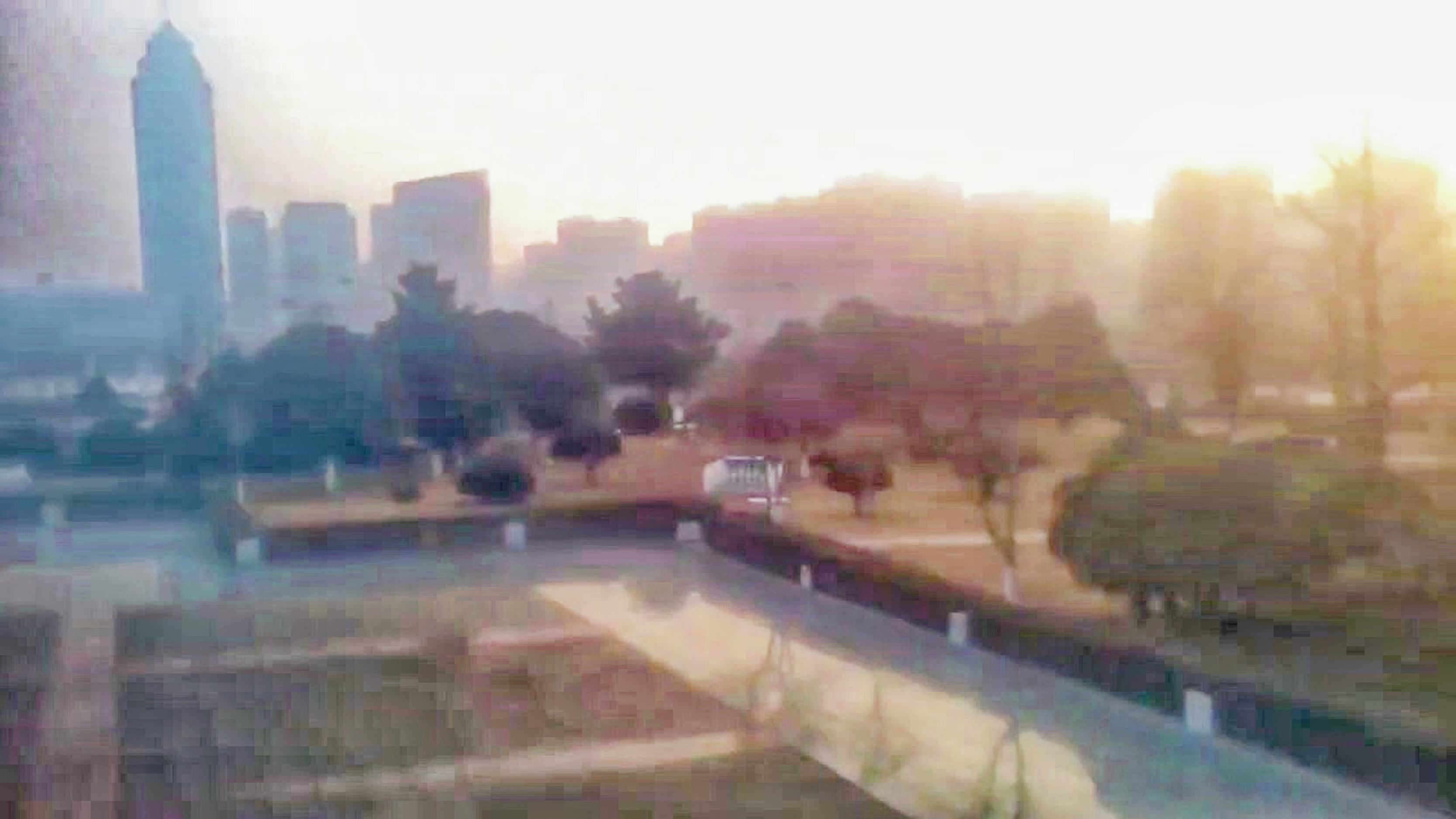}  \\  
			
			(a) Real-world input image &
			(b) DCP + Gamma &
			(c) Zero-Net \\

		\end{tabular}
	\end{center}
	\vspace{-4mm}
	\caption{We attempt to develop a pipeline to handle the image degradation problems ((a) $\rightarrow$ (b) $\rightarrow$ (c)).
		(a) indicates the issues caused by UAV photography with T-OLED in the real-world. (b) illustrates the results of our image enhancement using dark channel prior with Gamma correction, and the new hassles it brings. Zero-Net (LLE approach) is adopted to enhance the results of (b) in the (c).} 
	\vspace{-4mm}
	\label{f3}
\end{figure}

The conventional UDC problem (the image degradation) is usually addressed by an end-to-end network trained on a pair of blurred and clear images~\cite{cirudc,irudc}.
Unfortunately, since UAVs move video at high altitude, it is not easy to collect high-quality paired training data.
So far, we consider to enhance the degraded images based on our visual intuition.
%
%Notably GAN-based methods may be the way to solve such problems, but we aim to provide a simpler and more cost-effective scheme.
%
As shown in Figure~\ref{f3} (a), real-world images from the UAV with T-OLED can generate many cases of image degradation (overexposure, dark corners and blur, etc.). 
To solve these ill-posed problems, we propose a pipeline (Figure~\ref{f3} ($a \rightarrow b \rightarrow c$)) to enhance the images.
%
%The pipeline is considered from a \textbf{global perspective} and \textbf{local perspective}.
%
{\flushleft \textbf{The pipeline.}} We first need to solve the problem marked in \textcolor{green}{green box} in Figure~\ref{f3} (a). 
Based on visual experience, this is a case where the image is hazed.
DCP~\cite{DCP} is the most classical technique to solve the image with haze but it also leads to a series of problems, such as the whole image is lightless.
To mitigate this problem, Gamma correction is available as a general technique to relieve images from darkness.
We performed a combination of DCP and Gamma correction for Figure~\ref{f3} (a) and the results are shown in Figure~\ref{f3} (b). Although the haze-like phenomenon is resolved, it still leaves a massive spoil behind (see the \textcolor{blue}{artifacts} and the \textcolor{orange}{dark corner} parts in Figure~\ref{f3} (b)).
%
%The results as shown in Figure~\ref{f3} (b), we combined the above two techniques to enforce on Figure~\ref{f3} (a).
%
%Although, the haze-like phenomenon is solved by DCP and Gamma correction, it still left a massive spoil behind (see the Figure~\ref{f3} (b), \textcolor{blue}{artifacts} and \textcolor{orange}{dark corner}).
%
So we consider using a low-light image enhancement approach (Zero-Net) to further solve this problem, as shown in Figure~\ref{f3} (c).
The reason for considering the utilization of Zero-Net is that it is an unsupervised method, which provides us with technical support to solve the degradation problem of reference-free images.
Obviously, the overall image brightness is improved, but there is still some noise in local patches.

So far, we still have three unresolved challenges, one is sky artifacts, another is local blurring of the image, and the last is local overexposure.
Based on this empirical pipeline, we attempt to further enhance the visual performance by using the following solutions.
1) The dark channel prior loss~\cite{dcploss} is utilized to replace the DCP and Gamma correction strategy as it alleviates the problem of sky artifacts.
2) The loss functions provided by Zero-Net~\cite{zerolli} are used, and notably we have empirically adjusted the exposure loss function.
3) The brightness and the dark channel prior~\cite{dbloss} are also applied to modify the local blurring and over-exposure, and several such regular terms are enforced in this work.

Based on the above solutions, we design a zero-reference UDC-Net to run a degraded image of arbitrary (including ultra high definition images) resolution in real time.
\vspace{-2mm}
\subsection{ZRUDC-Net}
Given an arbitrary resolution input, our network first reconstructs the affine grid using a modified U-Net~\cite{UNET} on a fixed resolution of the input.
Capitalizing on the regressed affine grid, we can generate high-quality affine coefficients.
Moreover, to obtain the low-rank affine grid, we use pooling layers to limit its space characteristics.
Figure~\ref{framework} illustrates the architecture of the proposed single zero-reference UDC restoration approach, which consists of four parts: downsampling of the input image, generating an affine grid, obtaining the low-rank  affine grid and producing a clear image.

\begin{figure}[t] 
	\begin{center}
		\begin{tabular}{@{}c@{}}
			\includegraphics[width = 0.46\textwidth]{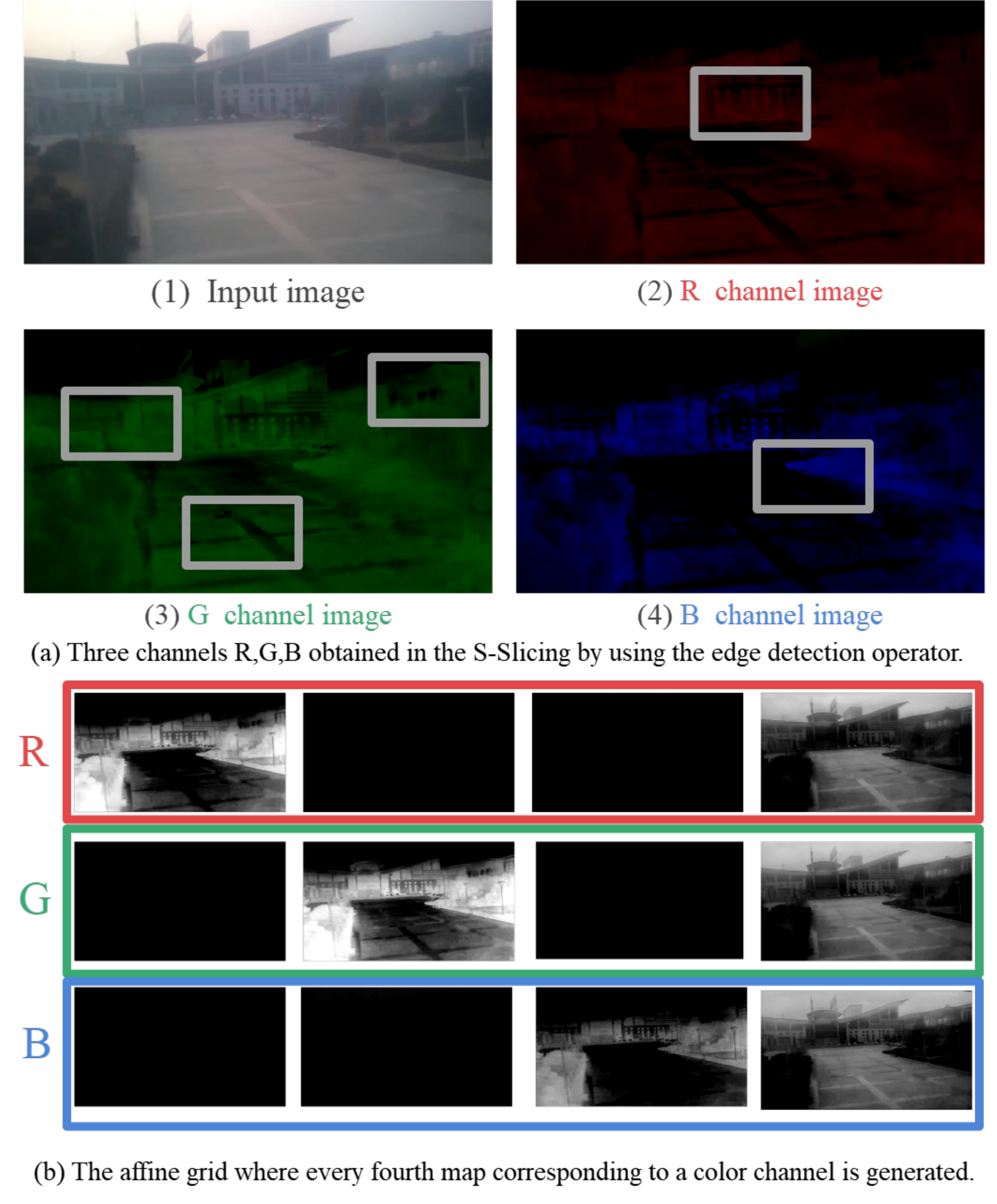}                   
		\end{tabular}
	\end{center}
	\vspace{-6mm}
	\caption{Coefficient maps for the affine color transform. R, G, B channel images generated by the affine grid from U-Net. White boxes indicate that using the affine grid generates richer edge and structure details in (a). (b) denotes the 12-channel characteristic of the affine grid, which can be clearly observed as an attention mechanism.} 
	\vspace{-4mm}
	\label{f4}
\end{figure}

{\flushleft \textbf{Affine grid learning.}} 
It is well-known that bilateral learning~\cite{hdrnet} in image restoration tasks is a well-reliable technique.
Its core lies the estimation of a bilateral grid to regress a high quality affine coefficient tensor (the channels are 12).
However, this depends heavily on the extraction of the guidance information, which is performed on a full-resolution image, and therefore only several convolution kernels can be used to estimate it (to reach the goal of real-time processing).
To further boost the modelling speed, we use the deep network directly to learn an affine grid.
First, we reduce the raw input image to a fixed resolution of $256 \times 256$ and extract low-level features by a U-Net and a convolution layer as shown in Figure~\ref{framework}.
This yields a $12 \times 256 \times 256$ array of the feature maps $T$ that contains rich textures and edges (see Figure~\ref{f4} (b)).
%
%This is due to the inherent characteristics of U-Net.
%
$T$ also can be viewed as a $256 \times 256$ information grid, where each grid cell contains 12 numbers,  one for each coefficient of a $3 \times 4$ affine transformation matrix.
%
%This workflow is more expressive than the bilateral grid learning, thanks to its extraordinary regression ability (not controlled by the guidance information).
%
Then we use the 2D pooling layer $MaxPool$ and a PReLU layer acting on $T$ to obtain a smaller scale $T^{'}$.
Specifically, a multi-channel affine grid $T$ whose third dimension has been unrolled:
\begin{equation}
T^{'}[x^{'}, y^{'}, c] = \text{PReLU} (MaxPool (T[x, y, c]))
\end{equation}
where $[x = 256, y=256]$ denotes the coordinates of the affine grid cell, and each cell has $c = 12$ number.
In addition, $[x^{'}, y^{'}, c]$ denotes the affine grid is downsampled in both height and width dimensions.
$T^{'}$ is regarded as an array of low rank for the full-resolution affine grid $T^{*}$ of the next stage.
That is because the procedure from $T^{'}$ to $T^{*}$ is implemented through the use of the linear operator (interpolation).
To demonstrate the effectiveness of the 2D pooling layer and the PReLU layer in the proposed network, we also performed an ablation study which is discussed in the subsequent section.

{\flushleft \textbf{Smoothing Slicing.}} 
Capitalizing on the predicted affine grid coefficient $T^{*}$,
we need to transfer this information back to the full-resolution content of the raw input to produce a high-quality clean image.
We introduce the S-Slicing operation on the three channels of color as shown Figure~\ref{f4}, and we give the PyTorch-style pseudo-code on the right of Figure~\ref{framework} for the details of the scheme.
Note that the bilateral learning uses an additive operator to squeeze the channels, while we use a learned convolution operator.
As shown in Figure~\ref{f4} (a), we visualize the by-products of the affine grids after the affine transformation of the original image, and we can see that the textures of all three of them complement each other and play a characteristic resembling that of bilateral learning.
This workflow maintains effectiveness while reducing inference time significantly.

\begin{figure*}[t]\scriptsize
	\begin{center}
		\tabcolsep 1pt
		\begin{tabular}{@{}ccccccc@{}}

			\includegraphics[width = 0.16\textwidth]{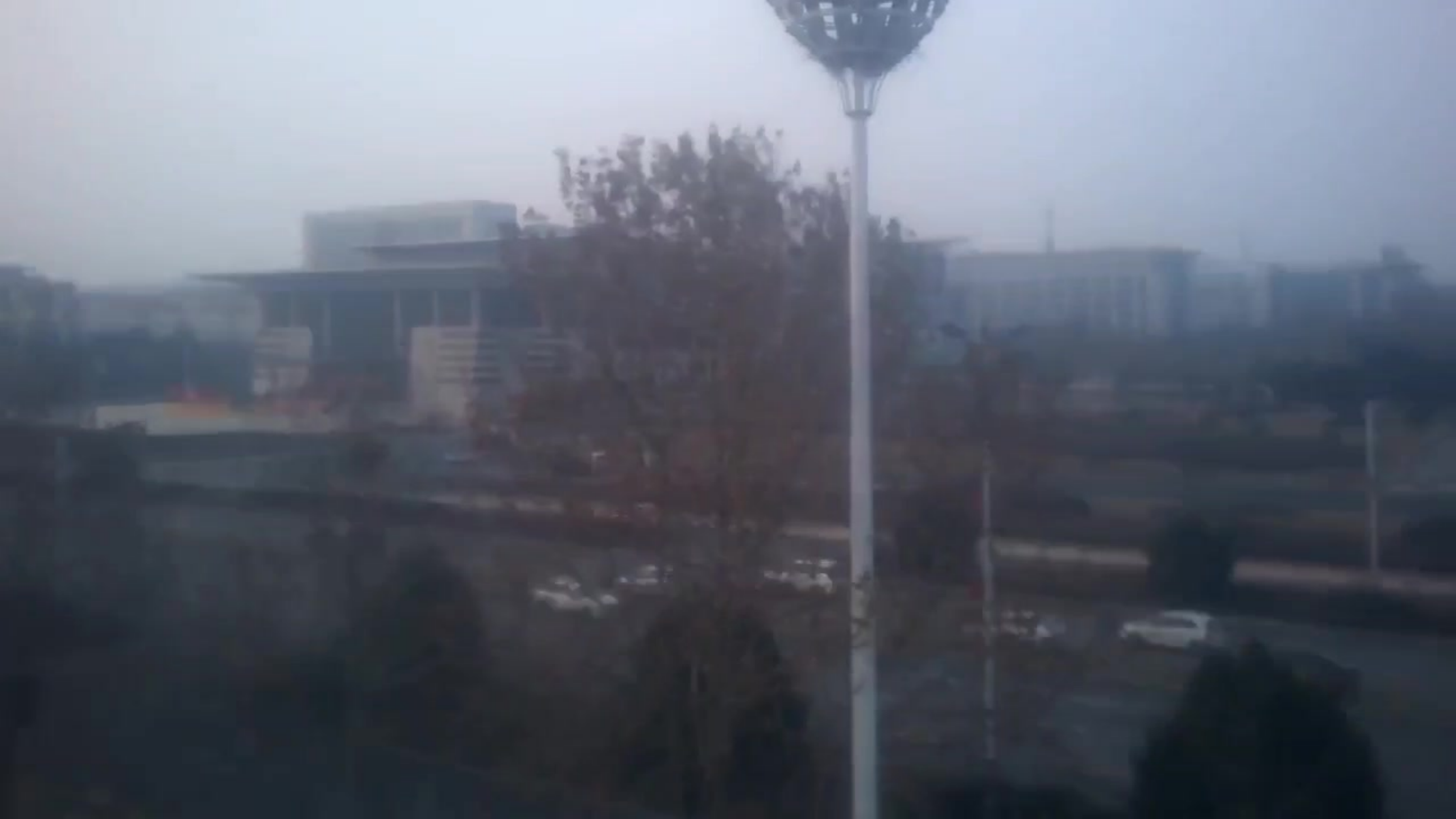}         &
\includegraphics[width = 0.16\textwidth]{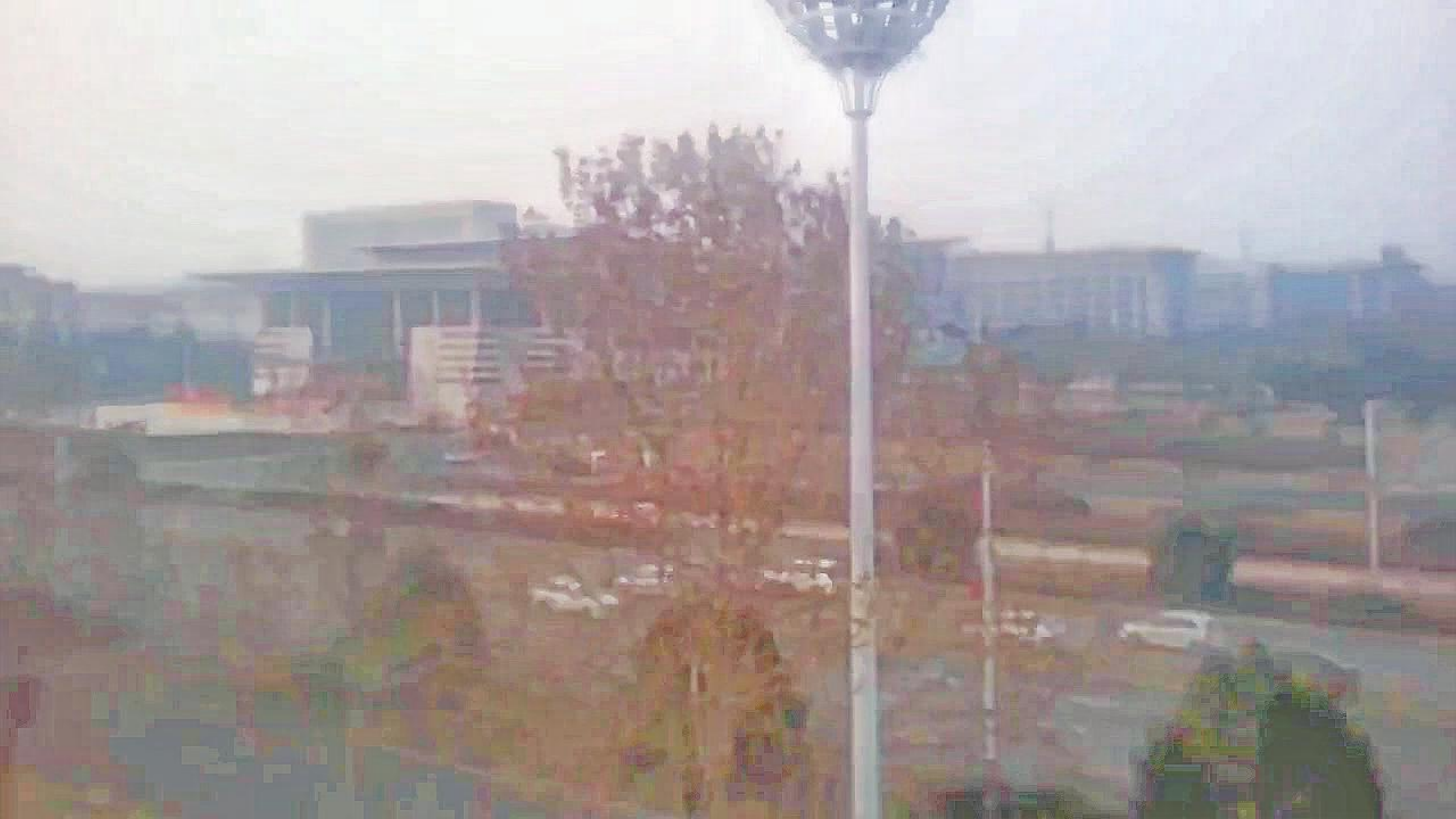}         &
\includegraphics[width = 0.16\textwidth]{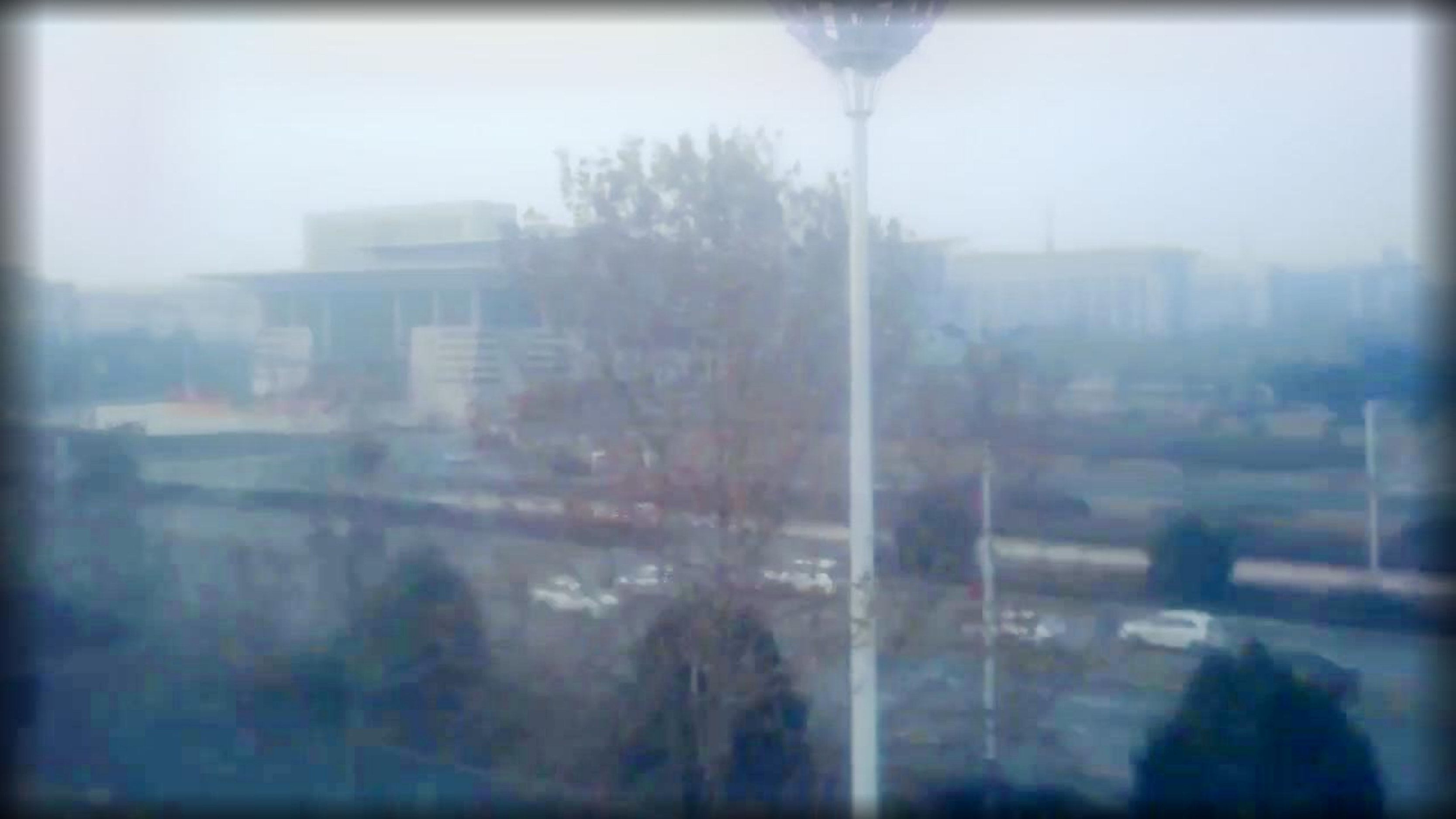}         &
\includegraphics[width = 0.16\textwidth]{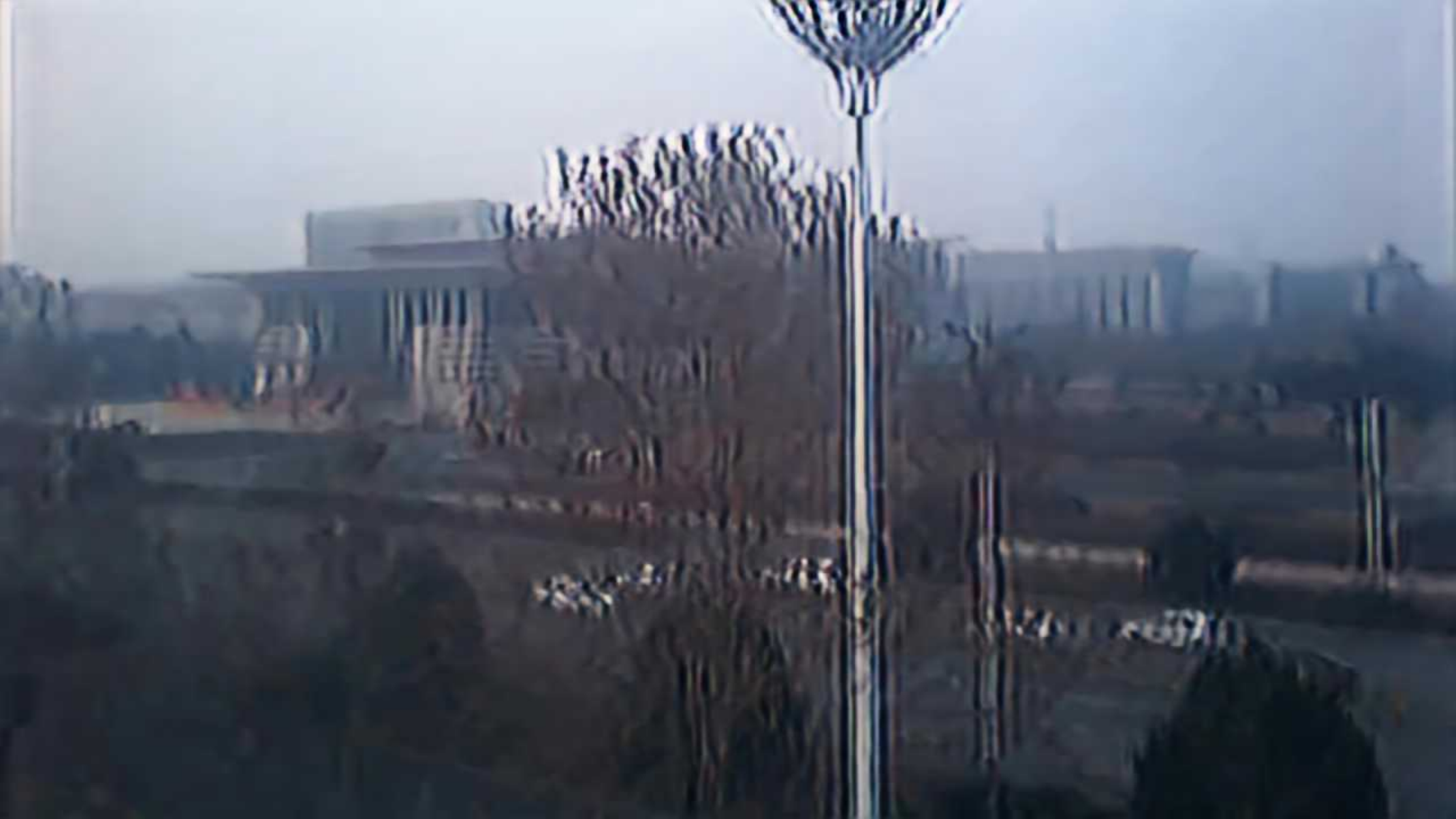}         &
\includegraphics[width = 0.16\textwidth]{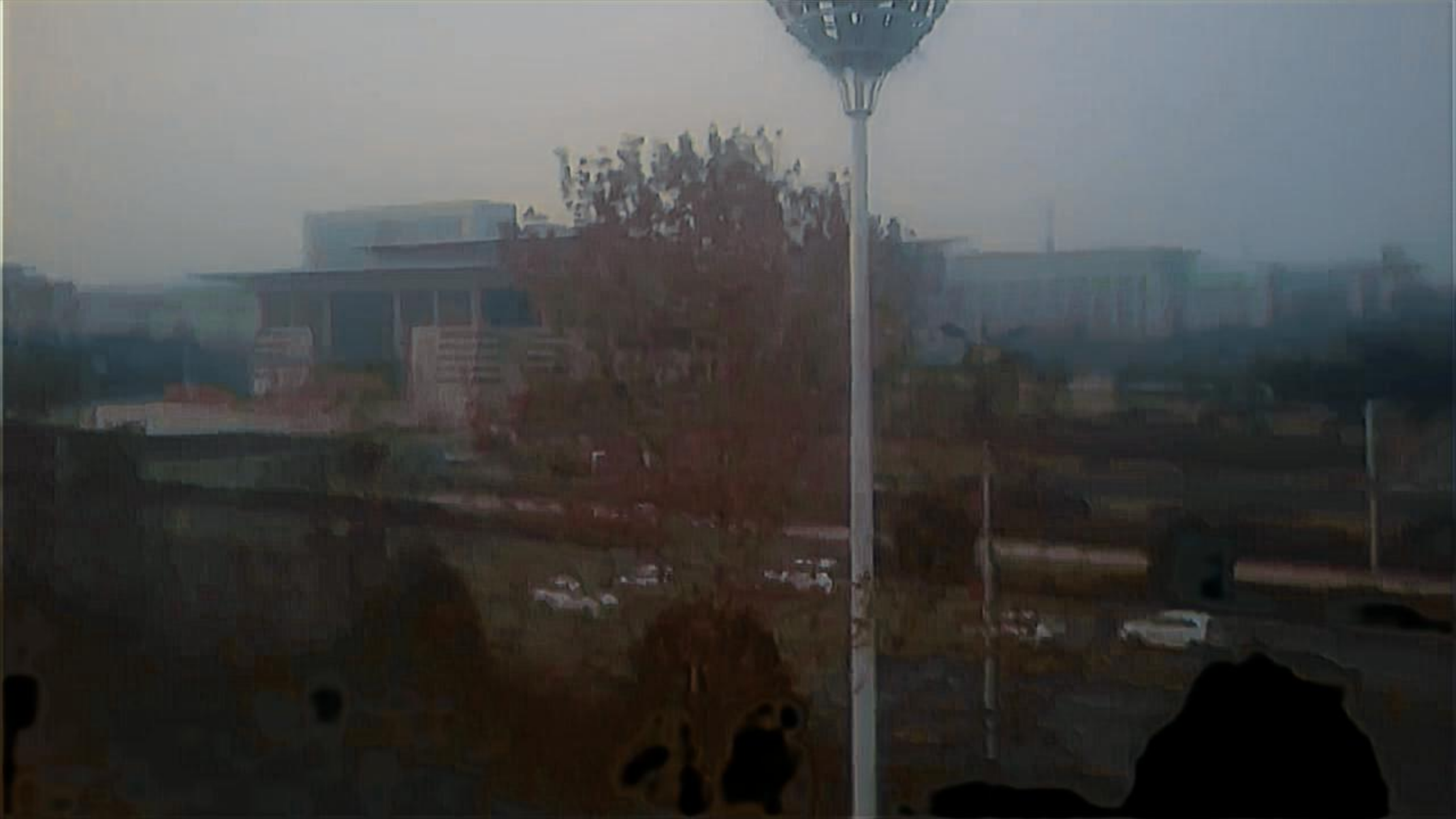}         &
\includegraphics[width = 0.16\textwidth]{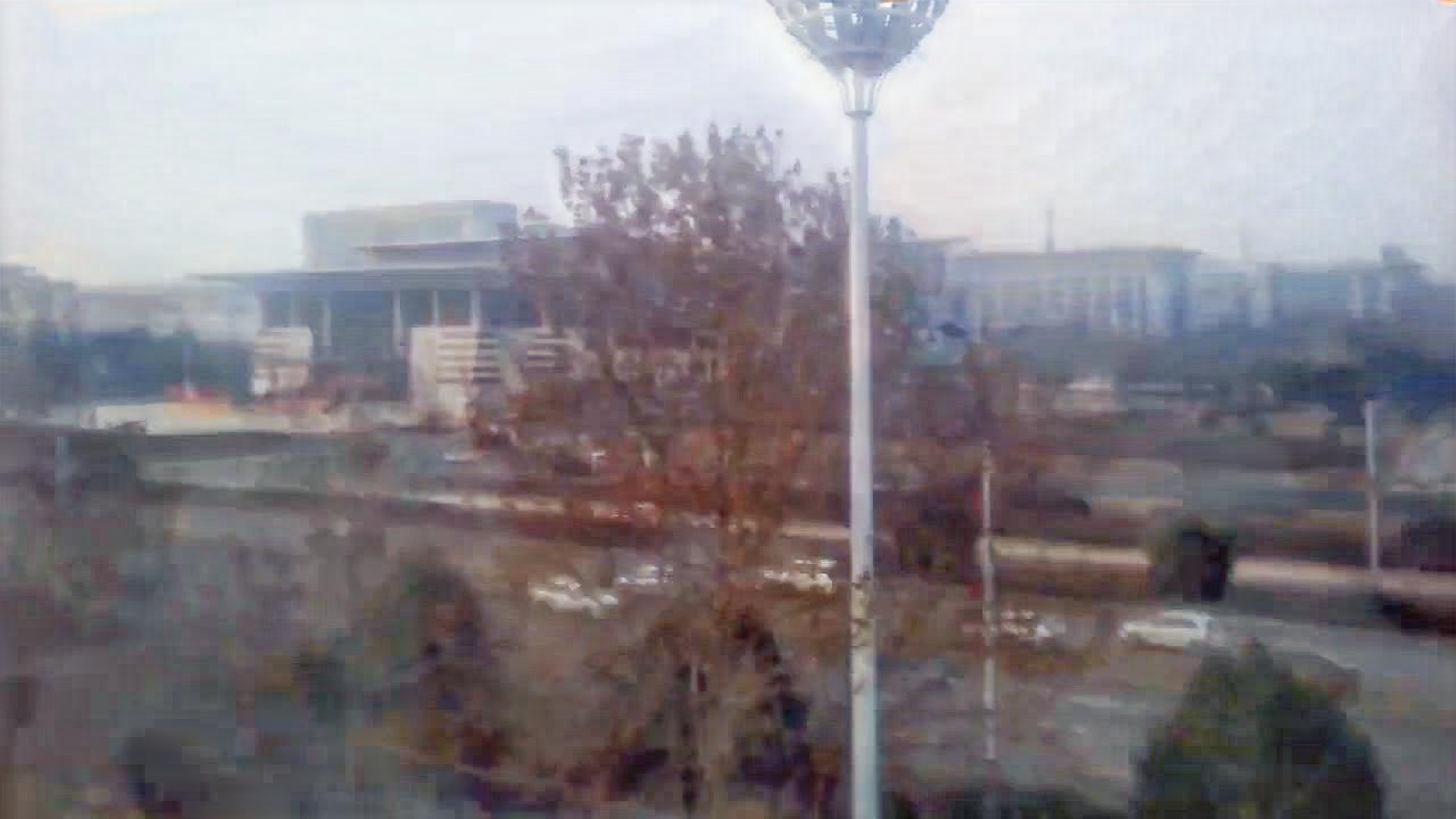}         \\

			\includegraphics[width = 0.16\textwidth]{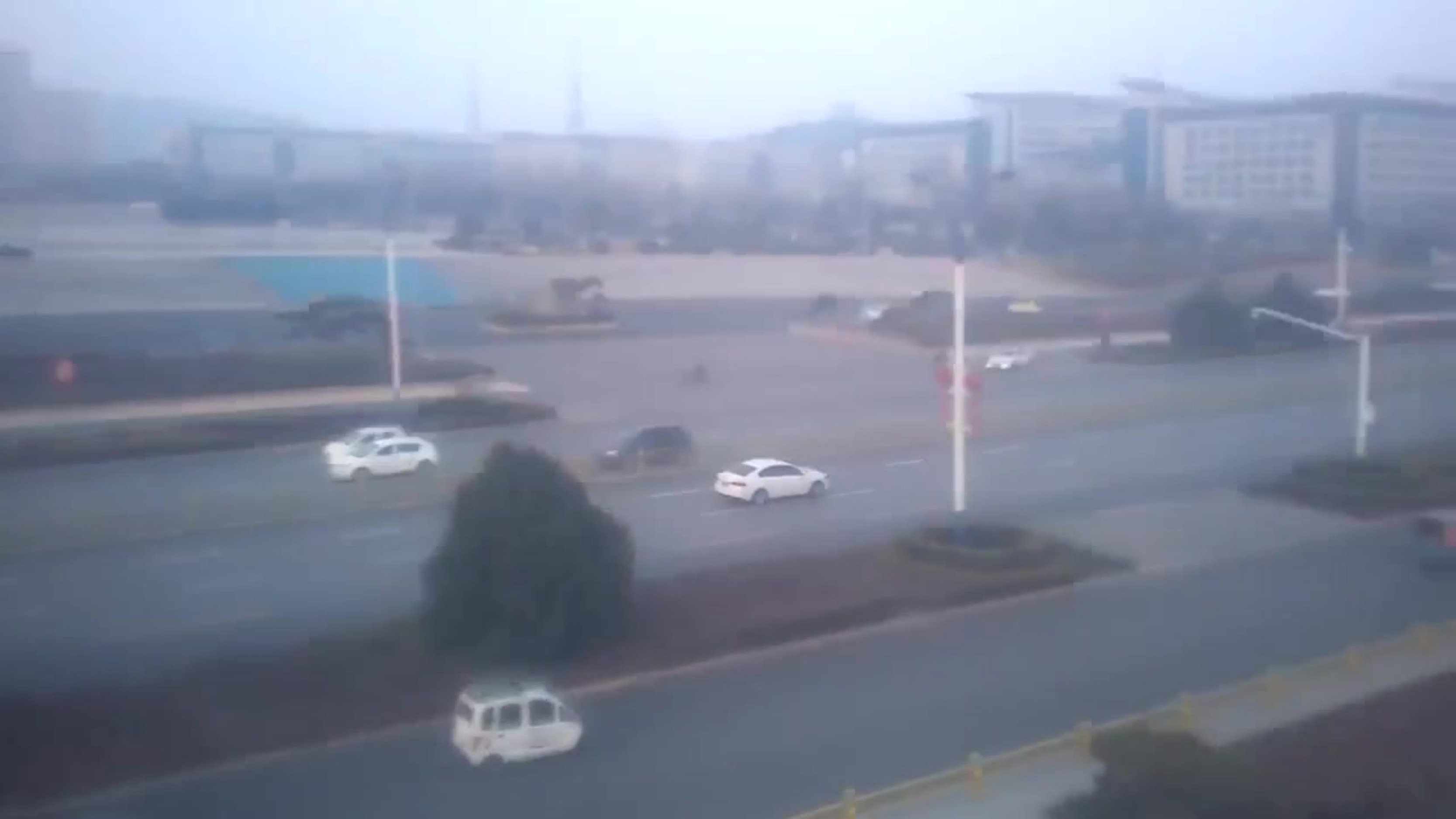}         &
\includegraphics[width = 0.16\textwidth]{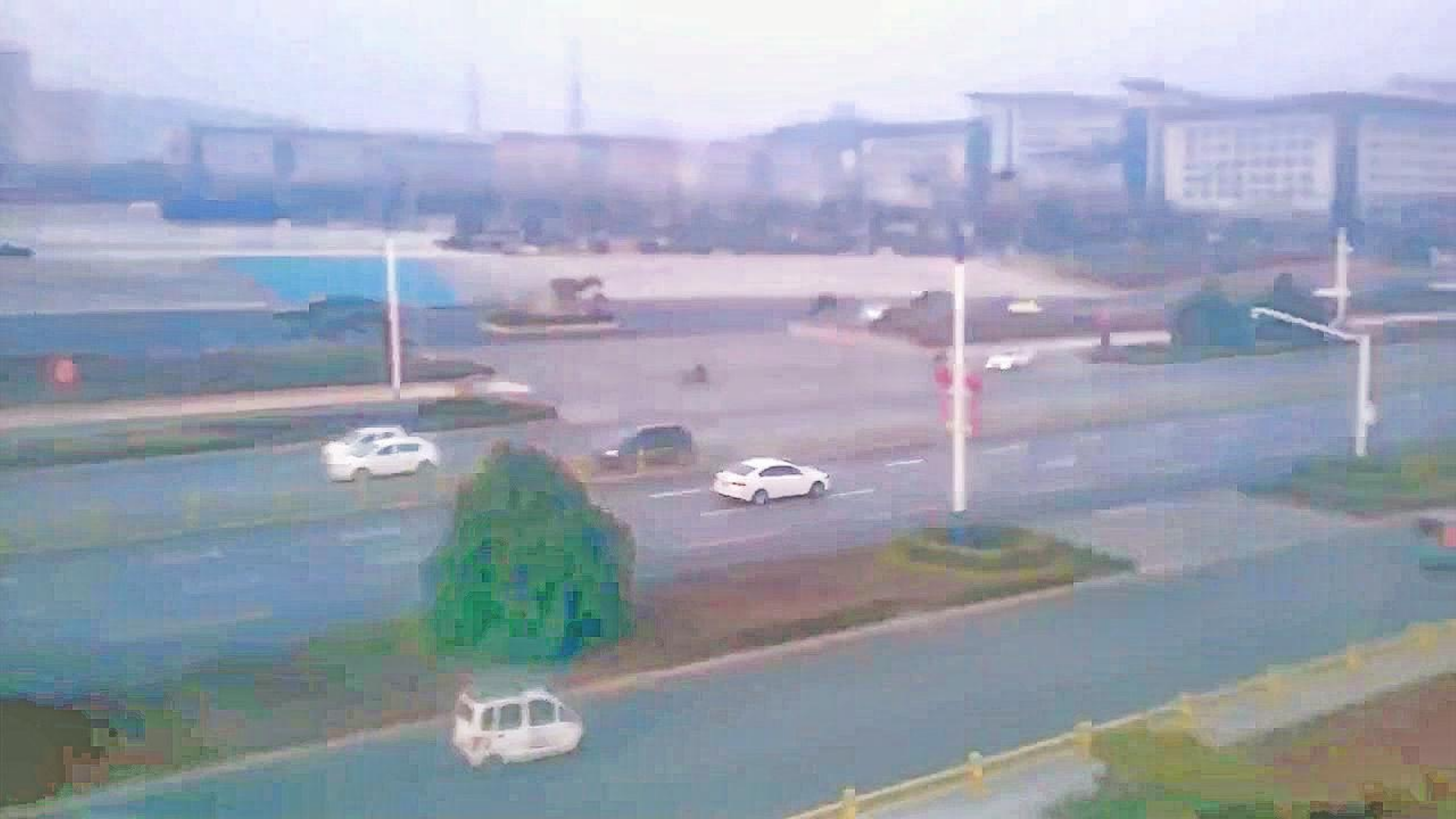}         &
\includegraphics[width = 0.16\textwidth]{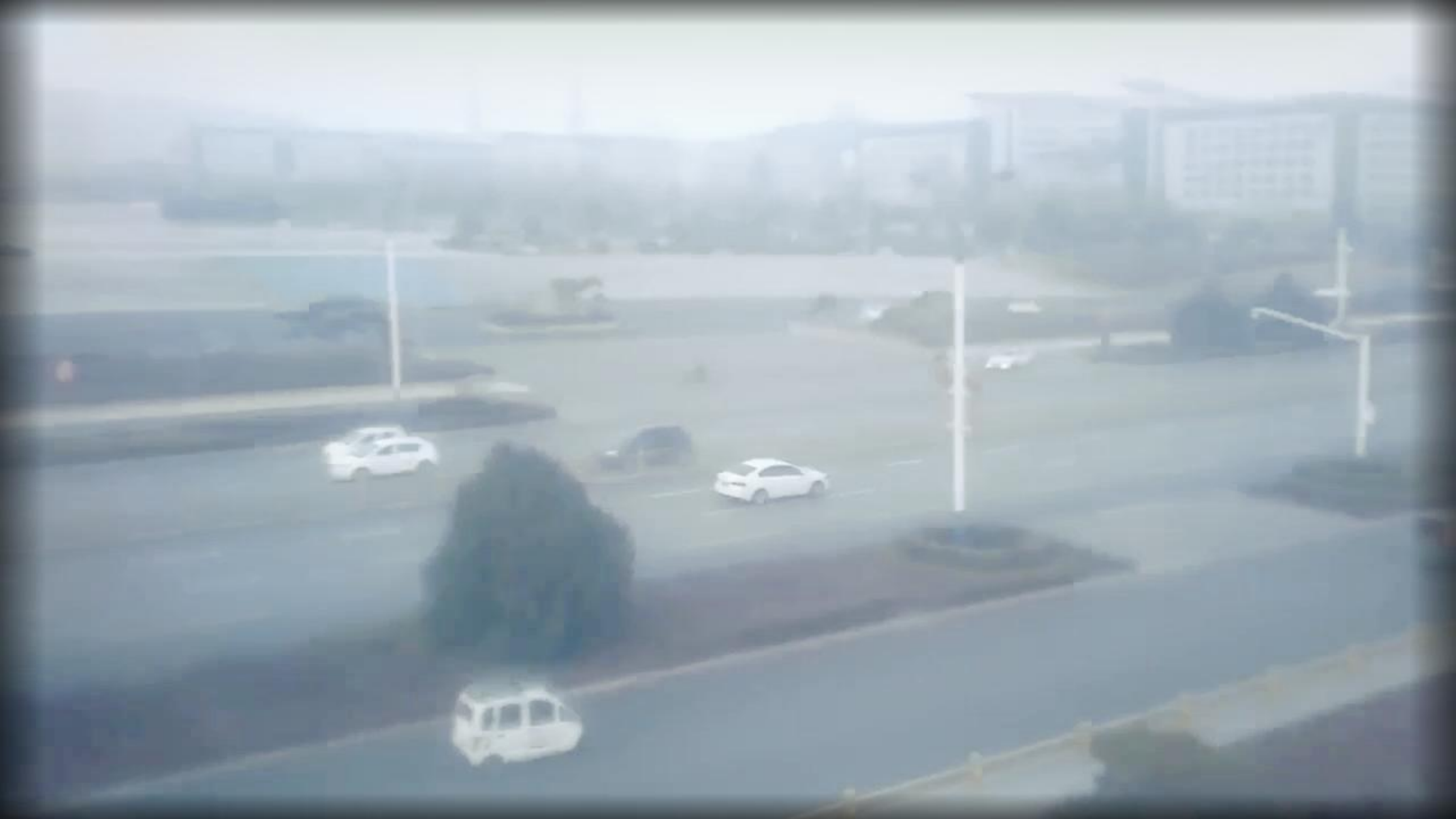}         &
\includegraphics[width = 0.16\textwidth]{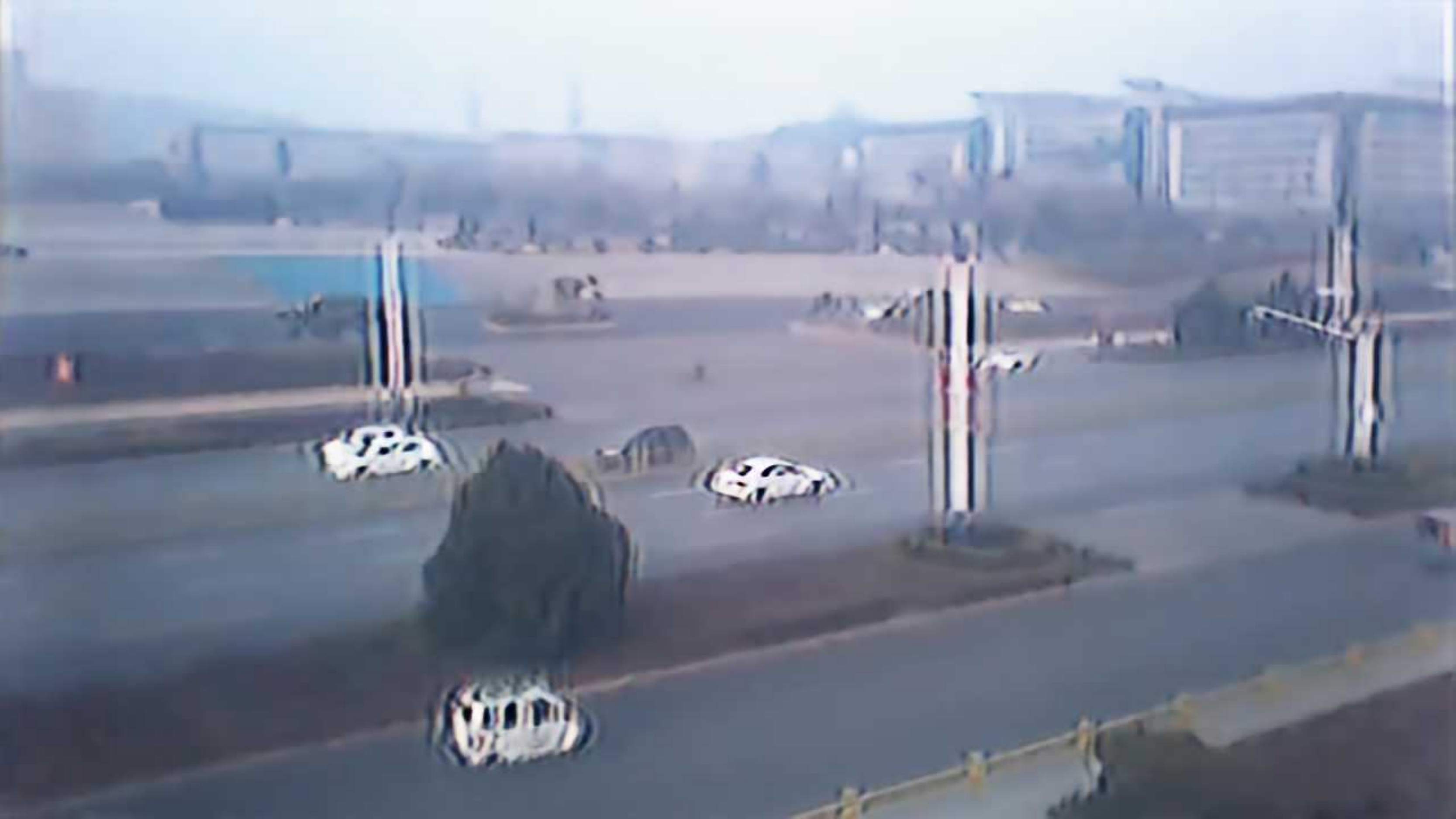}         &
\includegraphics[width = 0.16\textwidth]{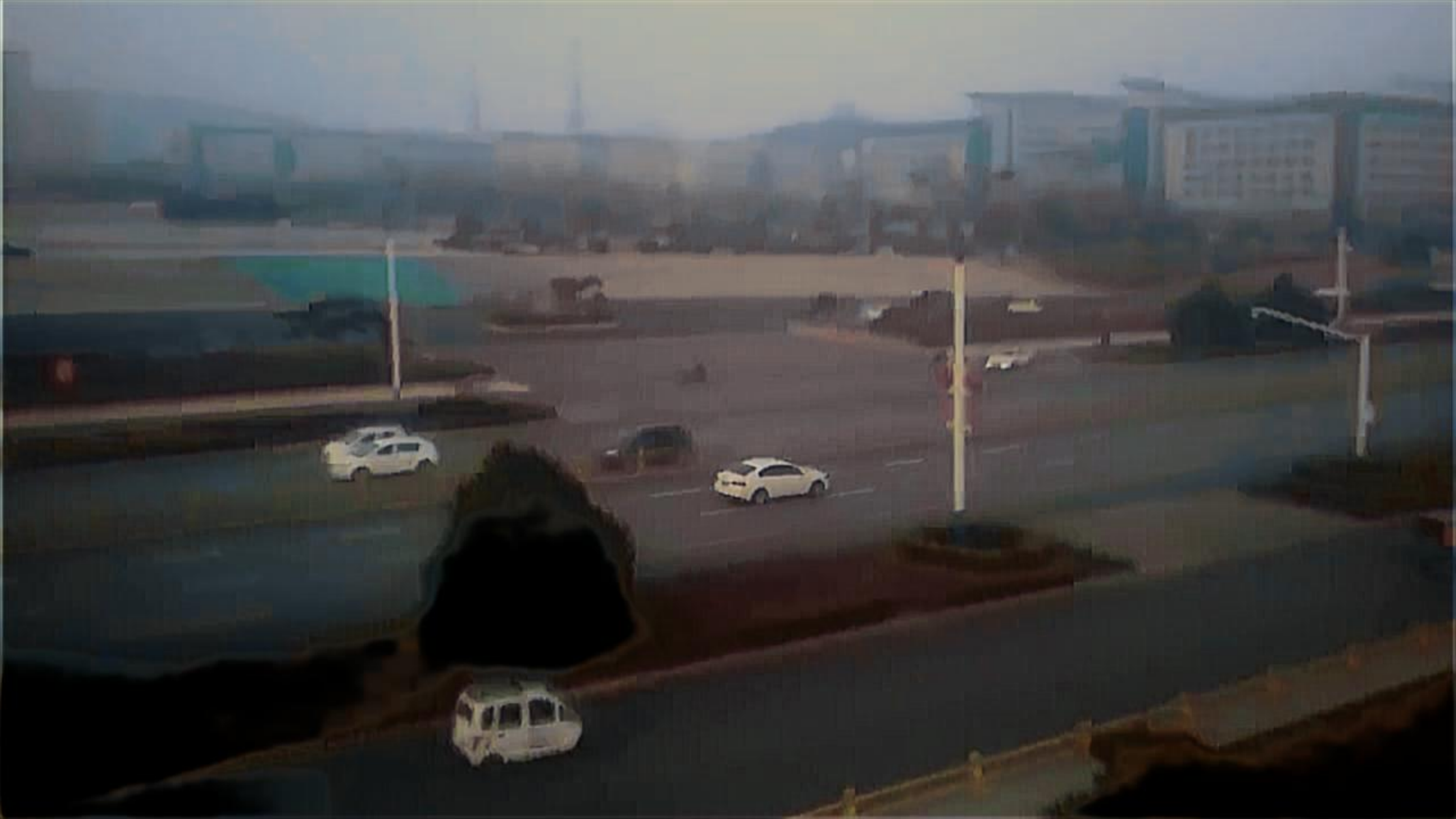}         &
\includegraphics[width = 0.16\textwidth]{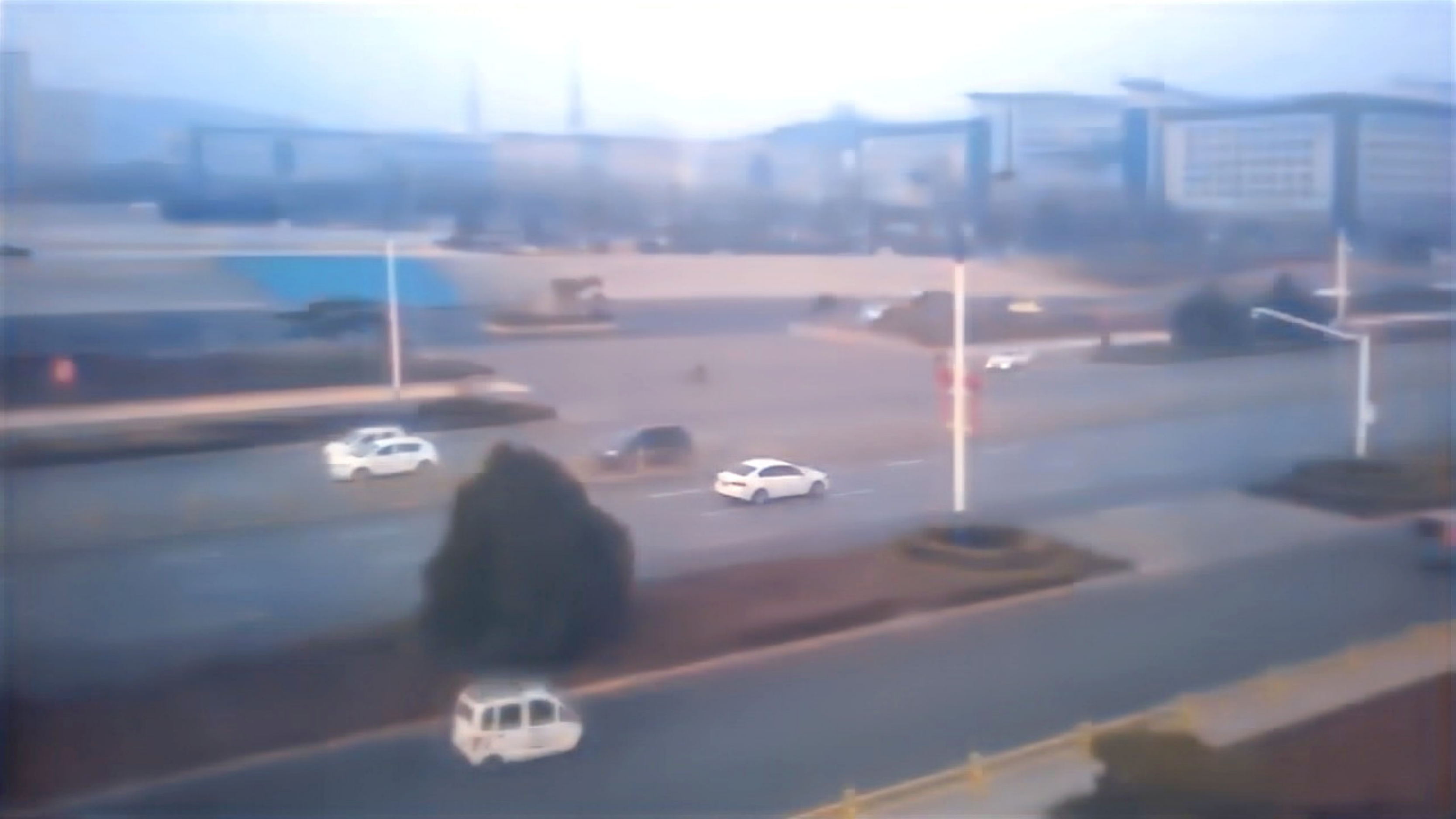}         \\

				\includegraphics[width = 0.16\textwidth]{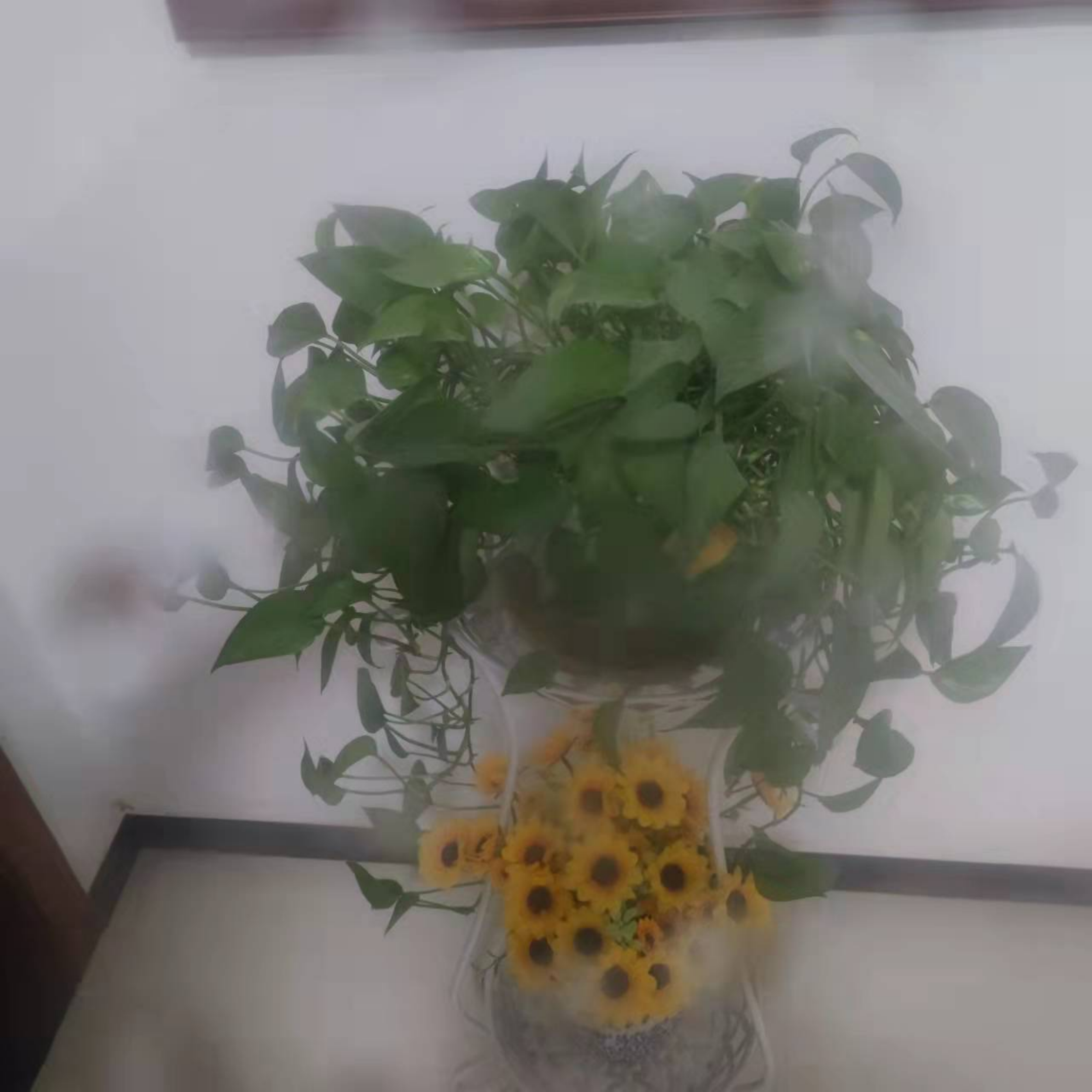}         &
				\includegraphics[width = 0.16\textwidth]{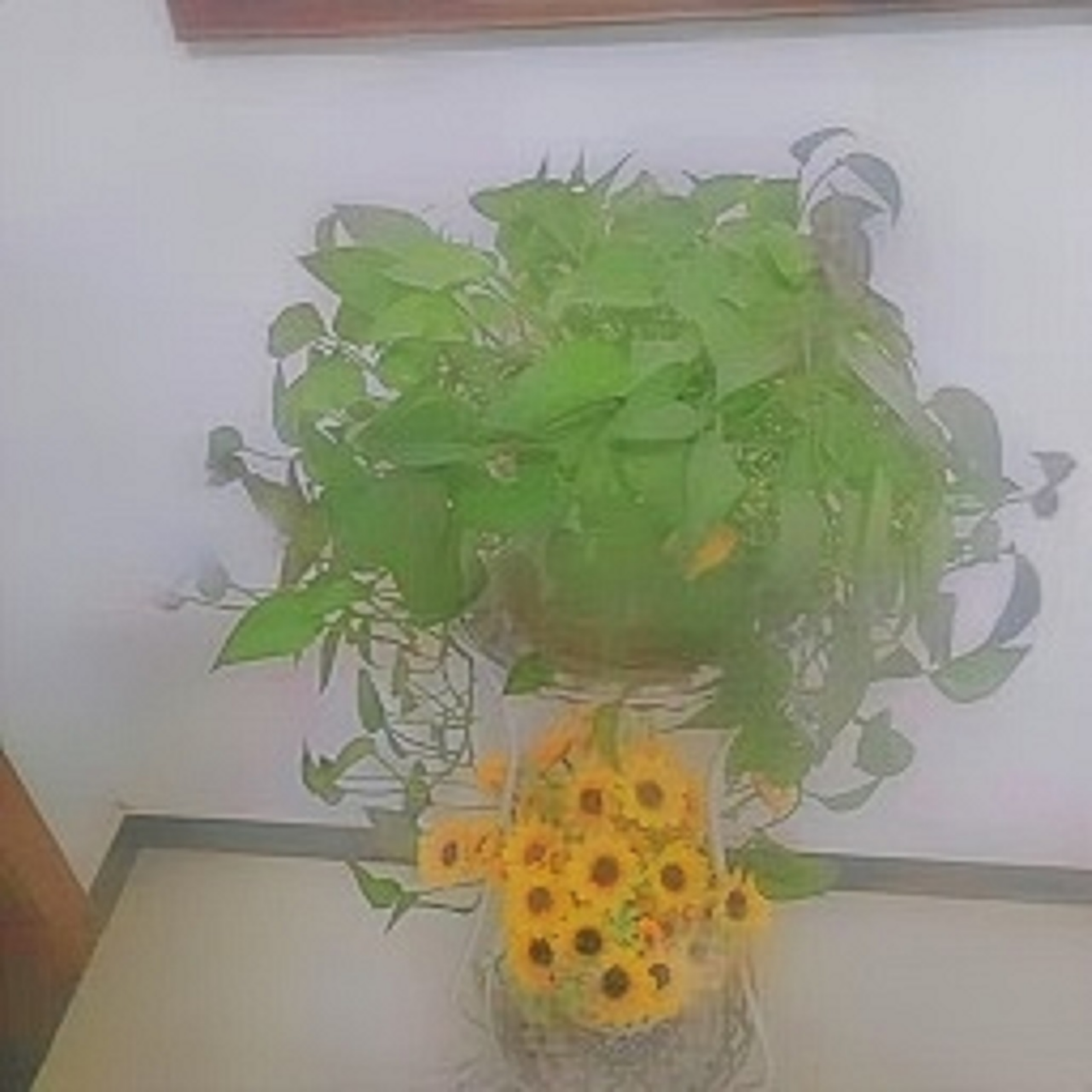}         &
				\includegraphics[width = 0.16\textwidth]{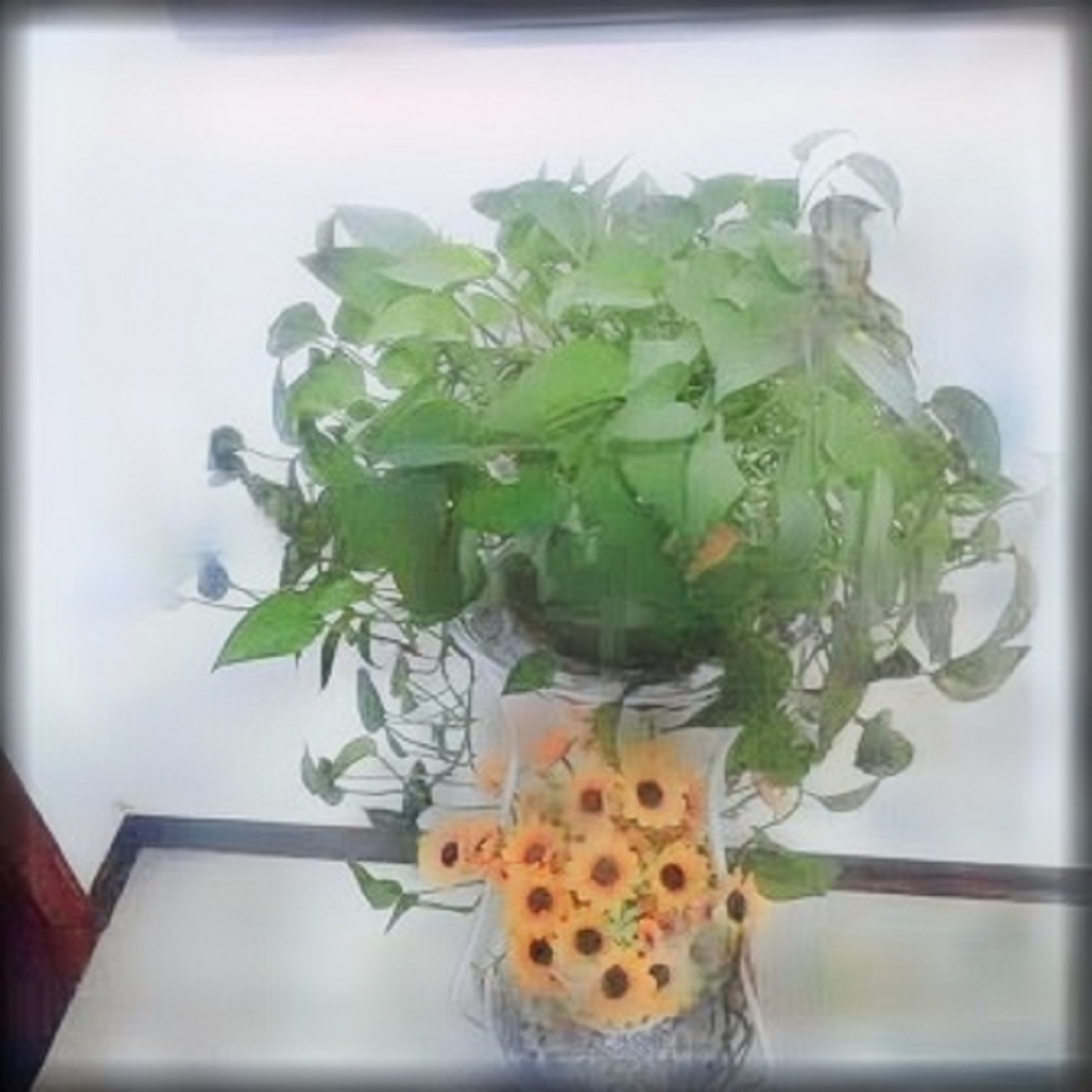}         &
				\includegraphics[width = 0.16\textwidth]{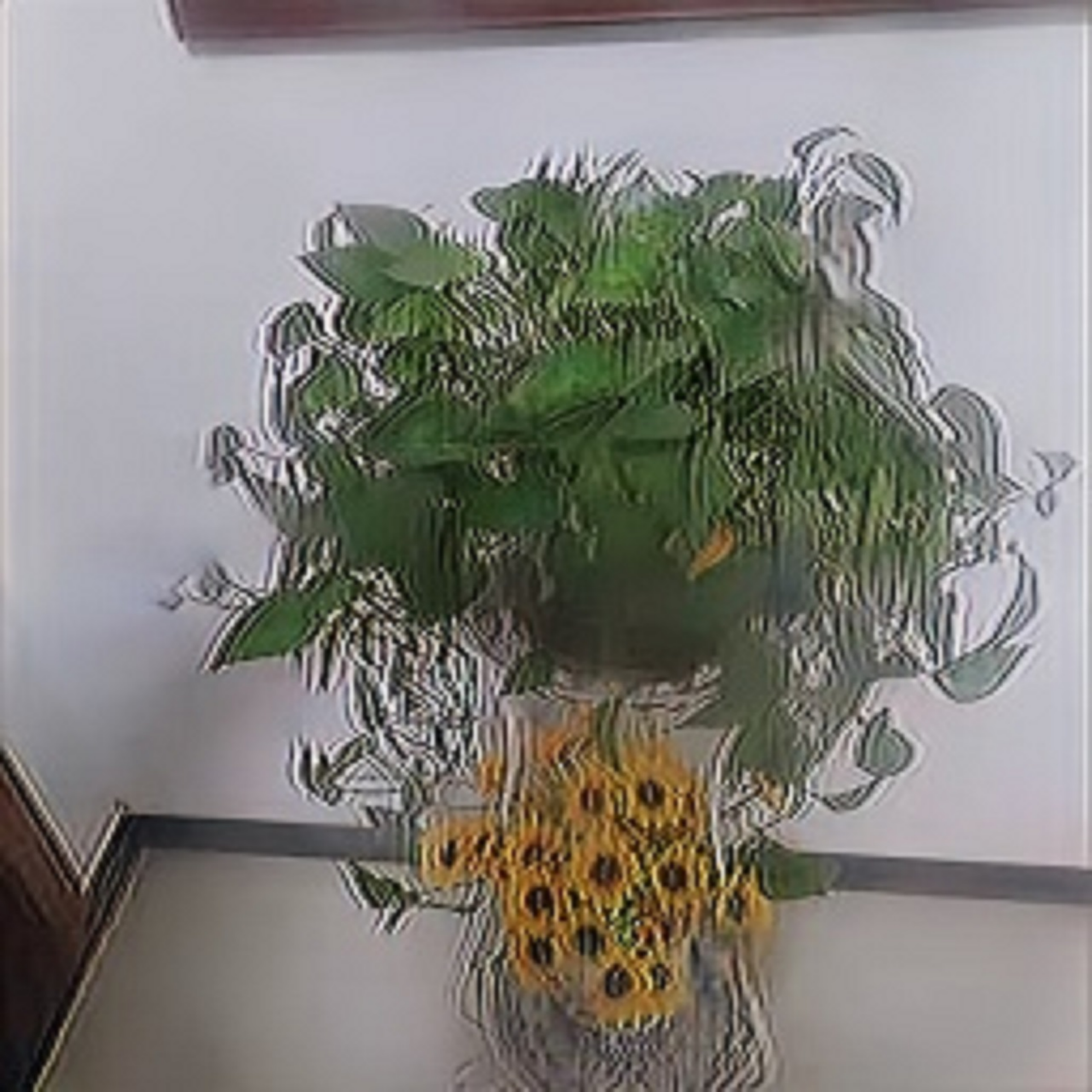}         &
				\includegraphics[width = 0.16\textwidth]{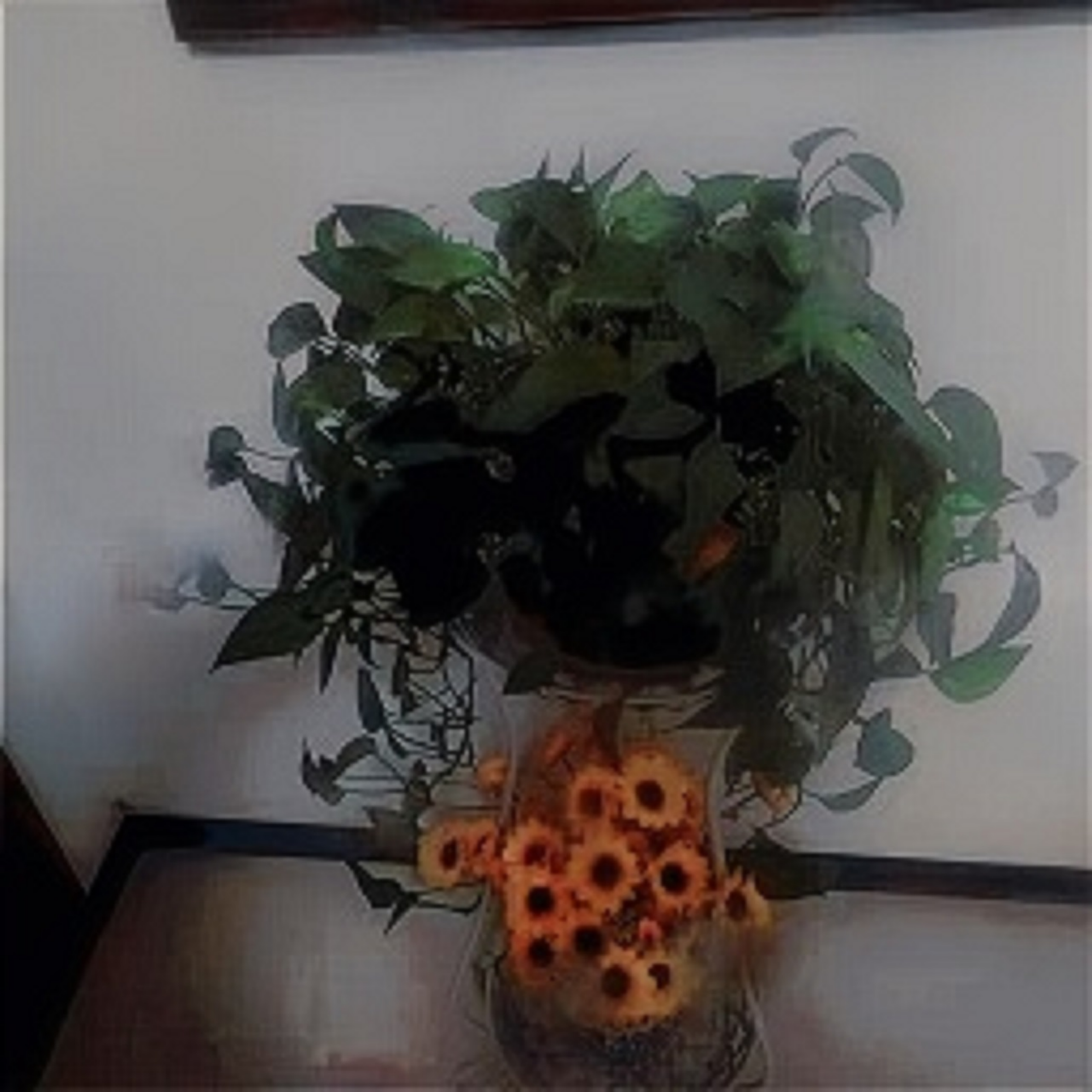}         &
				\includegraphics[width = 0.16\textwidth]{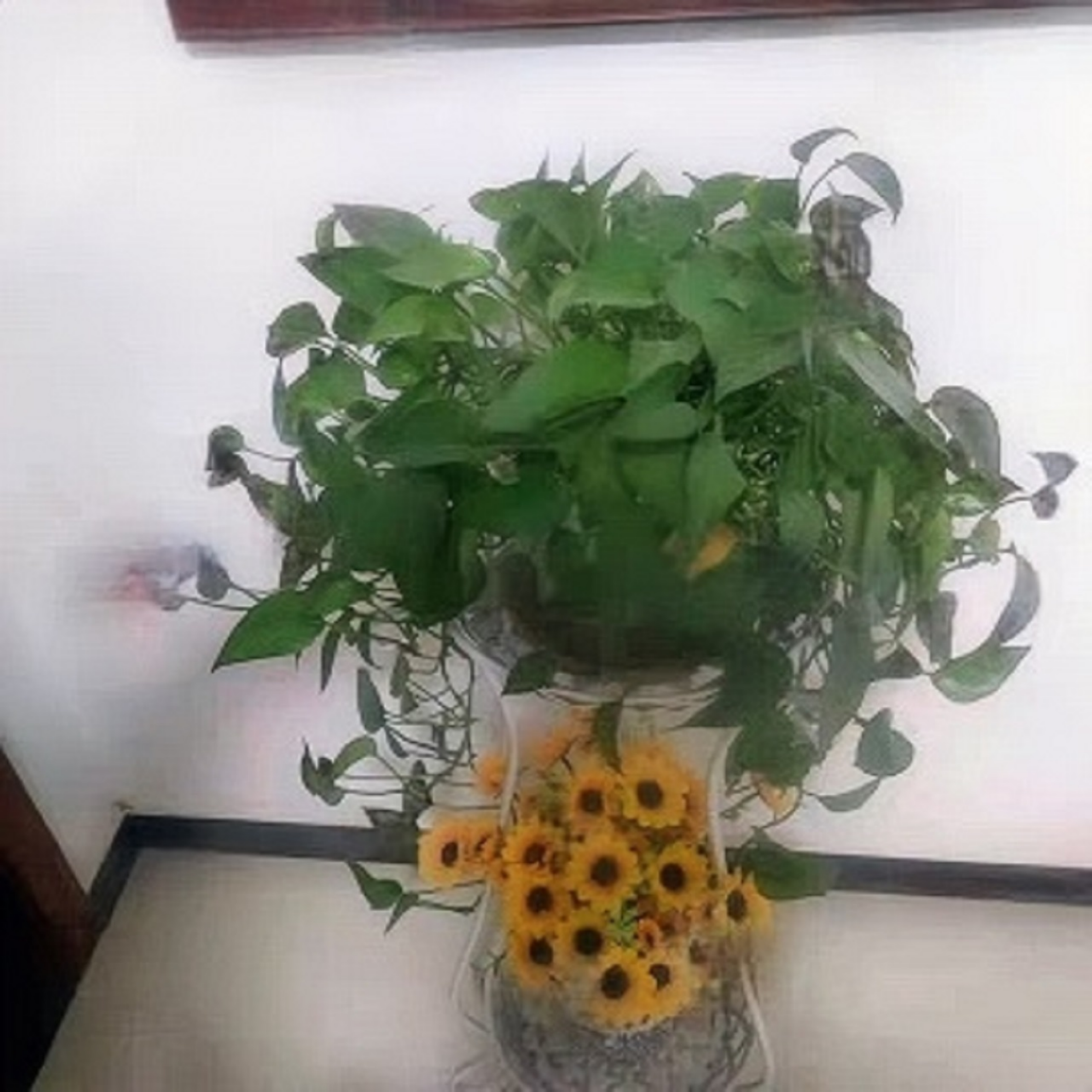}         \\
				
				\includegraphics[width = 0.16\textwidth]{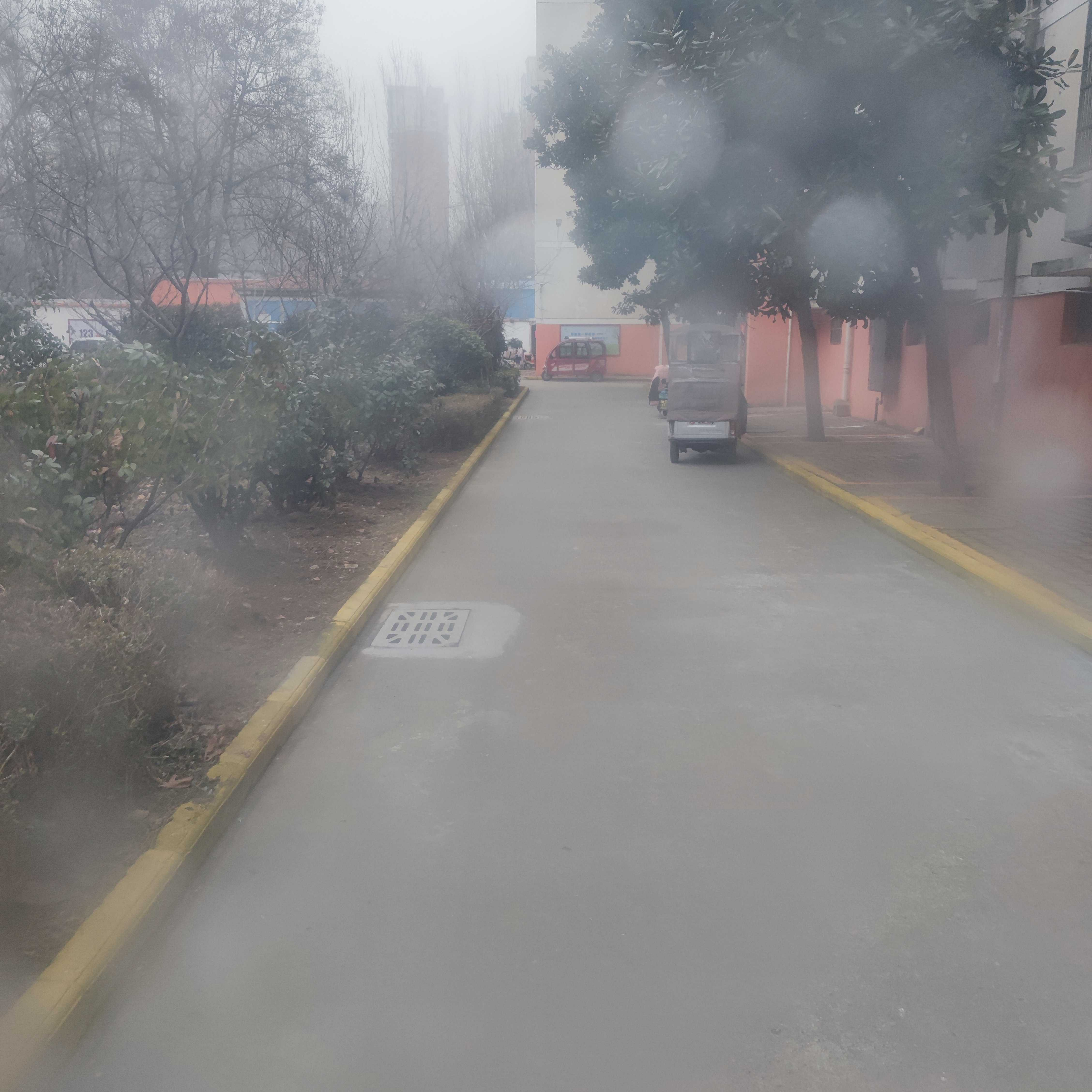}         &
				\includegraphics[width = 0.16\textwidth]{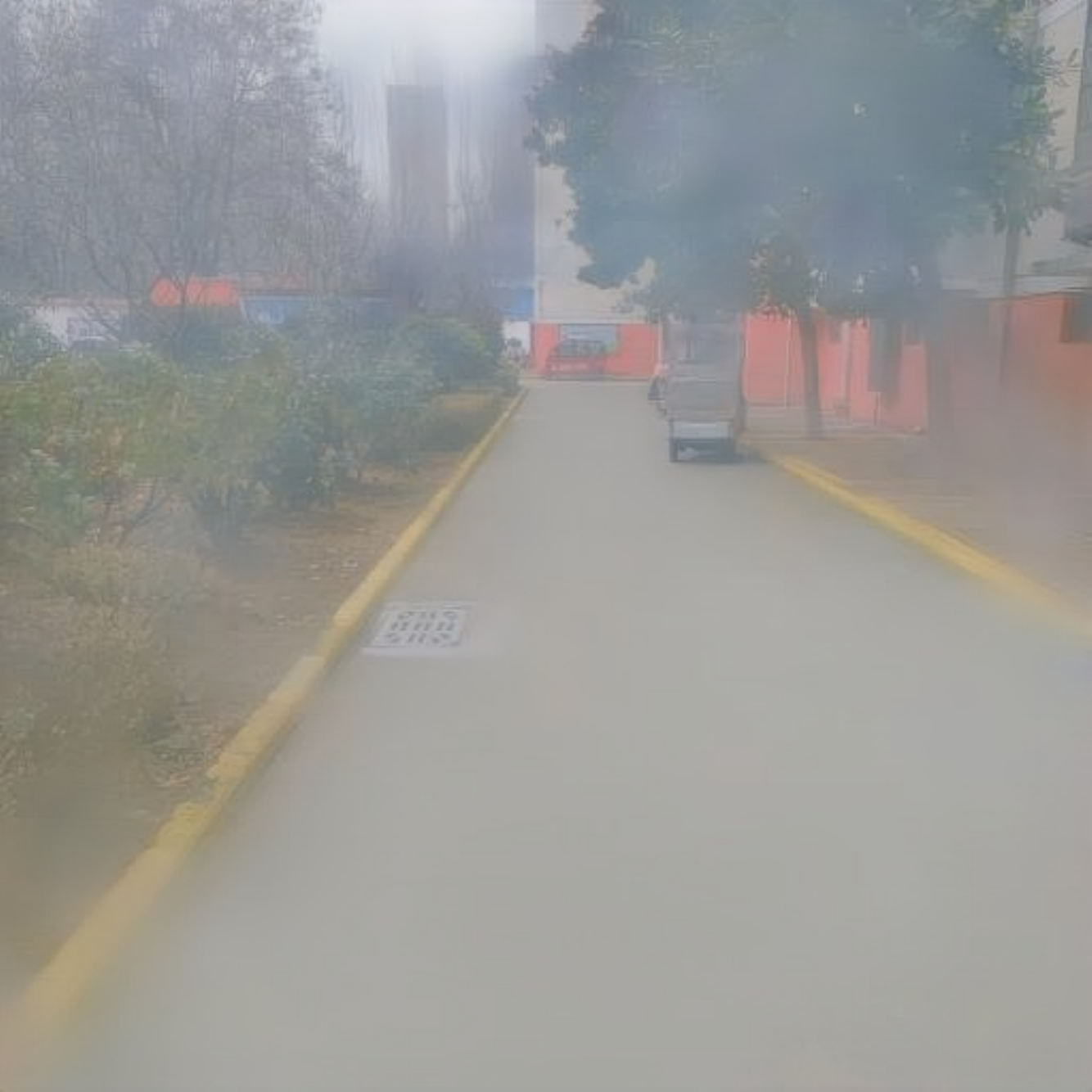}         &
				\includegraphics[width = 0.16\textwidth]{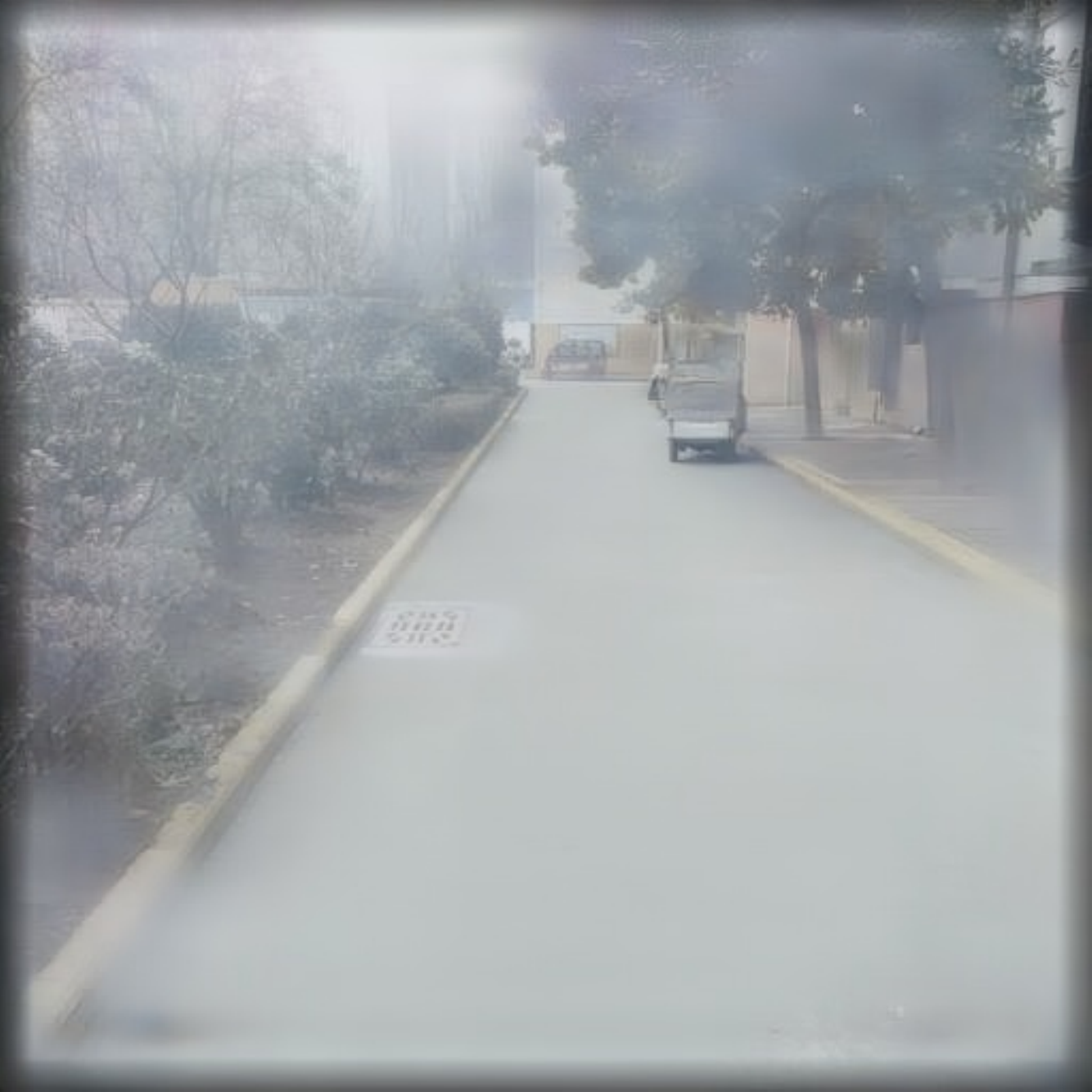}         &
				\includegraphics[width = 0.16\textwidth]{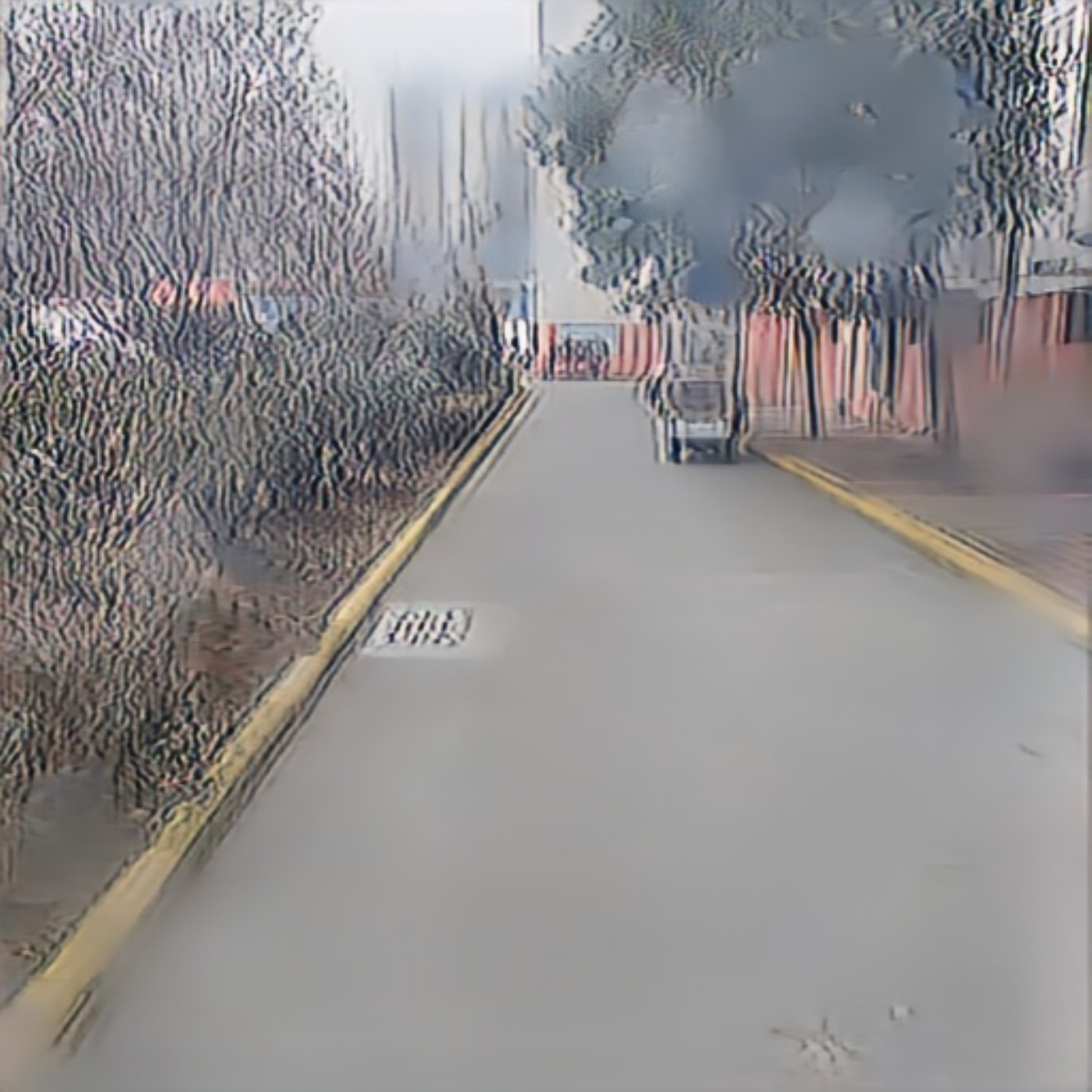}         &
				\includegraphics[width = 0.16\textwidth]{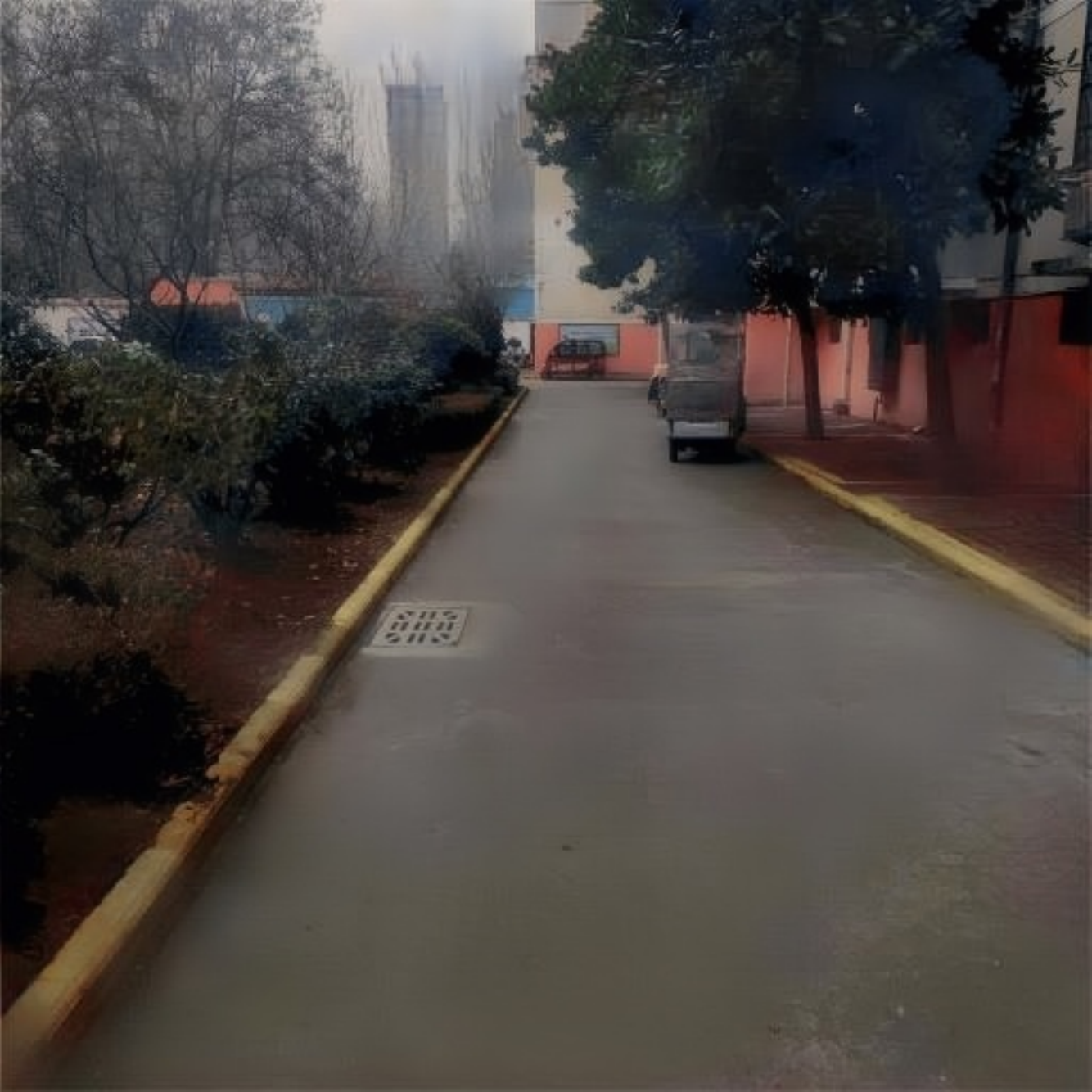}         &
				\includegraphics[width = 0.16\textwidth]{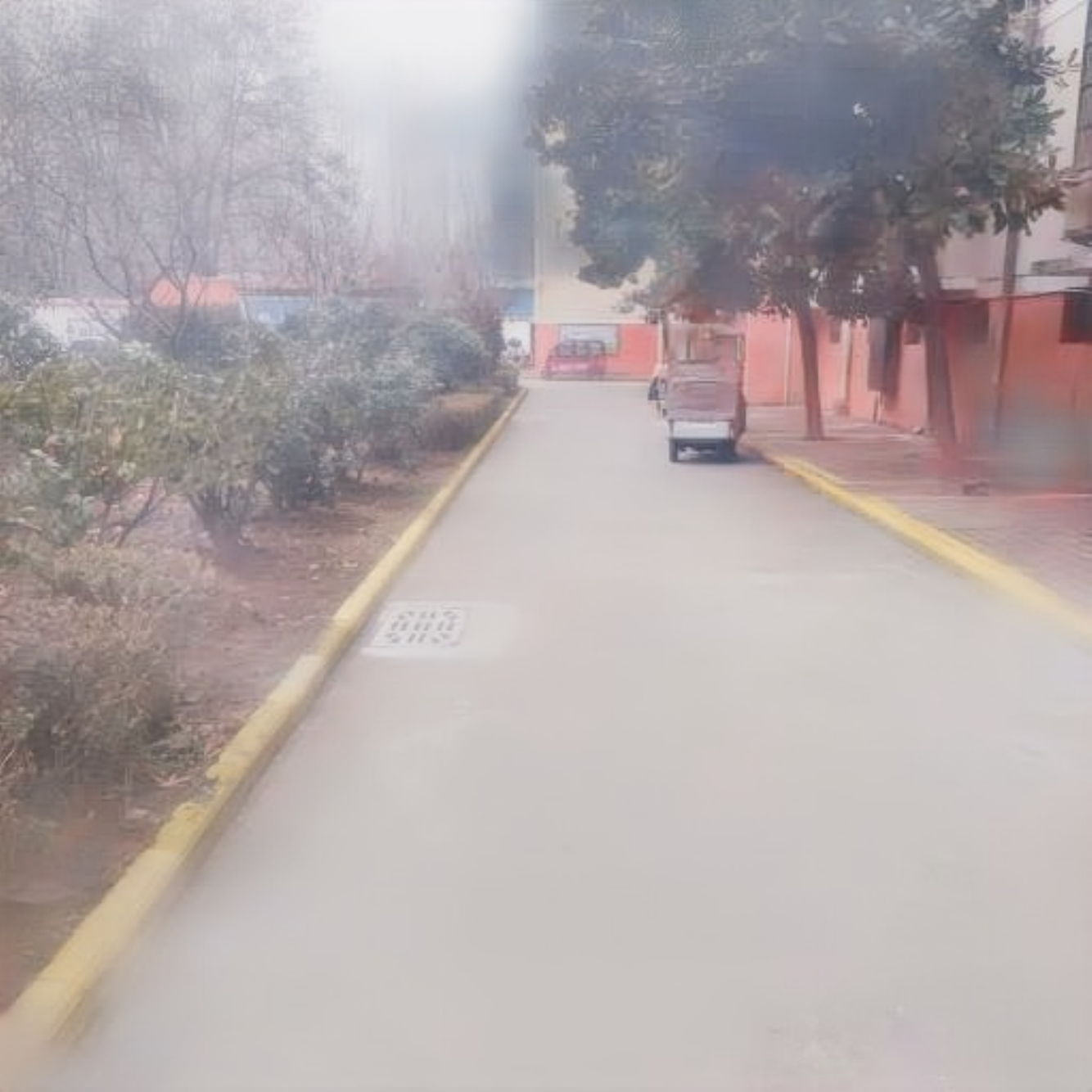}         \\

			(a) Input &
			(b) DCP + Zero-Net    &
			(c) HDRNet &  
			(d) DAGF   & 
			(e) DEUNet  & 
			(f) ZRUDC-Net    \\ 
			
			 &
			\textcolor{blue}{CVPR'09} + \textcolor{blue}{CVPR'20}    &
			\textcolor{blue}{TOG'17} &  
			\textcolor{blue}{ECCV'20} & 
			\textcolor{blue}{CVPR'21}  & 
		    \textcolor{blue}{Ours}
		    \\ 
			
		\end{tabular}
	\end{center}
	\vspace{-2mm}
	\caption{Enhanced results on real captured images on UAV with T-OLED. The proposed method generates much clear images as well as vivid colors. The images in the last two rows simulate the T-OLED being adhered by raindrops.
	Notably, we have used three resolutions for the images, $1280 \times 720$ resolution for the head two rows, $2180 \times 2180$ resolution for the third row, and $4180 \times 4180$ resolution for the last row. Except for HDRNet, Zero-Net and our approach, all other models use the downsampling-enhancement-upsampling method to process UHD resolution images on resource-constrained GPUs (24G RAM).	
}
	\vspace{-5mm}
	\label{fig-Real_mydata}
\end{figure*}

\begin{figure*}[t]\scriptsize
	\begin{center}
		\tabcolsep 1pt
		\begin{tabular}{@{}ccccccc@{}}
			
			\includegraphics[width = 0.16\textwidth]{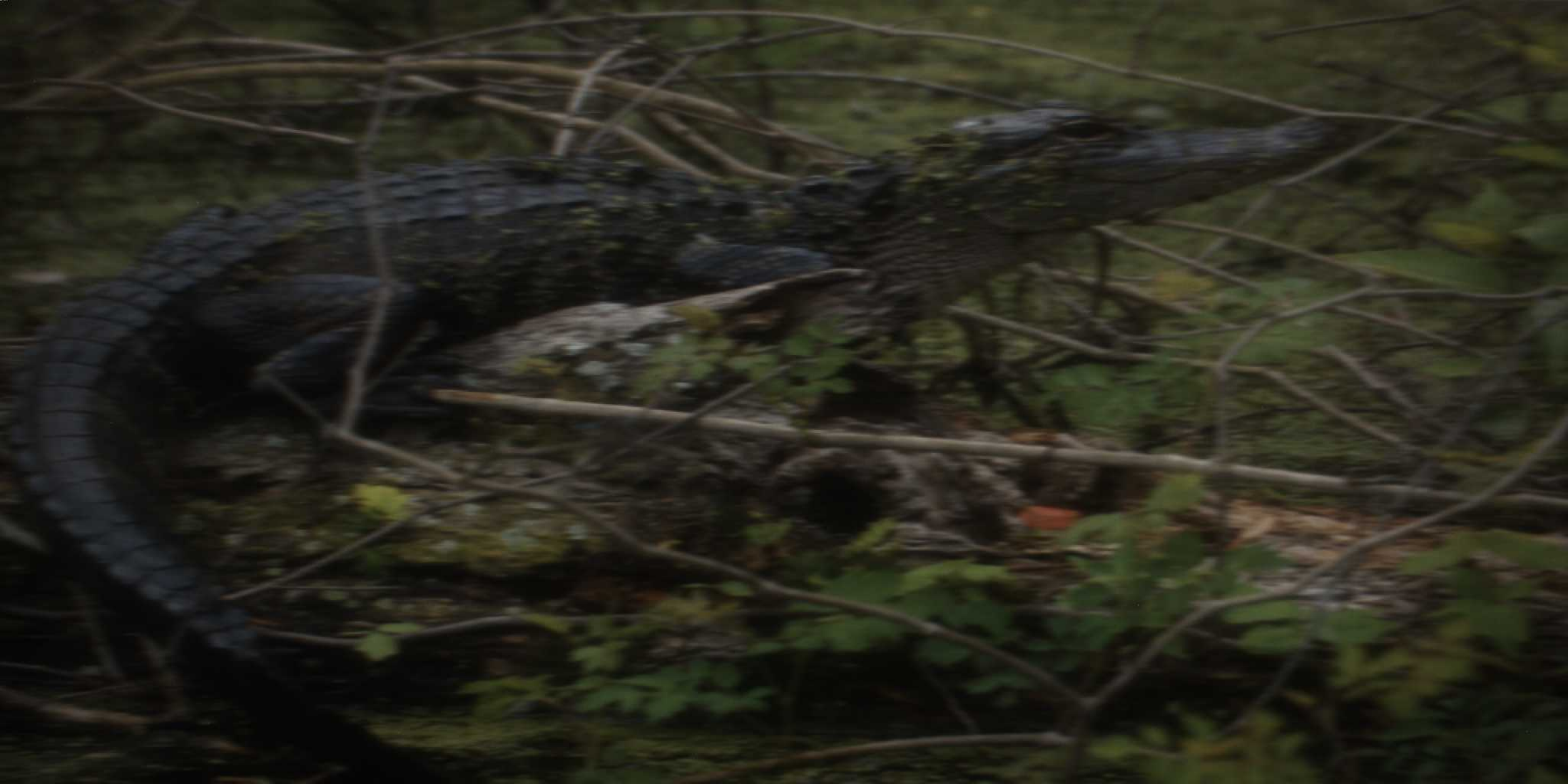}         &
			\includegraphics[width = 0.16\textwidth]{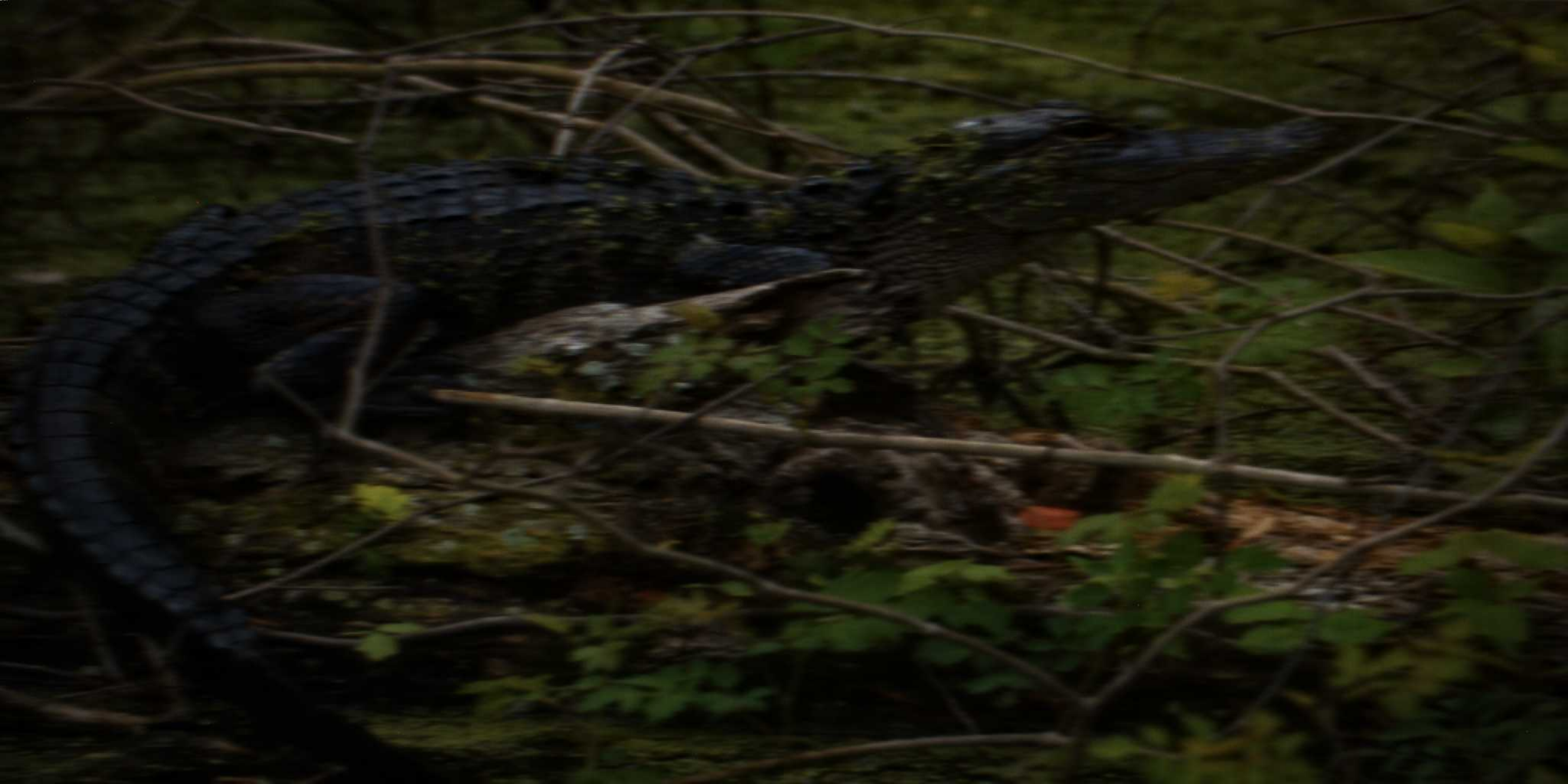}         &
			\includegraphics[width = 0.16\textwidth]{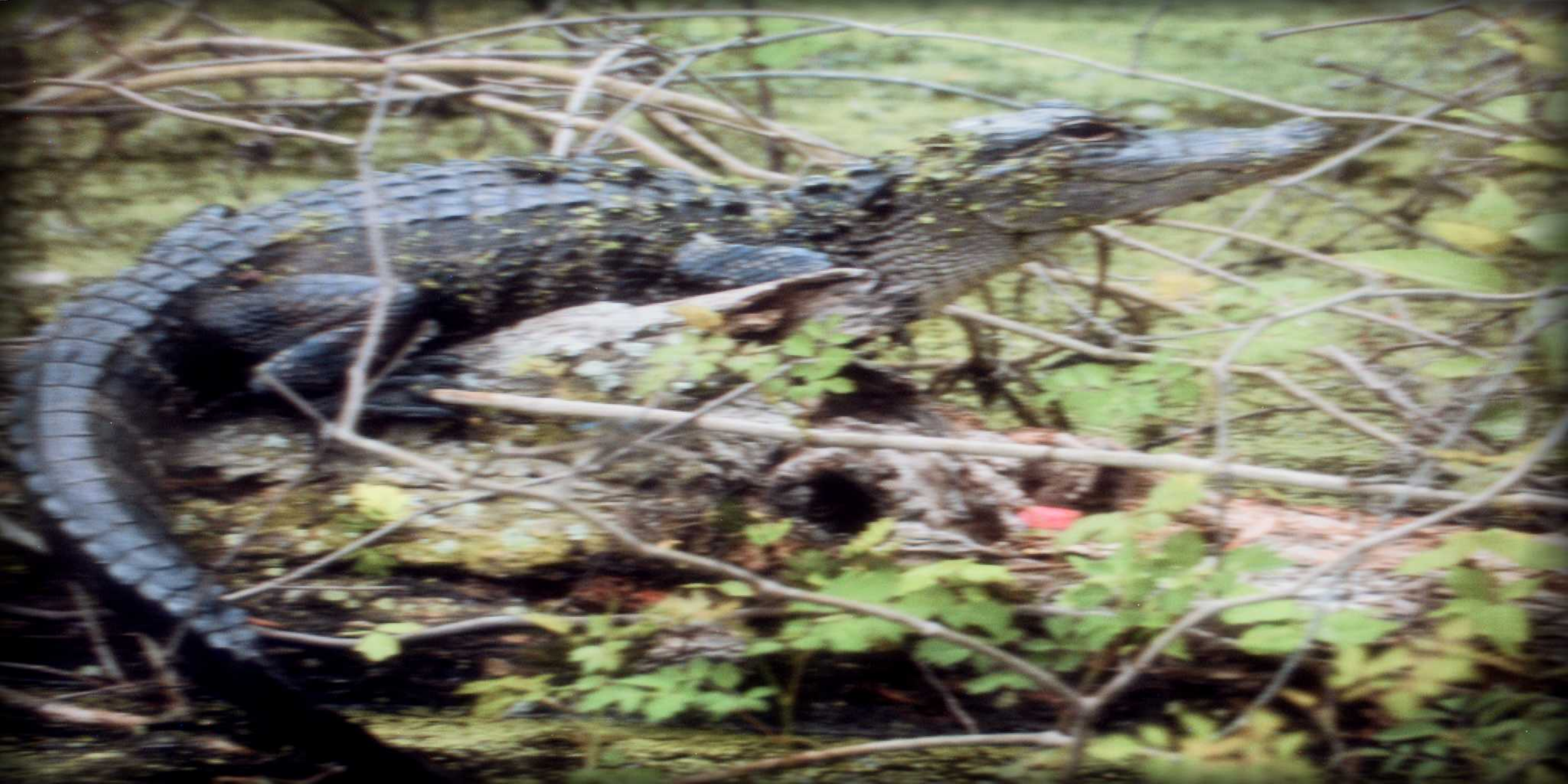}         &
			\includegraphics[width = 0.16\textwidth]{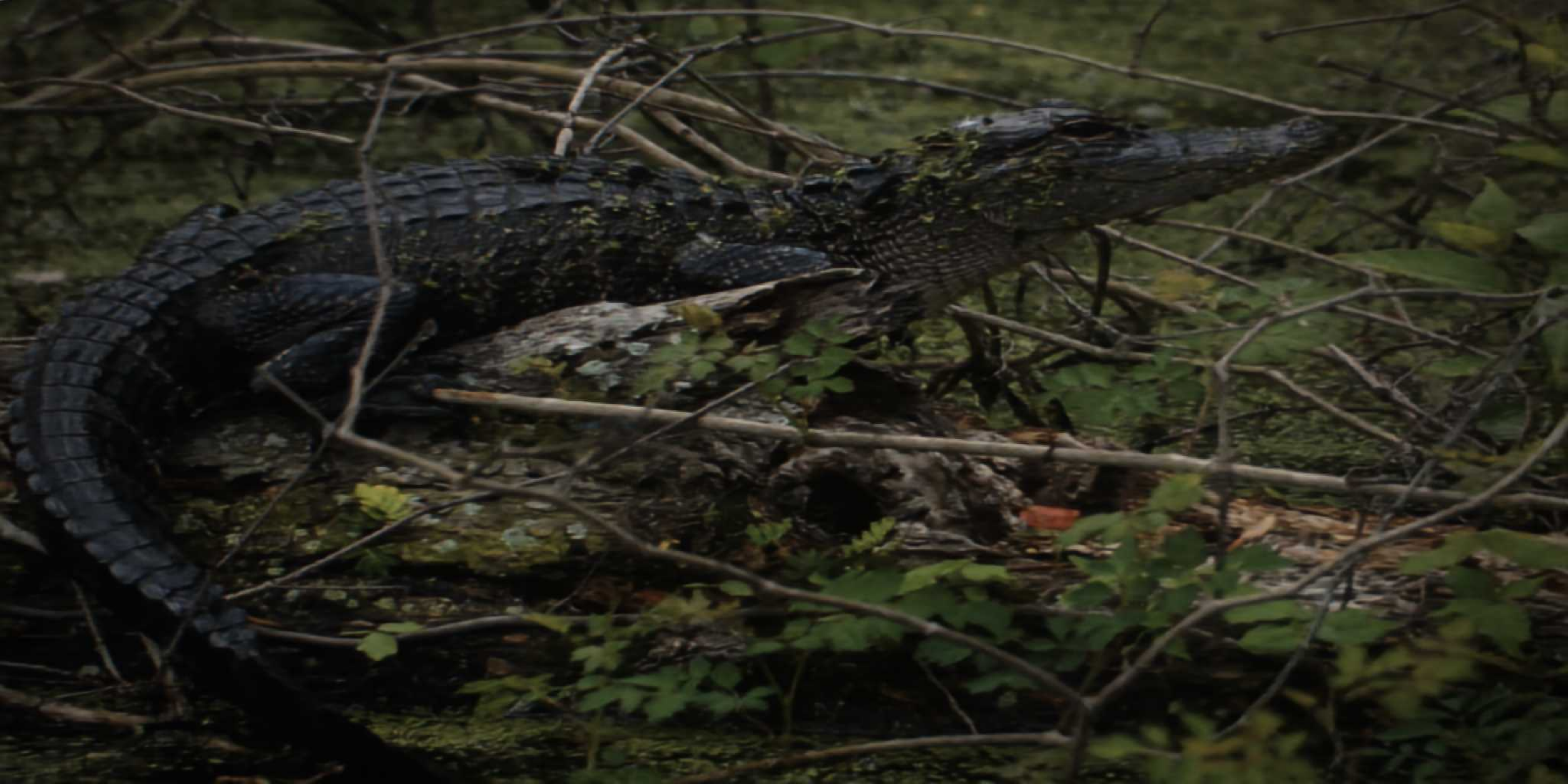}         &
			\includegraphics[width = 0.16\textwidth]{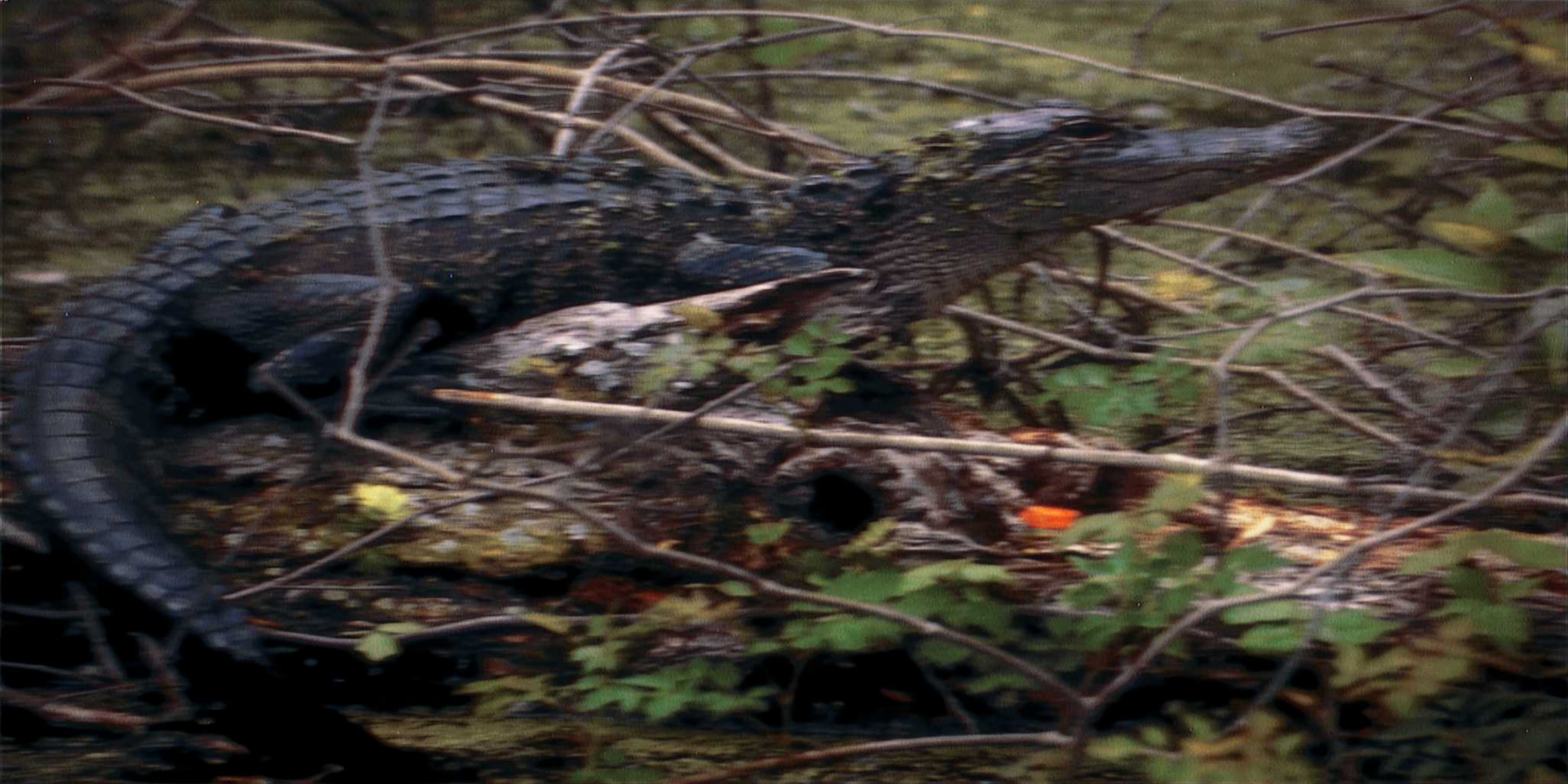}         &
			\includegraphics[width = 0.16\textwidth]{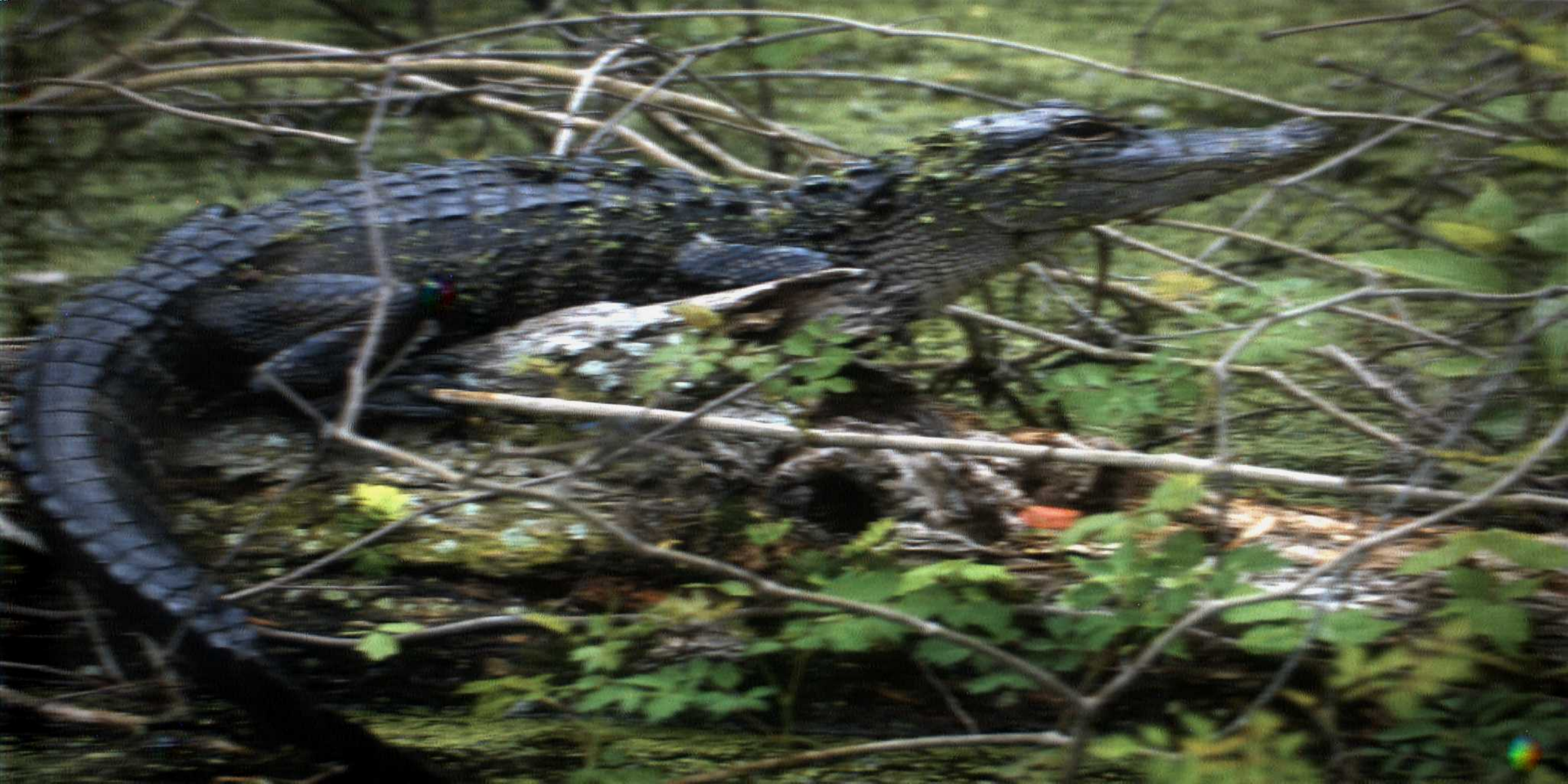}         \\
			
			\includegraphics[width = 0.16\textwidth]{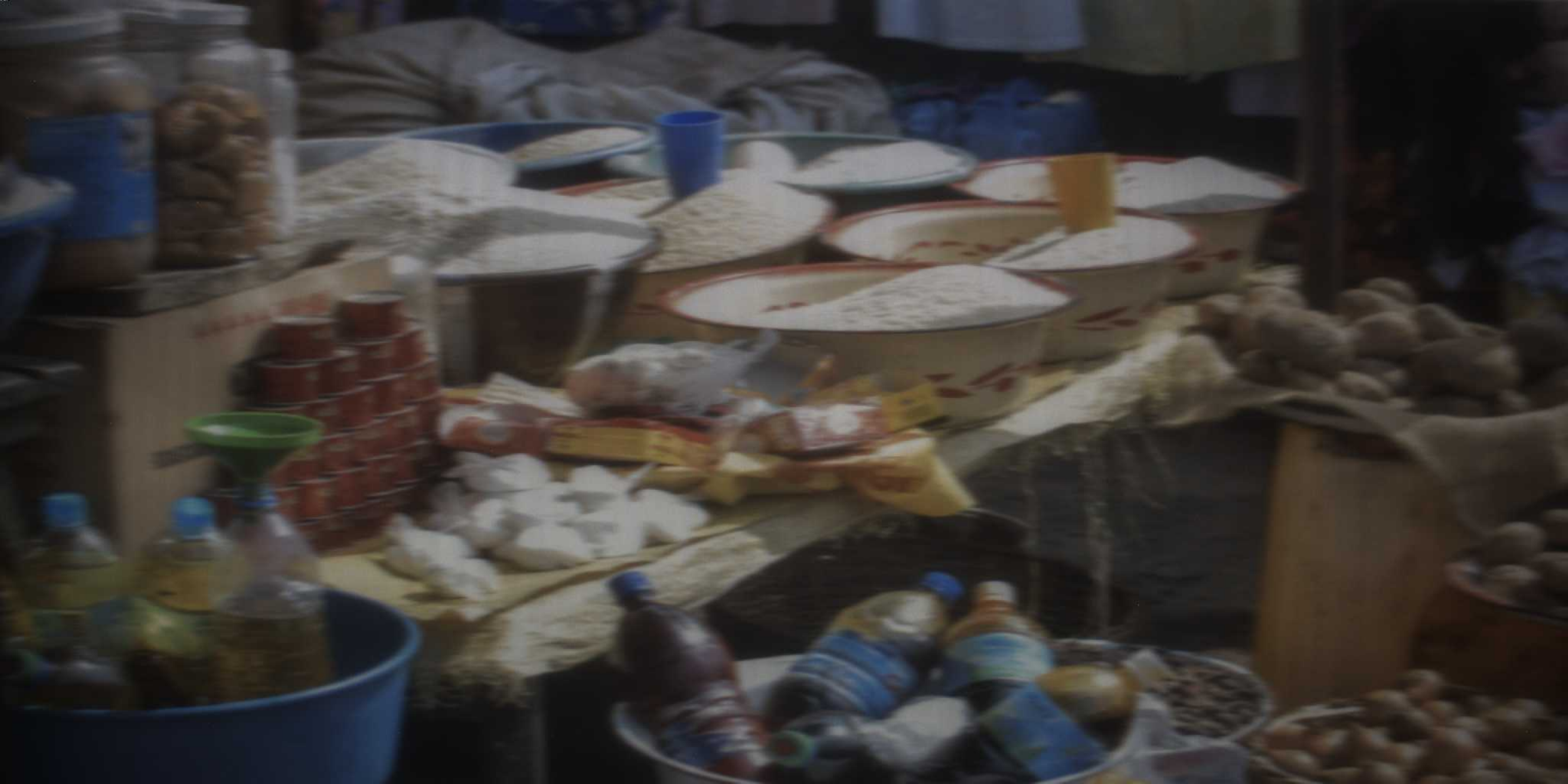}         &
			\includegraphics[width = 0.16\textwidth]{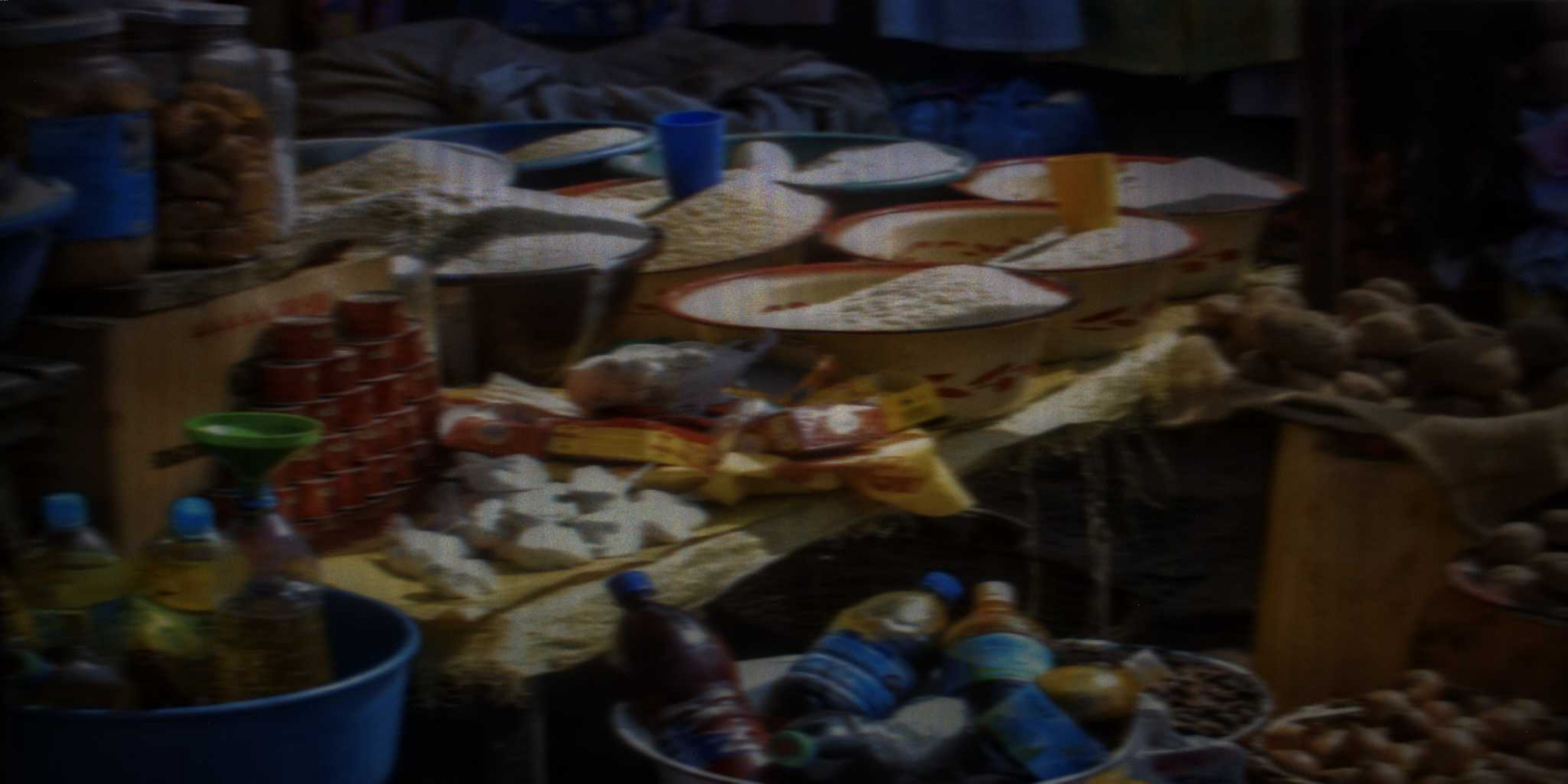}         &
			\includegraphics[width = 0.16\textwidth]{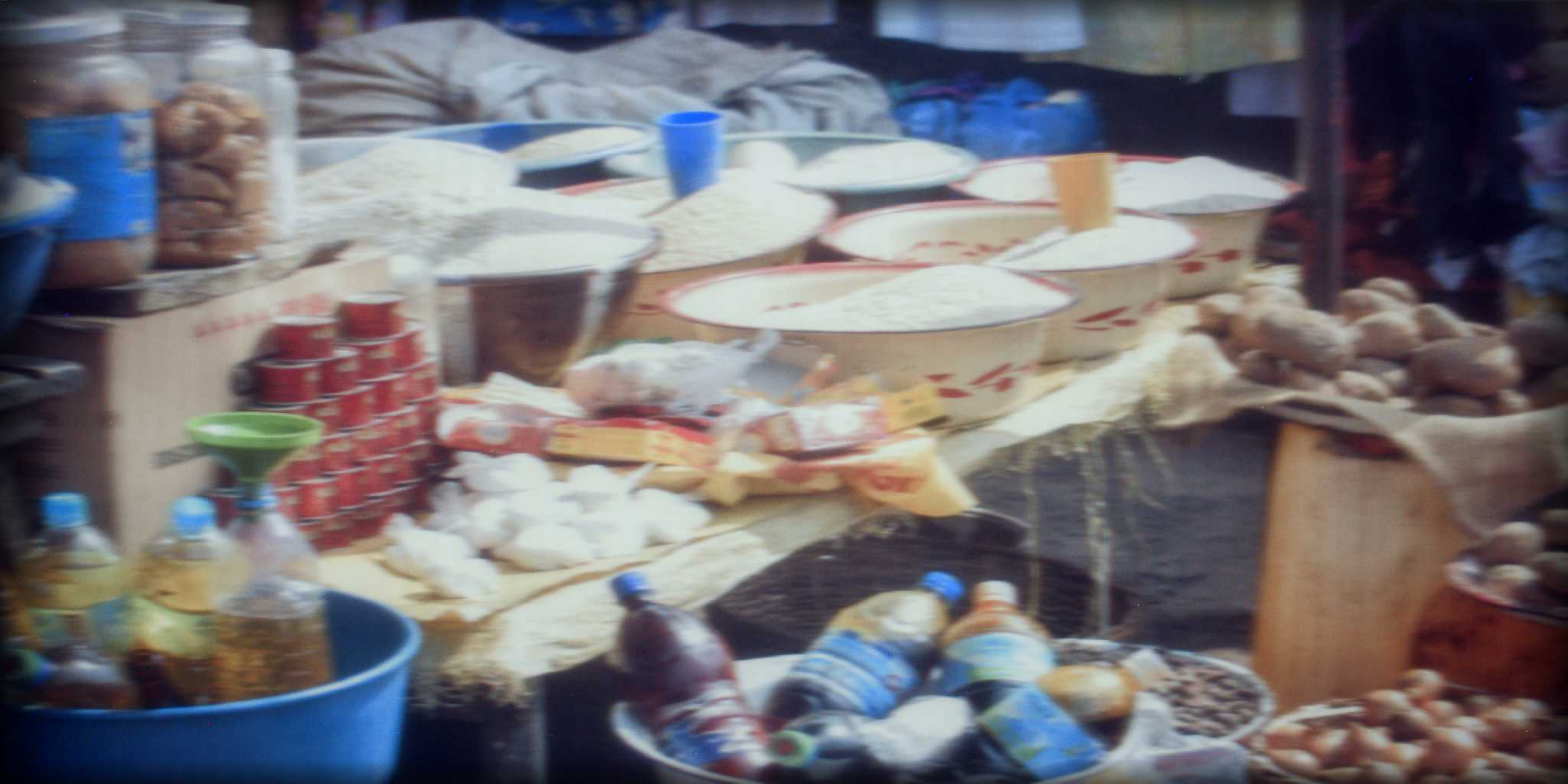}         &
			\includegraphics[width = 0.16\textwidth]{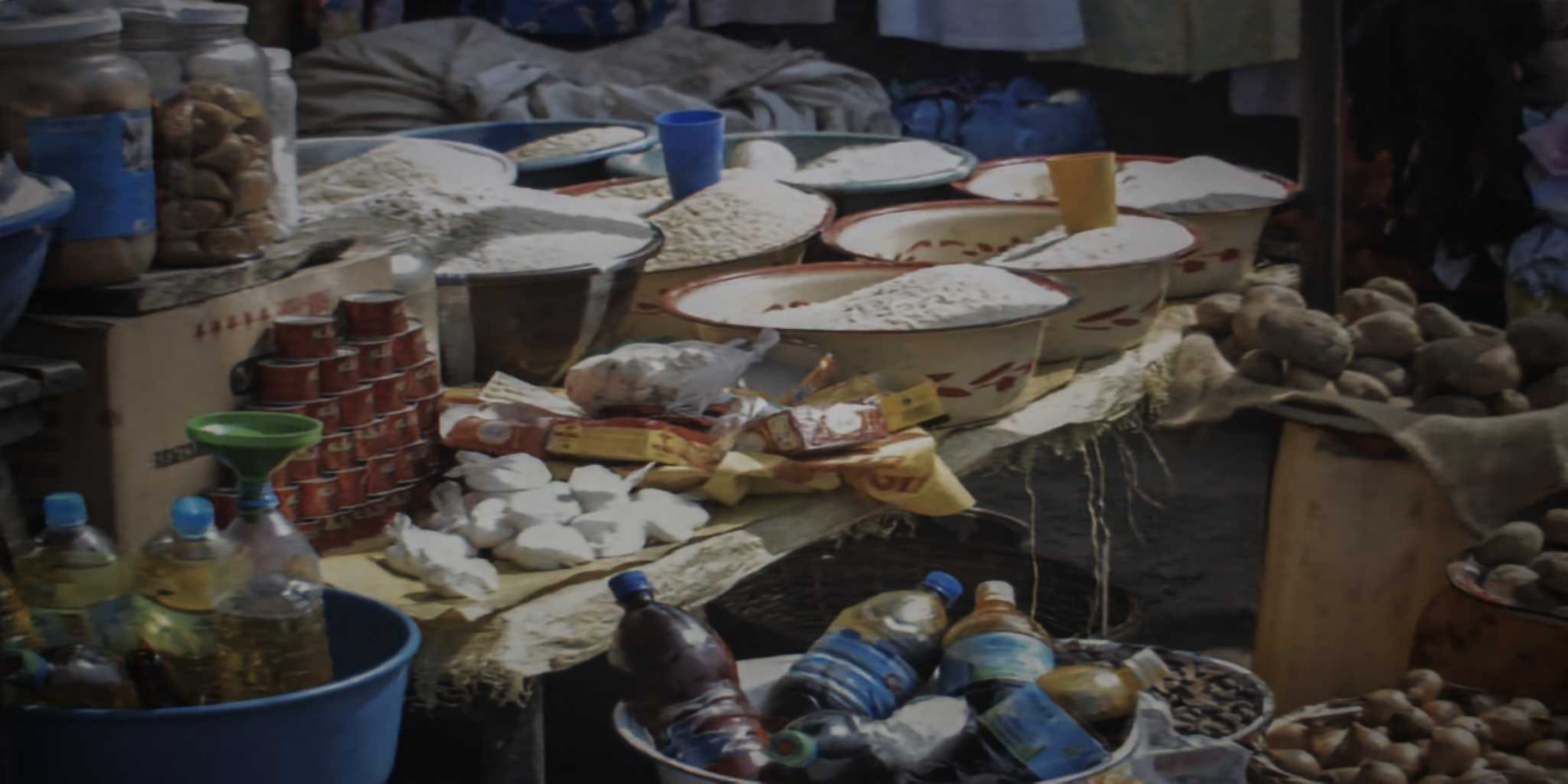}         &
			\includegraphics[width = 0.16\textwidth]{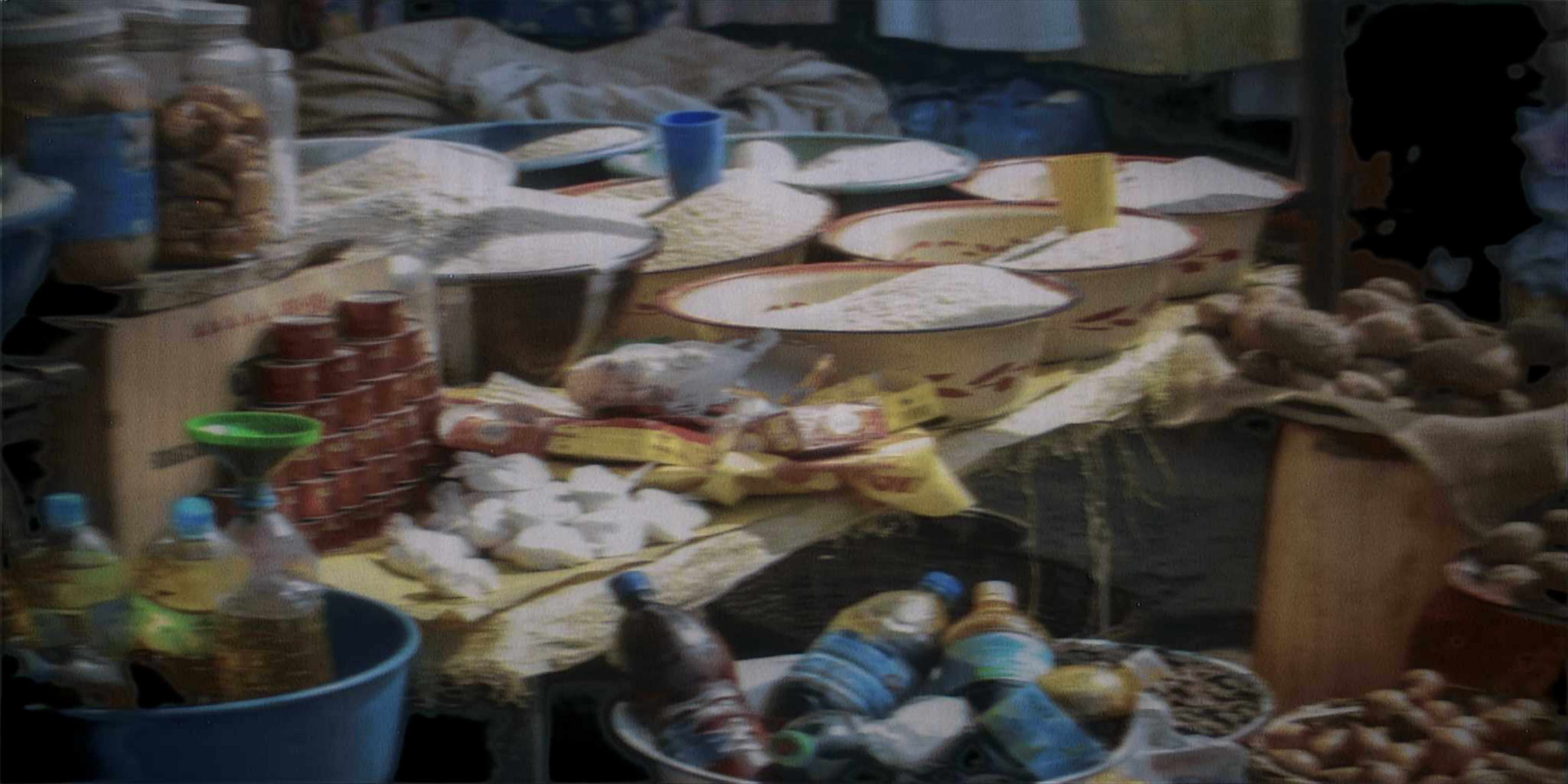}         &
			\includegraphics[width = 0.16\textwidth]{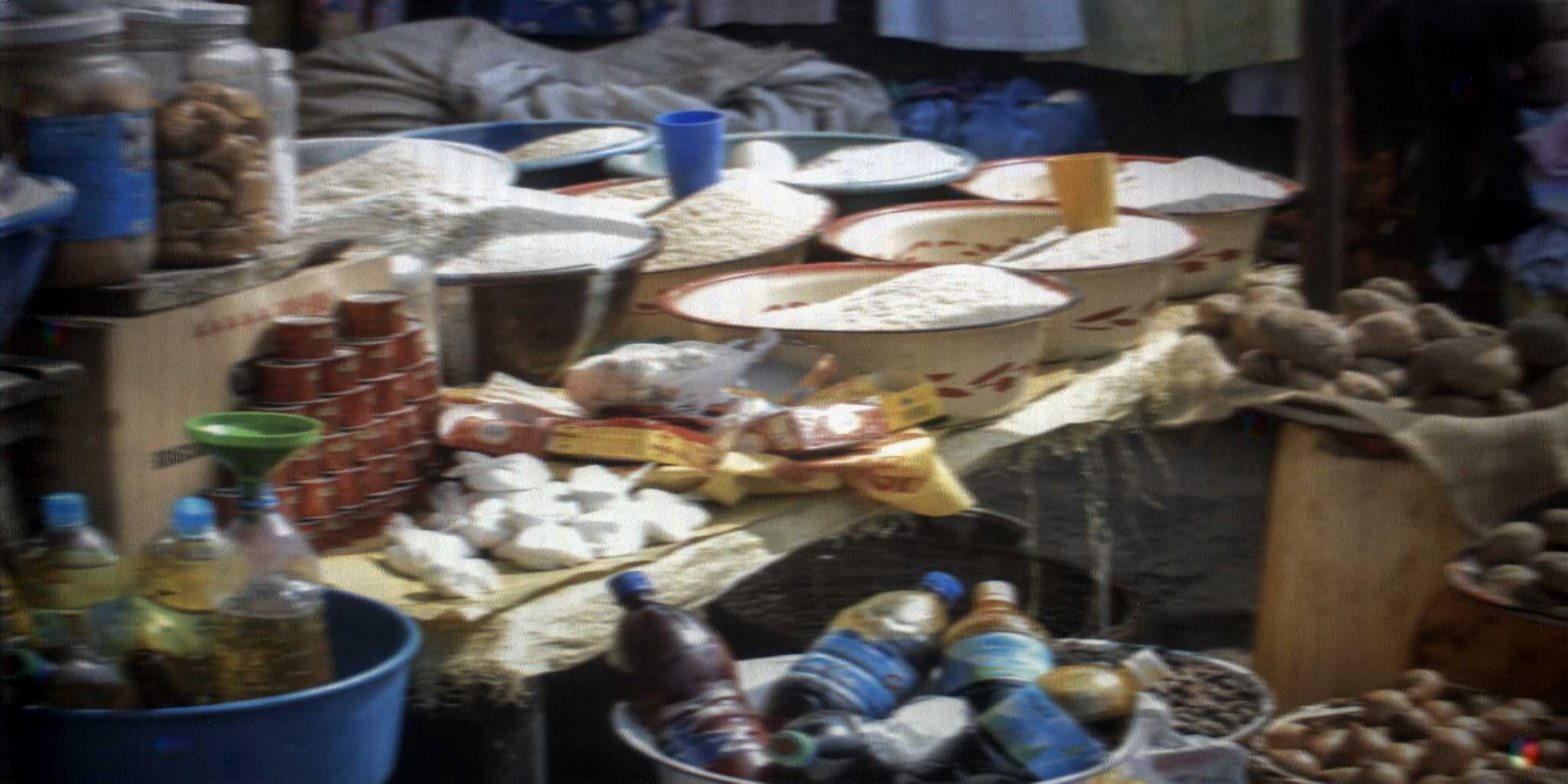}         \\
			
			\includegraphics[width = 0.16\textwidth]{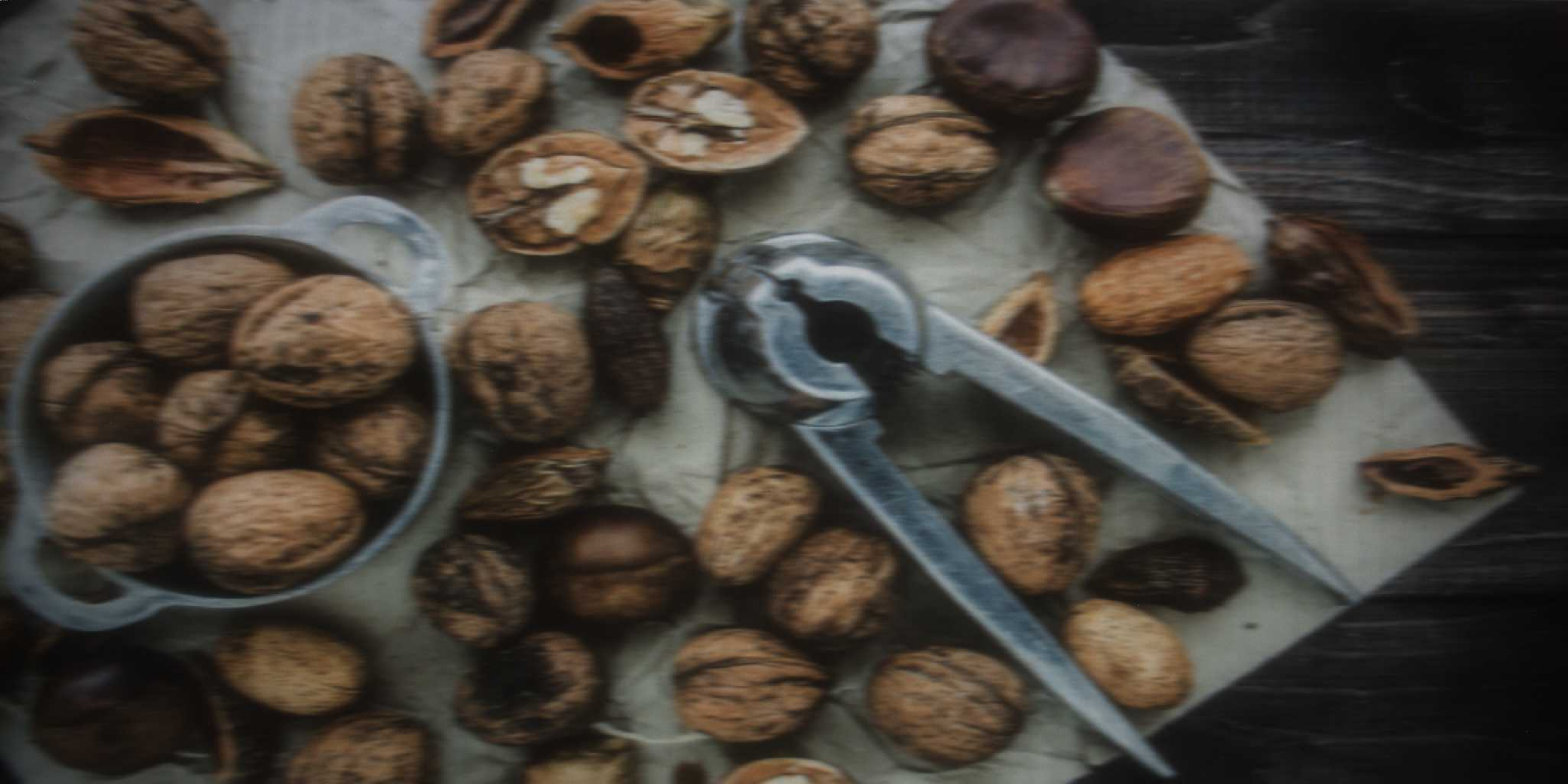}         &
			\includegraphics[width = 0.16\textwidth]{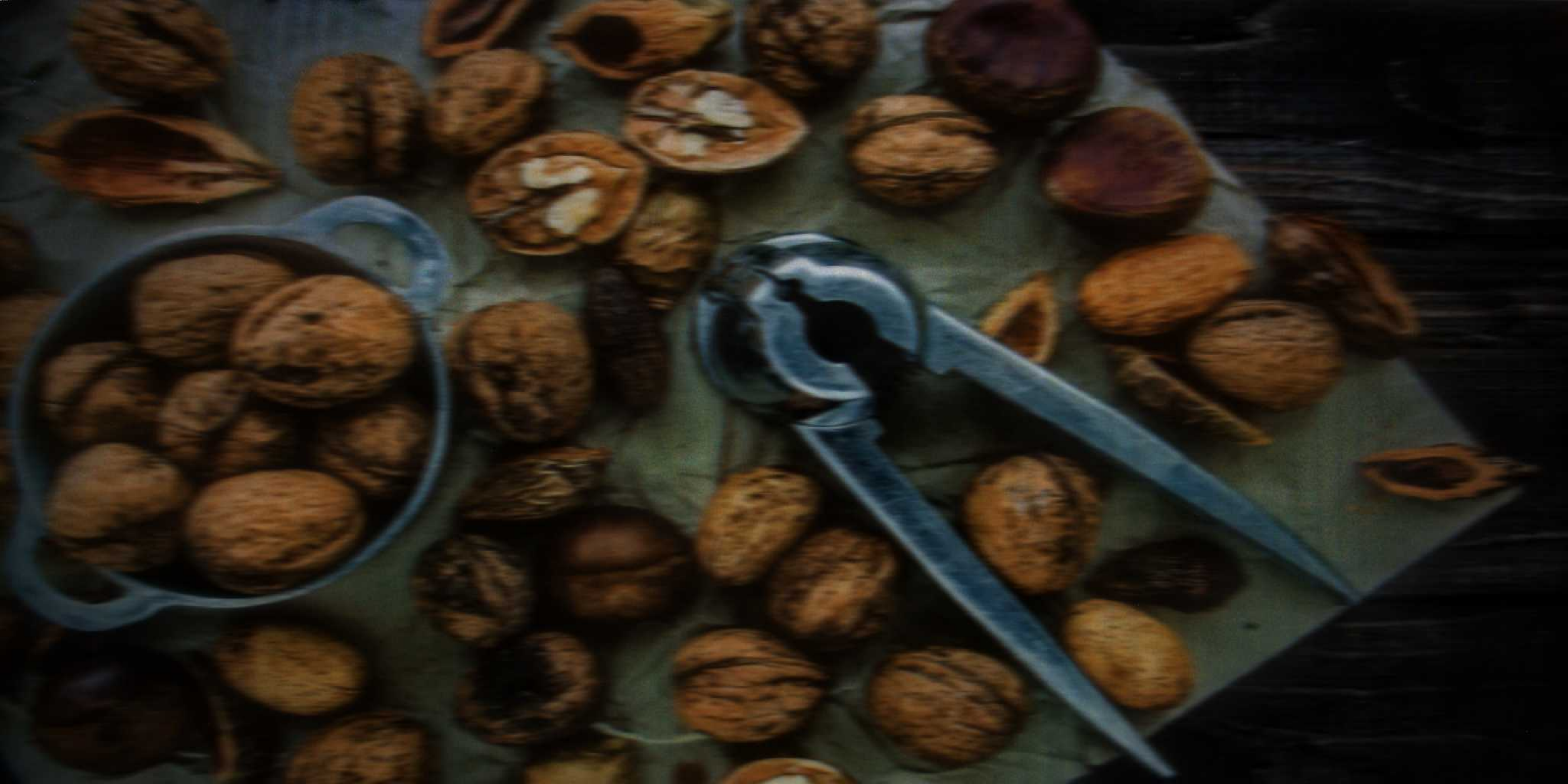}         &
			\includegraphics[width = 0.16\textwidth]{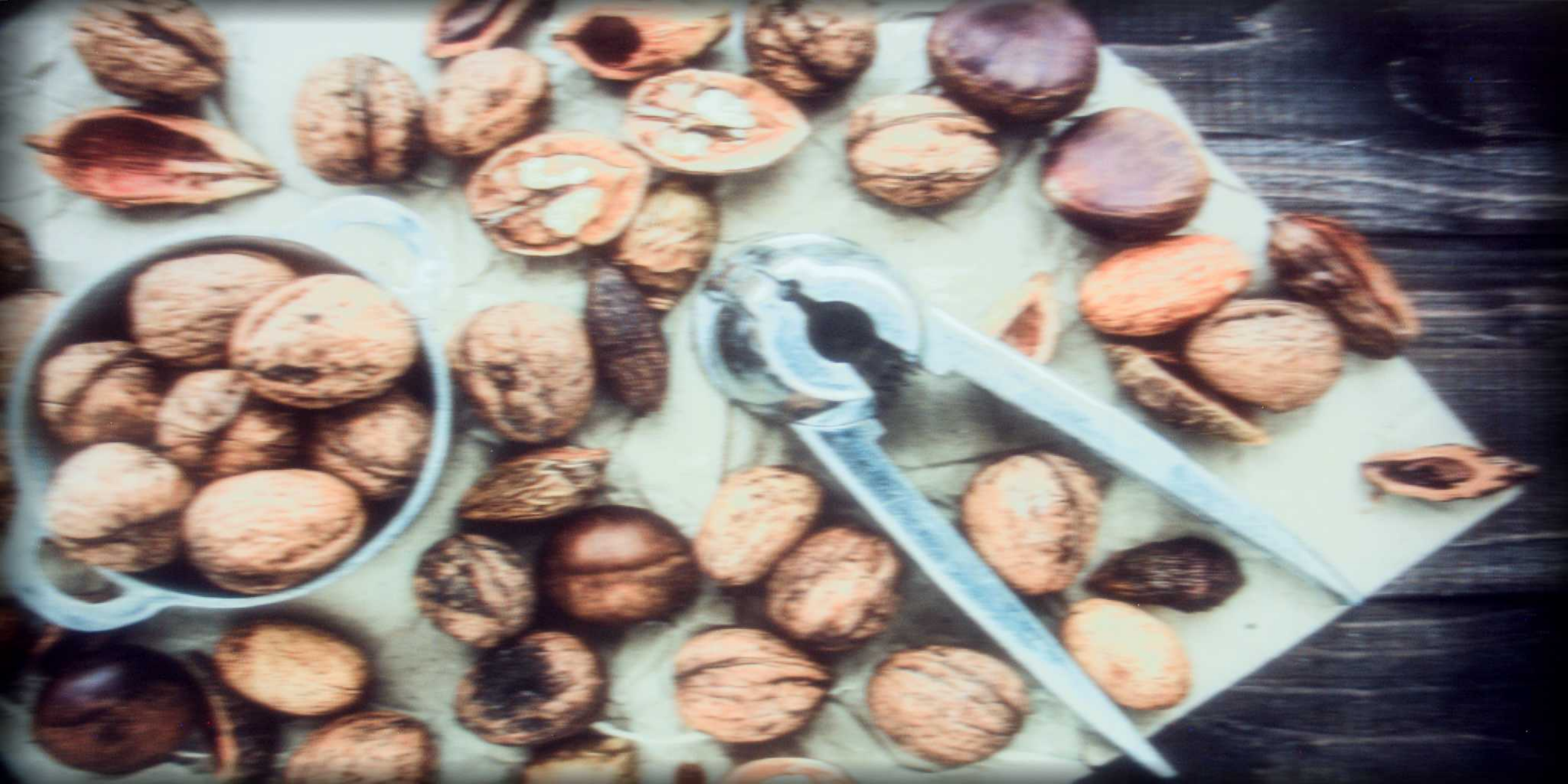}         &
			\includegraphics[width = 0.16\textwidth]{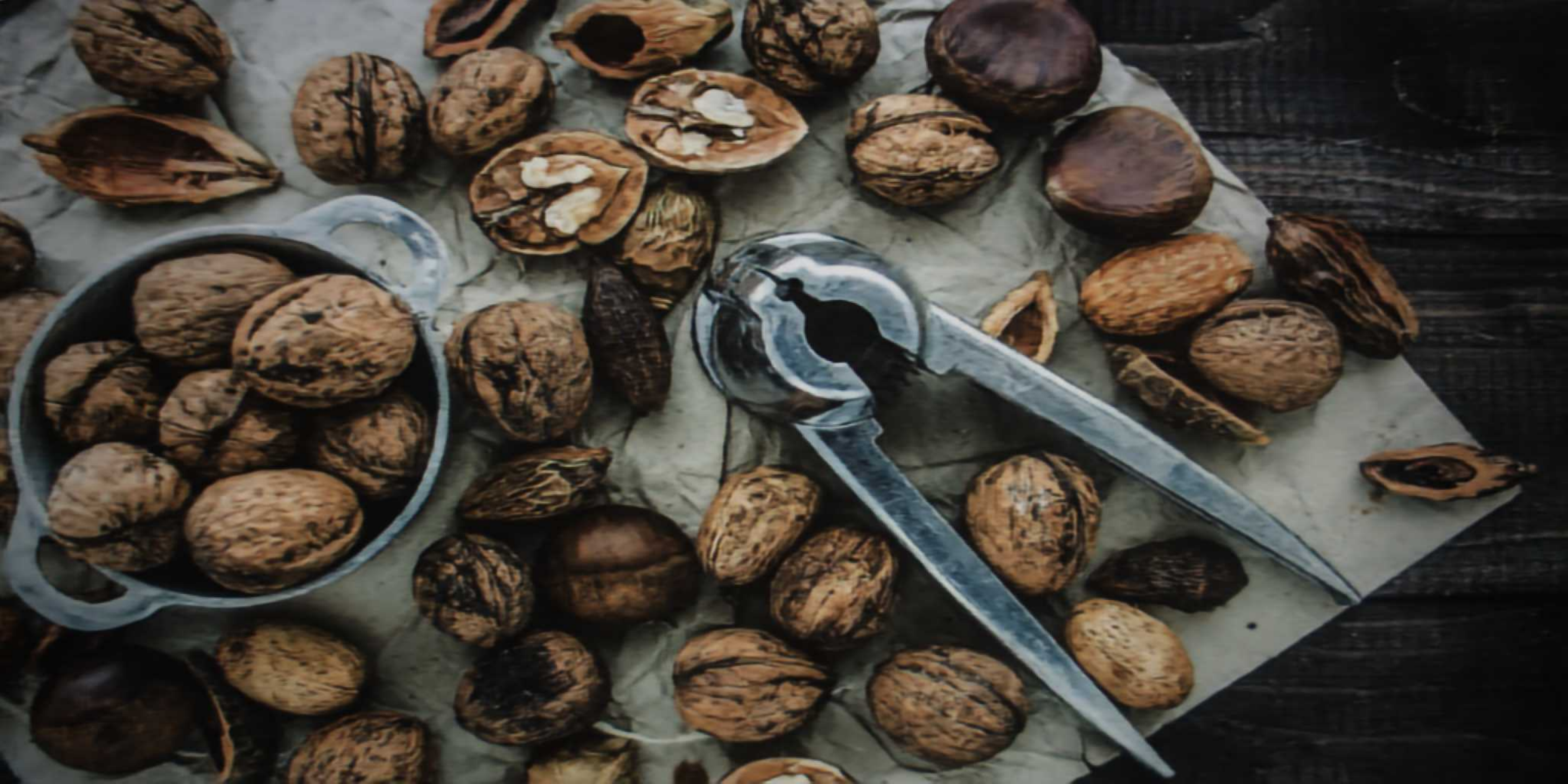}         &
			\includegraphics[width = 0.16\textwidth]{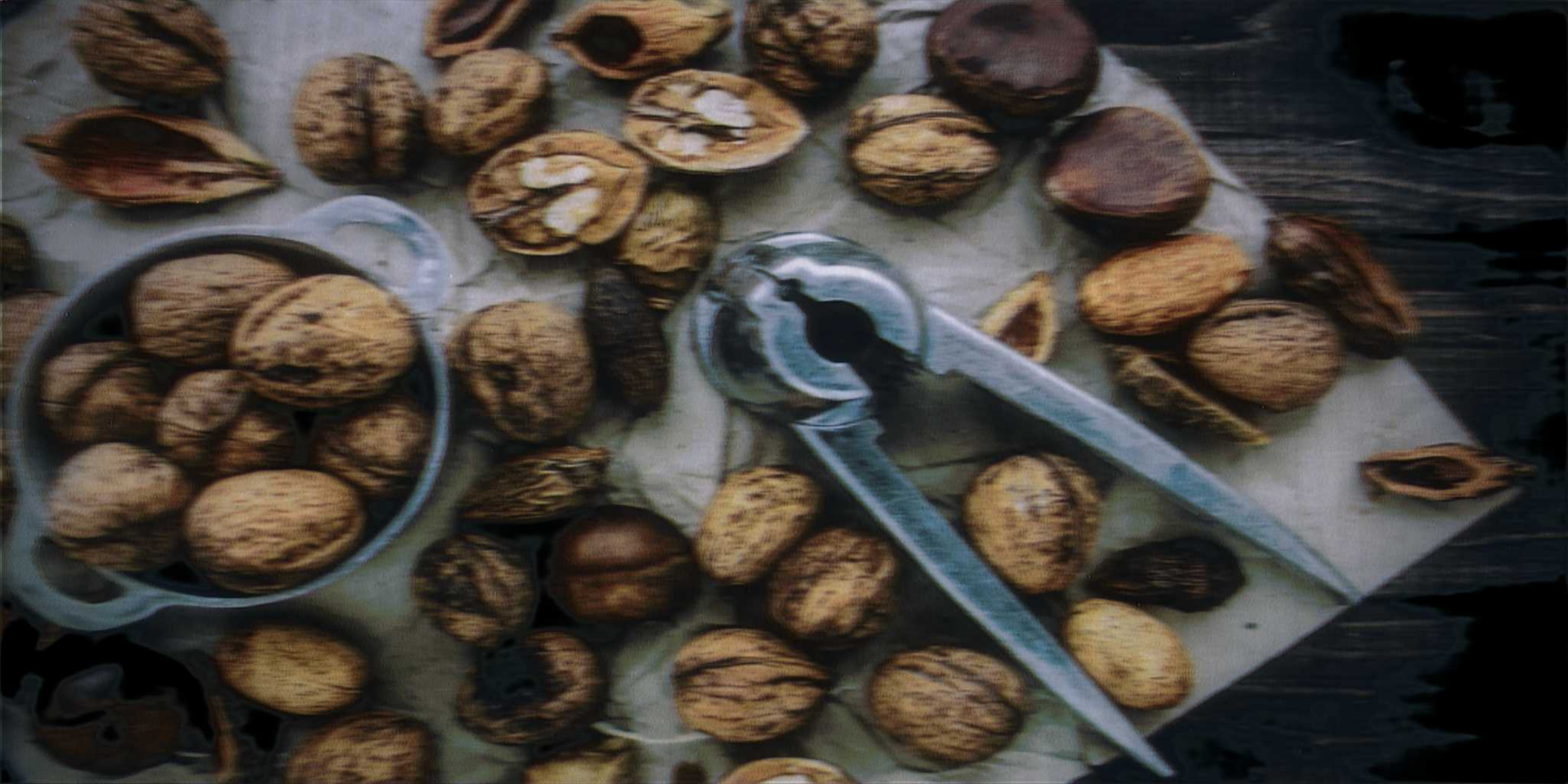}         &
			\includegraphics[width = 0.16\textwidth]{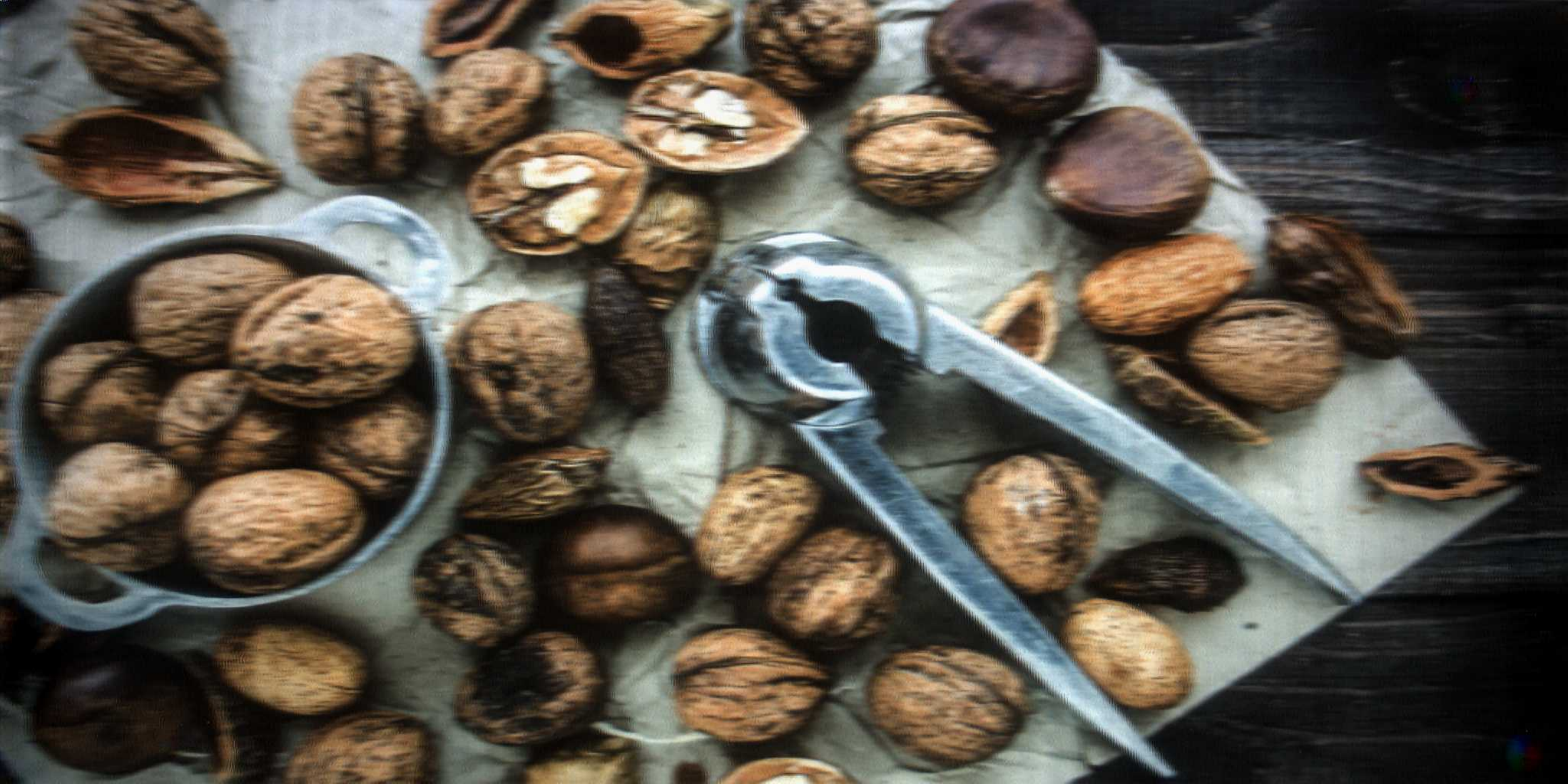}         \\

			(a) Input &
			(b) DCP + Zero-Net    &
			(c) HDRNet &  
			(d) DAGF   & 
			(e) DEUNet    & 
			(f) ZRUDC-Net    \\ 
			
			&
			\textcolor{blue}{CVPR'09} + \textcolor{blue}{CVPR'20}    &
			\textcolor{blue}{TOG'17} &  
			\textcolor{blue}{ECCV'20} & 
			\textcolor{blue}{CVPR'21}  & 
			\textcolor{blue}{Ours}
			\\ 
			
		\end{tabular}
	\end{center}
	\vspace{-3mm}
	\caption{Enhanced results on the T-OLED dataset. Our method yields a pleasing visual experience. HDRNet generates vivid colors on the test image, but many regions of the images are obviously overexposed.}
	\vspace{-2mm}
	\label{fig-sy_mydata}
\end{figure*}

\subsection{Non-Reference Loss Functions}
We propose a basket of differentiable non-reference losses to enable zero-reference learning on deep network.
The following six types of losses are enforced to train the ZRUDC-Net.

{\flushleft \textbf{Dark channel prior loss.}} 
The dark channel prior is an image statistical property, indicating that the darkest pixels in all color channels are very dark and close to zero in a small block of a haze-free outdoor image.
Golts et al.~\shortcite{dcploss} found an idea from the soft matting approach to design a loss function based on the dark channel prior.
%
%The approach designs a cyclic system on the optical model~\cite{Optics} to avoid overfitting while dehazing, which is an unsupervised strategy.
%
Mccartney and Hall~\shortcite{Optics} designed a cyclic system on the optical model to avoid overfitting while dehazing, which is an unsupervised strategy.
The dark channel prior loss $\mathcal{L}_{dcp}$ is a remarkably significant loss term for our algorithm, but it also depends much on the initialized configuration.
{\flushleft \textbf{Spatial consistency loss.}} 
We use a spatial consistency loss $\mathcal{L}_{s}$ to enhance image through preserving the difference of neighboring regions (left, right, down and top) between patches of the input image.
The enhanced formula is shown as follows:
\begin{equation}
\mathcal{L}_{s} = \frac{1}{P} \sum_{i=1}^{P} \sum_{j \in \Omega(i)}\left(\left|\left(O_{i}-O_{j}\right)\right|-\left|\left(I_{i}-I_{j}\right)\right|\right)^{2},
\end{equation}
%
%where $P$ is the number of local patches, $\Omega(i)$ is neighboring regions centered at the region $i$. $O$ and $I$ denote the average intensity value of the local region in the model output and input image.
%
where $P$ is the number of local patches, $\Omega(i)$ is neighboring regions centered at the region $i$. $O$ and $I$  denote the average intensity values of each local region in the model output and the input image, respectively.
The loss term empirically sets the size of the local region to $8 \times 8$ in this study.
{\flushleft \textbf{Exposure control loss.}} 
To address over-exposed regions, especially for the sky and the night lights, we use an exposure control loss $\mathcal{L}_{e}$ to keep the exposure degree.
The loss can be expressed as:
\begin{equation}
\mathcal{L}_{e} =\frac{1}{M} \sum_{k=1}^{M}\left|Y_{k}-E\right|,
\end{equation}
where $M$ represents the number of non-overlapping local regions of size $16 \times 16$, $Y$ is the average intensity value of a local region in the enhanced image.
$E$ represents the distance between the average intensity value of any local regions.
{\flushleft \textbf{Color constancy loss.}}
We use a color constancy loss $\mathcal{L}_{cc}$ to correct the color deviations in the enhanced image and reconstruct the relations among the three adjusted channels. 
This loss can be expressed as:
\begin{equation}
\mathcal{L}_{cc}=\sum_{\forall(p, q) \in \varepsilon}\left(J^{p}-J^{q}\right)^{2}, \varepsilon=\{(R, G),(R, B),(G, B)\},
\end{equation}
where $J^{p}$ denotes the average intensity value of $p$ channel in the enhanced image, $(p, q)$ represents a pair of channels.
{\flushleft \textbf{Illumination smoothness loss.}}
%
%We use the loss function to preserve the linear relations between neighboring pixels
%The illumination smoothness loss $L_{t v_{\mathcal{F}}}$ is defined as:
%
We use the illumination smoothness loss function $L_{t v_{\mathcal{F}}}$ to preserve the linear relations between neighboring pixels, which is defined as:
\begin{equation}
L_{t v_{\mathcal{F}}}=\frac{1}{N} \sum_{n=1}^{N} \sum_{c \in \xi}\left(\left|\nabla_{x} \mathcal{F}_{n}^{c}\right|+\nabla_{y} \mathcal{F}_{n}^{c} \mid\right)^{2}, \xi=\{R, G, B\}
\end{equation}
where $N$ represents the number of iteration, $\nabla_{x}$ and $\nabla_{y}$ represent the horizontal and the vertical gradient operations, respectively.
The four loss functions as described above are denoted by the term $\mathcal{L}_{lle}$ for convenience in the next description, and the weight coefficients of each term we adopt the native version~\cite{zerolli}.

{\flushleft \textbf{Dark and bright channels loss.}}
Enforcing a regularization term on the features of the bright and the dark channels is beneficial to restoring a sharp image, especially, for a sparse regularization term~\cite{dbloss}.
In this work, we use an  $\mathcal{L}_{1}$-regularization term to enforce sparsity on both dark and bright channels of shallow features.
The dark and bright channels loss $\mathcal{L}_{dbc}$ is defined as :
\begin{equation}
\mathcal{L}_{dbc} = ||D||_{1} + ||1 - B||_{1},
\end{equation}
where $D$ and $B$ are the conventional methods to extract the information of the dark channel and the bright channel, respectively.

The total loss can be expressed as :
\begin{equation}
L_{\text {total }}=w_{dcp}\mathcal{L}_{dcp} + w_{lle}\mathcal{L}_{lle} + w_{dbc}\mathcal{L}_{dbc},
\end{equation}
where $w_{dcp}$, $w_{lle}$ and $w_{dbc}$ are the weights of the losses.
In this work, we set the $w_{dcp}$, $w_{lle}$ and $w_{dbc}$ to $0.8$, $0.1$ and $0.1$, respectively.

\begin{figure*}[t]\scriptsize
	\begin{center}
		\tabcolsep 1pt
		\begin{tabular}{@{}cccccc@{}}
			
			\includegraphics[width = 0.192\textwidth]{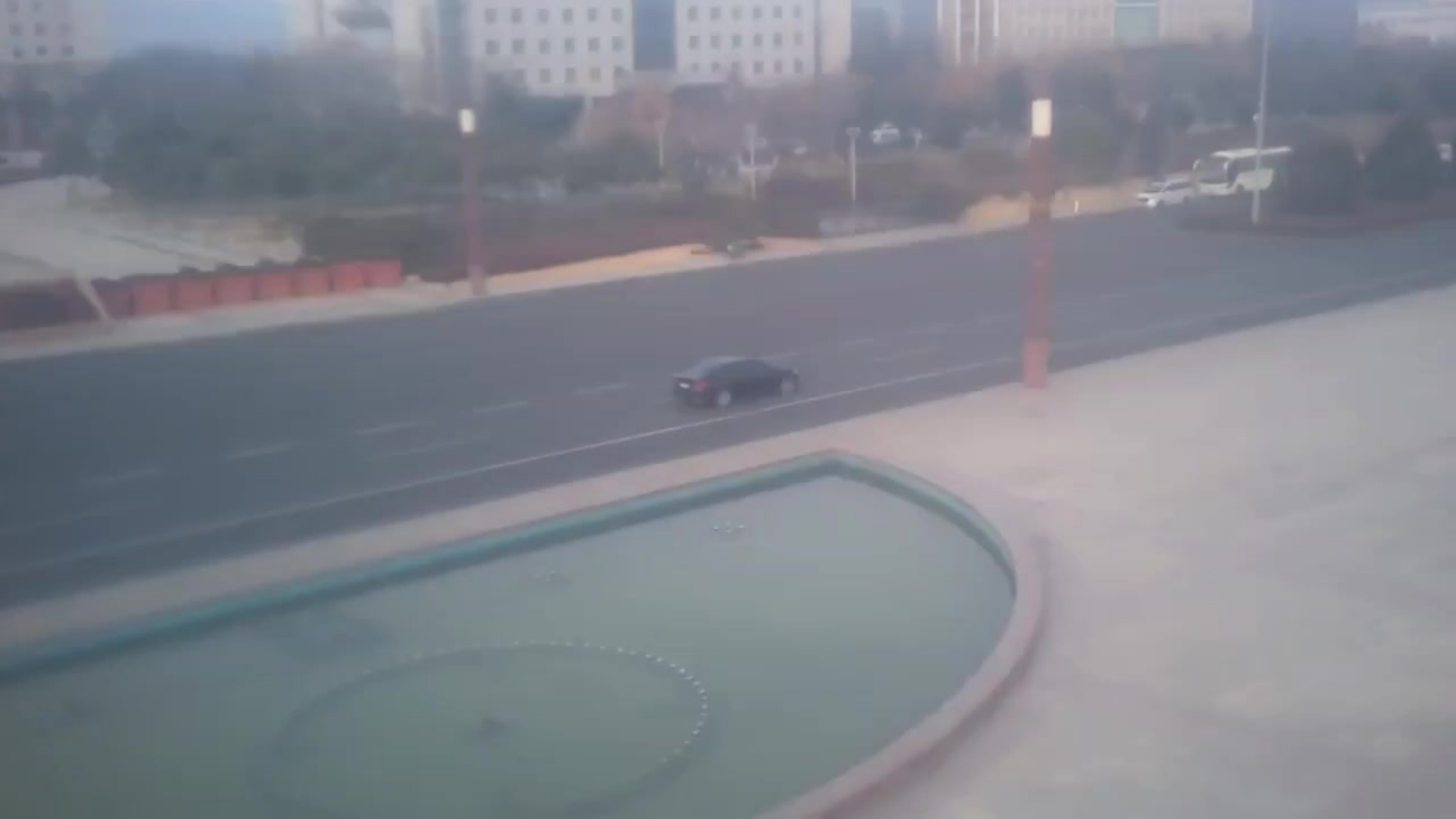}         &
			\includegraphics[width = 0.192\textwidth]{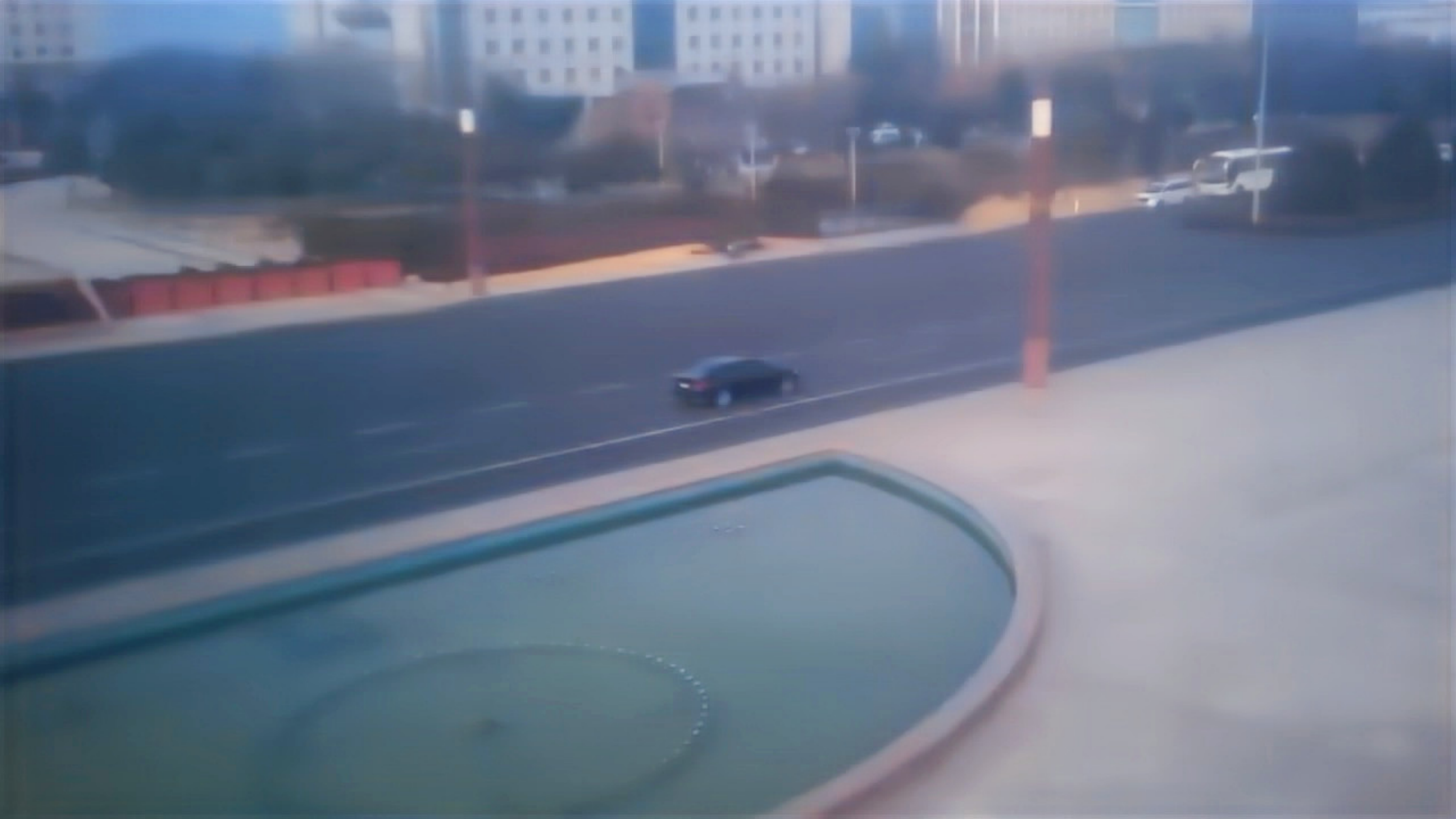}         &
			\includegraphics[width = 0.192\textwidth]{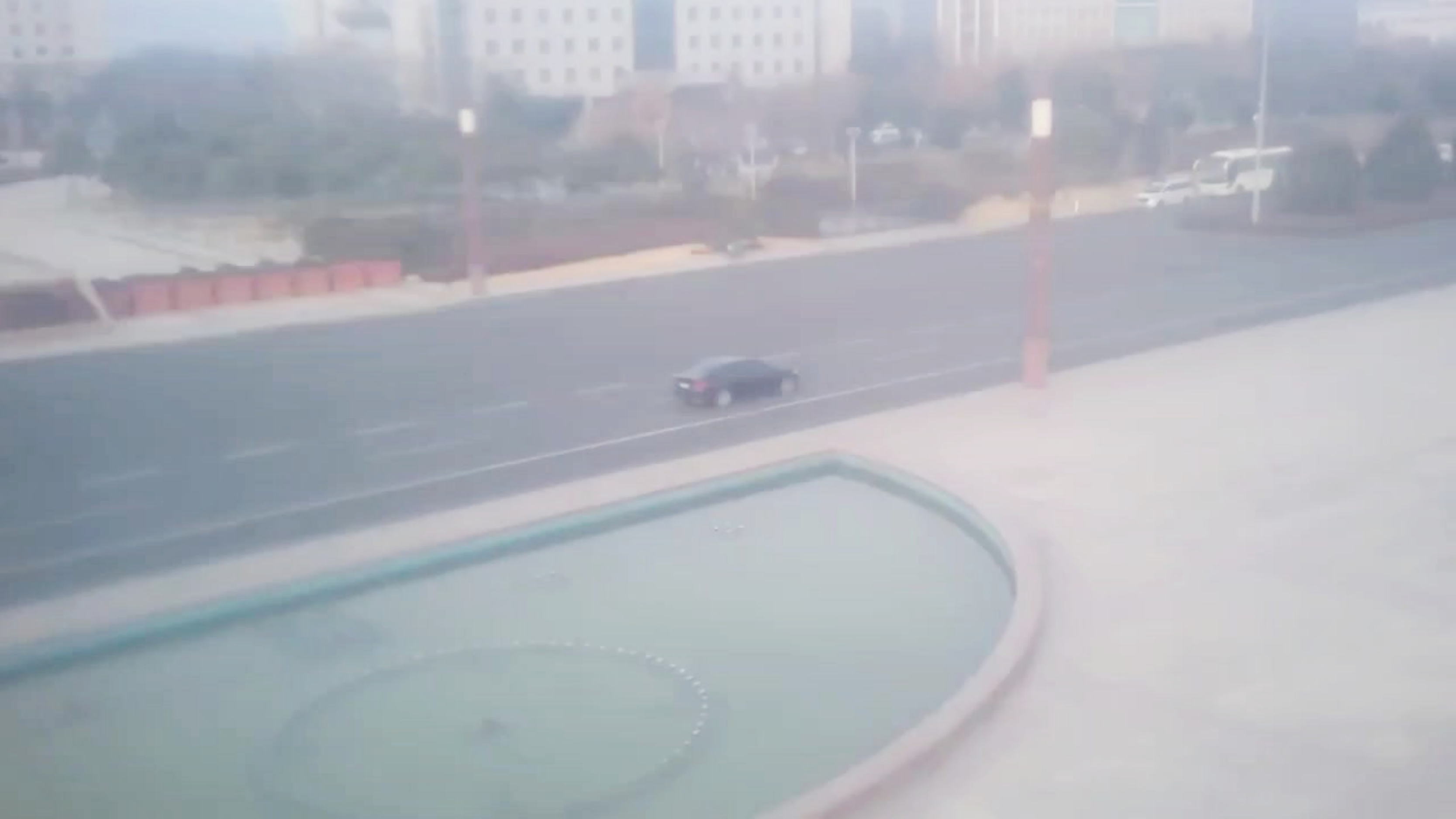}         &
			\includegraphics[width = 0.192\textwidth]{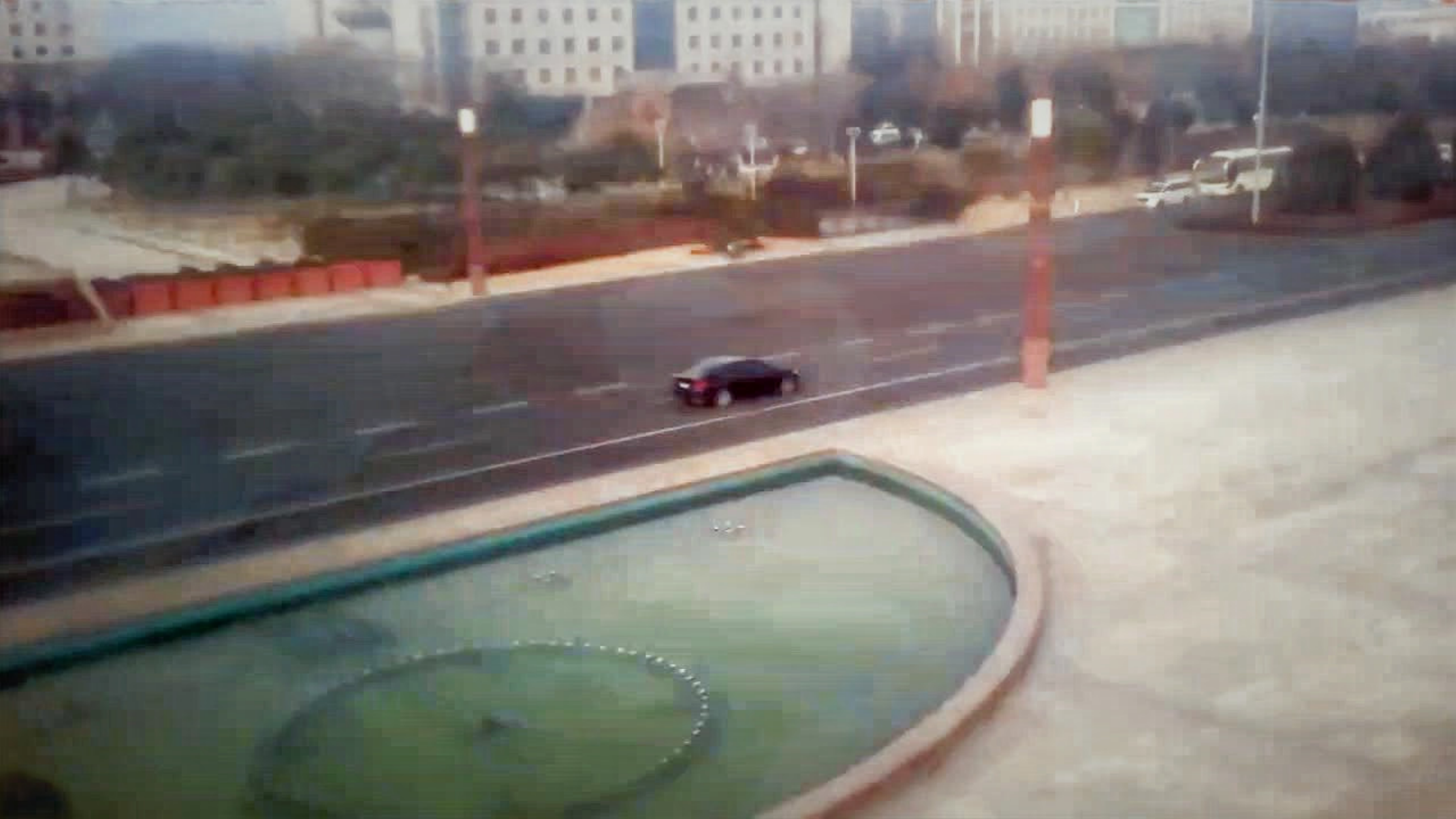}         &
			\includegraphics[width = 0.192\textwidth]{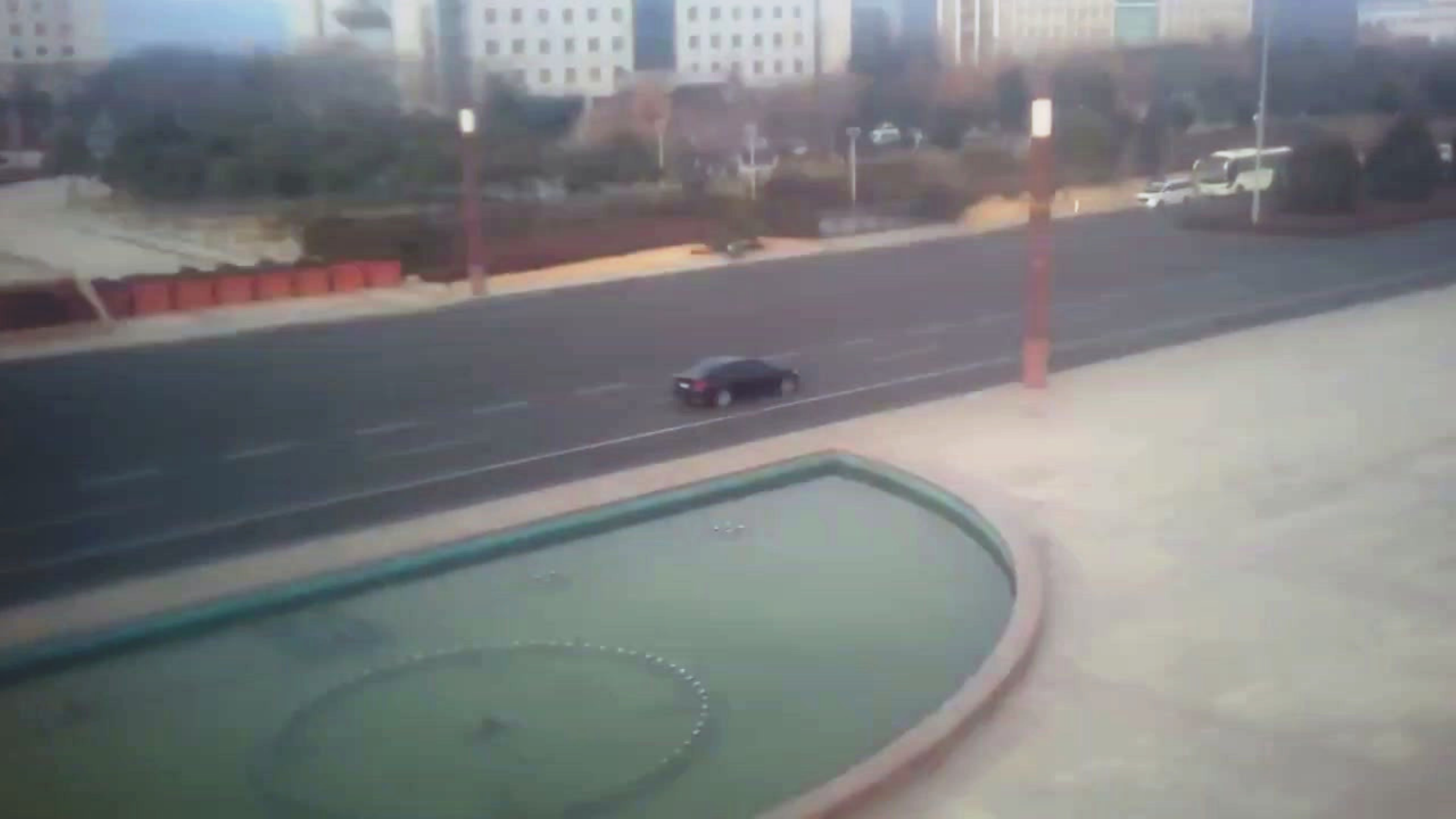}               \\
			
			(a) Input &
			(b) Ours    &
			(c) w/o $\mathcal{L}_{dcp}$  & 
			(d) w/o $\mathcal{L}_{lle}$   & 
			(e) w/o $\mathcal{L}_{dbc}$       \\ 
		\end{tabular}
	\end{center}
	\vspace{-6mm}
	\caption{Ablation experiments: without using $\mathcal{L}_{dcp}$ tends to generate haze in the image, without using $\mathcal{L}_{lle}$ tends to generate dark corner, and without using $\mathcal{L}_{dbc}$ makes the scene, such as vehicles and blurry.}
	\vspace{-6mm}
	\label{fig-abalation}
\end{figure*}

\section{Experiments}
In the section, we evaluate the proposed approach by conducting experiments on both a public dataset~\cite{eccvworks} and our collected real-world images (including records in extreme weather such as rainy).
All the results are compared against four image restoration methods:
DCP + Zero-Net~\cite{DCP,zerolli}, HDRNet~\cite{hdrnet}, DAGF~\cite{dagf}, DEUNet~\cite{irudc}. In addition, we conduct ablation studies to demonstrate the effectiveness of each module of our network.

{\flushleft \textbf{Implementation details.}} 
To train and test the proposed network as well as the comparison methods, we develop a new high resolution image restoration dataset on UAV with T-OLED, which consists of 30,000 frames blur images from 15 video clips by UAVs.
Since clean labels are missing, we can test the pre-trained model of the above method directly on this dataset.
The proposed model is implemented in PyTorch and Adam optimizer is used to train the model.
We use the full-resolution $1280 \times 720$ images with a batch size 21 to train the network.
The initial learning rate is set to 0.002. We train the network 100 epochs in total.
For DCP + Zero-Net method, we conduct different window sizes to trade-off the accuracy and the efficiency, and $45 \times 45$ is finally applied in the experiment.
In addition, to assess the situation in the extreme weather, we fine-tuned deep learning based models on the raindrop dataset~\cite{raindrop}. The fine-tuning methods all use a learning rate of 0.01, the SGD method, and a batchsize of 8.
%
%{\flushleft \textbf{Dataset.}} 
{\flushleft \textbf{Qualitative and Quantitative Evaluation.}}
The comparison experiments are evaluated on two datasets: Ours and the T-OLED dataset~\cite{irudc}.
Our model is fine tuned on the training data of the T-OLED dataset. 
Sample results on four images (including two images of rainy weather) from our dataset and four images from the T-OLED dataset, as shown in Figure~\ref{fig-Real_mydata} and Figure~\ref{fig-sy_mydata}.
DCP + Zero-Net and HDRNet tend to over-enhance the results (over-exposure),
while other models still cannot handle the real-world images (remain some haze in the results).
However, the restoration images generated by our method in Figure~\ref{fig-Real_mydata}-~\ref{fig-sy_mydata} (f) trade-offs the exposure and a pleasing visual perception is obtained.
The quantitative results on the T-OLED dataset reported in Table~\ref{tab:my-table} demonstrate the effectiveness of our proposal.
\vspace{-3mm}
\begin{table}[!h]
	%\small 
	\begin{center}
		\caption{Quantitative evaluations on the T-OLED (the resolution is 2K) test data in terms of PSNR, SSIM, LPIPS, and Time.}
		\vspace{-3mm}
		\label{tab:my-table}
		\begin{tabular}{cccccccccc}
			\hline
			\multicolumn{2}{l}{}     & \multicolumn{2}{l}{PSNR $\uparrow$}  & \multicolumn{2}{l}{SSIM $\uparrow$}   & \multicolumn{2}{l}{LPIPS $\downarrow$}  & \multicolumn{2}{c}{Time} \\ \hline
			\multicolumn{2}{l}{DCP + Zero-Net} & \multicolumn{2}{l}{21.22}          & \multicolumn{2}{l}{0.6990}          & \multicolumn{2}{l}{0.4289}   & \multicolumn{2}{c}{538ms}        \\
			\multicolumn{2}{l}{HDRNet}  & \multicolumn{2}{l}{29.81} & \multicolumn{2}{l}{0.8197} & \multicolumn{2}{l}{0.2779} & \multicolumn{2}{c}{20ms} \\
			\multicolumn{2}{l}{DAGF} & \multicolumn{2}{l}{35.89} & \multicolumn{2}{l}{0.9707} & \multicolumn{2}{l}{0.1255}  & \multicolumn{2}{c}{48ms}\\
			\multicolumn{2}{l}{DEUNet}     & \multicolumn{2}{l}{36.71}          & \multicolumn{2}{l}{\textbf{0.9713}} & \multicolumn{2}{l}{0.1209}   & \multicolumn{2}{c}{25ms}        \\
			\multicolumn{2}{l}{Ours}       & \multicolumn{2}{l}{\textbf{37.77}} & \multicolumn{2}{l}{0.9689}          & \multicolumn{2}{l}{\textbf{0.1189}} & \multicolumn{2}{c}{\textbf{18ms}} \\ \hline
		\end{tabular}%
	\end{center}
	\vspace{-4mm}
\end{table}
\vspace{-2mm}
{\flushleft \textbf{Ablation Study.}}
To demonstrate the effectiveness of each loss function and the 2D pool layer in ZRUDC-Net, we perform an ablation study involving the following two experiments:

1) w/o $\mathcal{L}_{dcp}$, w/o $\mathcal{L}_{lle}$, and w/o $\mathcal{L}_{dbc}$ :
We remove these loss functions respectively on training our proposed dataset.

2) Kernel size of the 2D pool layer: we use different kernel sizes of the 2D pool layer in our network to train the T-OLED dataset.
\vspace{-3mm}
\begin{table}[!h]
	\centering
	\caption{Effectiveness of the 2D pool scheme, and K denotes the kernel size, and K = None means removing the 2D pool layer in our network.}
	\vspace{-3mm}
	\label{t2}
	\begin{tabular}{clclclclcl}
		\hline
		\multicolumn{2}{c}{}     & \multicolumn{2}{c}{K=None}          & \multicolumn{2}{c}{K=3}            & \multicolumn{2}{c}{K=8}    & \multicolumn{2}{c}{K=16}   \\ \hline
		\multicolumn{2}{c}{PSNR} & \multicolumn{2}{c}{37.59}           & \multicolumn{2}{c}{\textbf{37.77}} & \multicolumn{2}{c}{36.13}  & \multicolumn{2}{c}{30.86}  \\
		\multicolumn{2}{c}{SSIM} & \multicolumn{2}{c}{\textbf{0.9701}} & \multicolumn{2}{c}{0.9689}         & \multicolumn{2}{c}{0.9554} & \multicolumn{2}{c}{0.9020} \\ \hline
	\end{tabular}%
	\vspace{-4mm}
\end{table}

Corresponding visual comparisons are shown in Figure~\ref{fig-abalation}.
Table~\ref{t2} compares our method against these three baselines on the T-OLED dataset, we use K=3 as the parameter in our network ($Adaptpool$ may be better, but inference time can increase).
\vspace{-3mm}
\section{Limitations and Discussion}
\begin{figure}[h] 
	\begin{center}
		\begin{tabular}{@{}c@{}}
			\includegraphics[width = 0.46\textwidth]{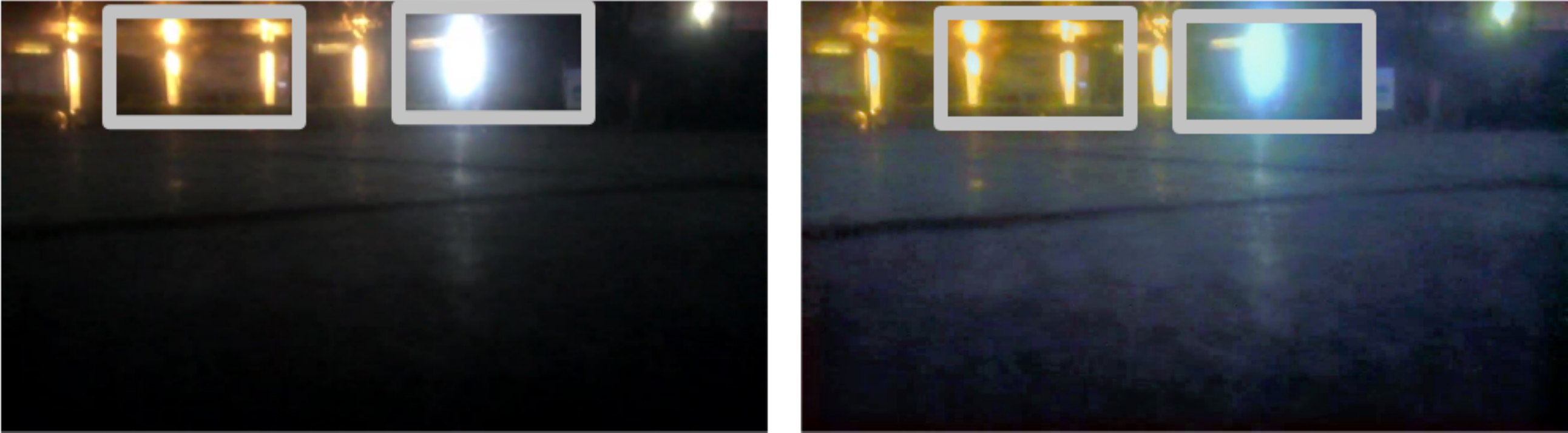}                   
		\end{tabular}
	\end{center}
	\vspace{-4mm}
	\caption{Image captured by the UAV with T-OLED in the dark (left), enhanced image by using our approach (right). The white box marked locations indicate the effect of the light information restoration.} 
	\vspace{-4mm}
	\label{f5}
\end{figure}

\vspace{-1mm}
In this study, we have two limitations: 1) we do not consider the UAV camera situation under P-LED, and 2) as shown in the Figure~\ref{f5}, we have bad image recovery at night, especially for the illumination information.
For our proposed deep network, we provide a more direct and efficient strategy to generate arbitrary resolution of the enhanced image (see Figure~\ref{fig-Real_mydata}) compared to bilateral learning methods.
In addition, for HDRNet~\cite{hdrnet}, we replace the additive operator with a convolutional operator in the slicing process, and the PSNR value is increased on the FIVEK dataset. More discussions are described in the supplementary materials.
%
%In addition, we discuss in the supplemental material about the effectiveness of battling HDRNet on the FIVEK dataset.

\vspace{-3mm}
\section{Conclusion}
\vspace{-1mm}
In this paper, we propose a zero-reference image restoration method via ZRUDC-Net and visual experiences. The key to our method is using different loss functions to build regression task, which effectively maintains the detailed edges and enhances the color of the image.
At the same time, we establish ZRUDC-Net to handle images of arbitrary resolution (including ultra high definition image) in real time. Quantitative and qualitative results show that our proposed method can generate visually-pleasing results on the non-clear images.

\clearpage

\bibliographystyle{named}
\bibliography{ijcai22}

\begin{thebibliography}{}

\bibitem[\protect\citeauthoryear{Afifi \bgroup \em et al.\egroup
  }{2021}]{HistoGAN}
Mahmoud Afifi, Marcus~A. Brubaker, and Michael~S. Brown.
\newblock {HistoGAN}: Controlling colors of {GAN}-generated and real images via
  color histograms.
\newblock In {\em CVPR}, 2021.

\bibitem[\protect\citeauthoryear{Cai \bgroup \em et al.\egroup }{2020a}]{dsd}
Jianrui Cai, Wangmeng Zuo, and Lei Zhang.
\newblock Dark and bright channel prior embedded network for dynamic scene
  deblurring.
\newblock {\em TIP}, 29:6885--6897, 2020.

\bibitem[\protect\citeauthoryear{Cai \bgroup \em et al.\egroup
  }{2020b}]{dbloss}
Jianrui Cai, Wangmeng Zuo, and Lei Zhang.
\newblock Dark and bright channel prior embedded network for dynamic scene
  deblurring.
\newblock {\em TIP}, 29:6885--6897, 2020.

\bibitem[\protect\citeauthoryear{Cheng \bgroup \em et al.\egroup }{2019}]{dtp}
Chi~Jui Cheng, Tzu~Chin Huang, Wen~Tsan Lin, Cheng~Chih Hsieh, Peng~Yu Chen,
  Peter Lu, and Hoang~Yan Lin.
\newblock Evaluation of diffraction induced background image quality
  degradation through transparent {OLED} display.
\newblock In {\em SID Symposium Digest of Technical Papers}, 2019.

\bibitem[\protect\citeauthoryear{Emerton \bgroup \em et al.\egroup
  }{2020}]{icta}
Neil Emerton, David Ren, and Tim Large.
\newblock Image capture through {TFT} arrays.
\newblock In {\em SID Symposium Digest of Technical Papers}, 2020.

\bibitem[\protect\citeauthoryear{Feng \bgroup \em et al.\egroup
  }{2021}]{removeudc}
Ruicheng Feng, Chongyi Li, Huaijin Chen, Shuai Li, Chen~Change Loy, and Jinwei
  Gu.
\newblock Removing diffraction image artifacts in under-display camera via
  dynamic skip connection network.
\newblock In {\em CVPR}, pages 662--671, 2021.

\bibitem[\protect\citeauthoryear{Gharbi \bgroup \em et al.\egroup
  }{2017}]{hdrnet}
Micha{\"{e}}l Gharbi, Jiawen Chen, Jonathan~T. Barron, Samuel~W. Hasinoff, and
  Fr{\'{e}}do Durand.
\newblock Deep bilateral learning for real-time image enhancement.
\newblock {\em TOG}, 36(4):118:1--118:12, 2017.

\bibitem[\protect\citeauthoryear{Golts \bgroup \em et al.\egroup
  }{2020}]{dcploss}
Alona Golts, Daniel Freedman, and Michael Elad.
\newblock Unsupervised single image dehazing using dark channel prior loss.
\newblock {\em TIP}, 29:2692--2701, 2020.

\bibitem[\protect\citeauthoryear{Guo \bgroup \em et al.\egroup
  }{2020}]{zerolli}
Chunle Guo, Chongyi Li, Jichang Guo, Chen~Change Loy, Junhui Hou, Sam Kwong,
  and Runmin Cong.
\newblock Zero-reference deep curve estimation for low-light image enhancement.
\newblock In {\em CVPR}, 2020.

\bibitem[\protect\citeauthoryear{He \bgroup \em et al.\egroup }{2009}]{DCP}
Kaiming He, Jian Sun, and Xiaoou Tang.
\newblock Single image haze removal using dark channel prior.
\newblock In {\em CVPR}, 2009.

\bibitem[\protect\citeauthoryear{Hirsch \bgroup \em et al.\egroup
  }{2009}]{bidi}
Matthew Hirsch, Douglas Lanman, Henry Holtzman, and Ramesh Raskar.
\newblock {BiDi} screen: a thin, depth-sensing {LCD} for {3D} interaction using
  light fields.
\newblock {\em TOG}, 28(5):1--9, 2009.

\bibitem[\protect\citeauthoryear{Jiang \bgroup \em et al.\egroup
  }{2021}]{englightengan}
Yifan Jiang, Xinyu Gong, Ding Liu, Yu~Cheng, Chen Fang, Xiaohui Shen, and
  Jianchao Yang.
\newblock {EnlightenGAN}: Deep light enhancement without paired supervision.
\newblock {\em TIP}, 30:2340--2349, 2021.

\bibitem[\protect\citeauthoryear{Kwon \bgroup \em et al.\egroup }{2016}]{mlt}
Hyeok-Jun Kwon, Chang-Mo Yang, Min-Cheol Kim, Choon-Woo Kim, Ji-Young Ahn, and
  Pu-Reum Kim.
\newblock Modeling of luminance transition curve of transparent plastics on
  transparent {OLED} displays.
\newblock {\em Electronic Imaging}, 2016(20):1--4, 2016.

\bibitem[\protect\citeauthoryear{Kwon \bgroup \em et al.\egroup
  }{2021}]{cirudc}
Kinam Kwon, Eunhee Kang, Sangwon Lee, and Su{-}Jin Lee.
\newblock Controllable image restoration for under-display camera in
  smartphones.
\newblock In {\em CVPR}, 2021.

\bibitem[\protect\citeauthoryear{Mccartney and Hall}{1977}]{Optics}
E.~J. Mccartney and F.~F. Hall.
\newblock Optics of the atmosphere: Scattering by molecules and particles.
\newblock {\em Phys. Today}, 1977.

\bibitem[\protect\citeauthoryear{Qian \bgroup \em et al.\egroup
  }{2018}]{raindrop}
Rui Qian, Robby~T. Tan, Wenhan Yang, Jiajun Su, and Jiaying Liu.
\newblock Attentive generative adversarial network for raindrop removal from a
  single image.
\newblock In {\em CVPR}, pages 2482--2491, 2018.

\bibitem[\protect\citeauthoryear{Qin \bgroup \em et al.\egroup }{2016}]{sti}
Zong Qin, Yu-Hsiang Tsai, Yen-Wei Yeh, Yi-Pai Huang, and Han-Ping~David Shieh.
\newblock See-through image blurring of transparent organic light-emitting
  diodes display: calculation method based on diffraction and analysis of pixel
  structures.
\newblock {\em Journal of Display Technology}, 12(11):1242--1249, 2016.

\bibitem[\protect\citeauthoryear{Qin \bgroup \em et al.\egroup }{2017}]{dsti}
Zong Qin, Jing Xie, Fang-Cheng Lin, Yi-Pai Huang, and Han-Ping~D Shieh.
\newblock Evaluation of a transparent display's pixel structure regarding
  subjective quality of diffracted see-through images.
\newblock {\em IEEE Photonics Journal}, 9(4):1--14, 2017.

\bibitem[\protect\citeauthoryear{Ronneberger \bgroup \em et al.\egroup
  }{2015}]{UNET}
Olaf Ronneberger, Philipp Fischer, and Thomas Brox.
\newblock {U-Net}: Convolutional networks for biomedical image segmentation.
\newblock In {\em MICCAI}, 2015.

\bibitem[\protect\citeauthoryear{Sethumadhavan \bgroup \em et al.\egroup
  }{2020}]{tdp}
Hrishikesh~Panikkasseril Sethumadhavan, Densen Puthussery, Melvin Kuriakose,
  and Jiji~Charangatt Victor.
\newblock Transform domain pyramidal dilated convolution networks for
  restoration of under display camera images.
\newblock In {\em ECCV}, 2020.

\bibitem[\protect\citeauthoryear{Suh \bgroup \em et al.\egroup
  }{2012}]{suh2012p}
Sungjoo Suh, Changkyu Choi, Kwonju Yi, Dusik Park, and Changyeong Kim.
\newblock An {LCD} display system with depth-sensing capability based on coded
  aperture imaging.
\newblock In {\em SID Symposium Digest of Technical Papers}, 2012.

\bibitem[\protect\citeauthoryear{Suh \bgroup \em et al.\egroup }{2013}]{add}
Sungjoo Suh, Changkyu Choi, Dusik Park, and Changyeong Kim.
\newblock Adding depth-sensing capability to an {OLED} display system based on
  coded aperture imaging.
\newblock In {\em SID Symposium Digest of Technical Papers}, 2013.

\bibitem[\protect\citeauthoryear{Sundar \bgroup \em et al.\egroup
  }{2020}]{dagf}
Varun Sundar, Sumanth Hegde, Divya Kothandaraman, and Kaushik Mitra.
\newblock Deep atrous guided filter for image restoration in under display
  cameras.
\newblock In {\em ECCV Workshops}, 2020.

\bibitem[\protect\citeauthoryear{Yang and Sankaranarayanan}{2021}]{ddpludc}
Anqi Yang and Aswin~C. Sankaranarayanan.
\newblock Designing display pixel layouts for under-panel cameras.
\newblock {\em TPAMI}, 43(7):2245--2256, 2021.

\bibitem[\protect\citeauthoryear{{Yuqian Zhou, David Ren, Kyle Tolentino, et
  al.}}{2020}]{eccvworks}
{Yuqian Zhou, David Ren, Kyle Tolentino, et al.}
\newblock {UDC} 2020 challenge on image restoration of under-display camera:
  Methods and results.
\newblock In {\em ECCV Workshops}, 2020.

\bibitem[\protect\citeauthoryear{Zhang}{2020}]{id}
Zhenhua Zhang.
\newblock Image deblurring of camera under display by deep learning.
\newblock In {\em SID Symposium Digest of Technical Papers}, 2020.

\bibitem[\protect\citeauthoryear{Zhou \bgroup \em et al.\egroup }{2021}]{irudc}
Yuqian Zhou, David Ren, Neil Emerton, Sehoon Lim, and Timothy~A. Large.
\newblock Image restoration for under-display camera.
\newblock In {\em CVPR}, 2021.

\end{thebibliography}

\end{document}